# A Review of Findings from Neuroscience and Cognitive Psychology as Possible Inspiration for the Path to Artificial General Intelligence

## Florin Leon

Department of Computer Science and Engineering
Faculty of Automatic Control and Computer Engineering
"Gheorghe Asachi" Technical University of Iași, Romania
Email: florin.leon@academic.tuiasi.ro

**Abstract**

This review aims to contribute to the quest for artificial general intelligence by examining neuroscience and cognitive psychology methods for potential inspiration. Despite the impressive advancements achieved by deep learning models in various domains, they still have shortcomings in abstract reasoning and causal understanding. Such capabilities should be ultimately integrated into artificial intelligence systems in order to surpass data-driven limitations and support decision making in a way more similar to human intelligence. This work is a vertical review that attempts a wide-ranging exploration of brain function, spanning from lower-level biological neurons, spiking neural networks, and neuronal ensembles to higher-level concepts such as brain anatomy, vector symbolic architectures, cognitive and categorization models, and cognitive architectures. The hope is that these concepts may offer insights for solutions in artificial general intelligence.

**Short Table of Contents**

# 1. Introduction

Artificial intelligence (AI) has witnessed substantial advancements in recent years, primarily driven by the emergence of deep learning models. They have proven to be highly effective in tackling complex problems that were previously very difficult to solve, particularly in the domain of primary processing dealing, e.g., with image, sound and language. Additionally, deep learning has led to significant breakthroughs in content generation through the development of generative models, allowing for the creation of text, question answering, and image synthesis.

The successes of deep learning models have been evident across various applications. In image processing, these models have achieved remarkable results in tasks such as object detection, image recognition, and semantic segmentation, often surpassing human-level performance in specific scenarios. Similarly, in sound processing, deep learning algorithms have demonstrated their capabilities in speech recognition and audio synthesis, giving rise to virtual assistants. Furthermore, in the domain of natural language processing, deep learning techniques have enabled machines to process and generate human language, leading to advancements in machine translation, sentiment analysis, and text summarization.

Generative models represent a significant advancement in AI, offering the ability to produce novel and realistic content across different modalities. For example, text-based conversational agents exhibit the capacity to generate coherent and contextually relevant responses to a given prompt. On the other hand, image generating models demonstrate the ability to create high-quality images based on textual descriptions. These advancements mark a notable leap in the expressive capabilities of AI systems and their potential for creative interactions with humans.

Despite these significant achievements, deep learning models have inherent limitations that hinder their progress towards artificial general intelligence (AGI). Their shortcomings become evident when tasked with abstract reasoning and the identification of causal relations. While deep learning methods excels at pattern recognition and data-driven tasks, they often struggle to comprehend the underlying causal mechanisms responsible for the observed patterns. This limitation is particularly noticeable in tasks demanding advanced abstract reasoning, where human intelligence has a distinct advantage over AI.

Another substantial drawback of deep learning models is their need for large amounts of training data to achieve good performance, making them less practical in scenarios where data is scarce, expensive to acquire, or simply unavailable. In contrast, humans possess a clear ability to learn from relatively limited data by leveraging abstract reasoning, understanding causality, and employing compact rules to generalize knowledge effectively.

The road to achieving AGI lies in addressing these limitations. Integrating reasoning and causality into AI systems will be instrumental in enabling them to move beyond data-driven tasks and make sound decisions and inferences, similar to human reasoning capabilities.

Furthermore, the accumulation of knowledge is very important in the pursuit of AGI. Beyond learning from large volumes of labeled data, AI systems should be equipped with the ability to accumulate knowledge similar to human common sense. This entails incorporating background knowledge about the world, contextual understanding, and the capacity to make decisions based on prior experiences and reasoning. In this manner, such systems could more closely resemble human abilities to compensate for incomplete facts or to act in unexpected situations.

In this review, we will provide an overview of research directions with the potential to advance AGI. While deep learning currently dominates the AI research landscape, it is important to recognize several areas that may not receive the mainstream attention they deserve. These fields, while seeming to be below the radar of the AI community, continue to engage in active research. Moreover, we should emphasize the significance of connecting AI research with the fields of neuroscience and cognitive psychology. Drawing inspiration from these disciplines can offer novel insights and opportunities to bring us closer to achieving AGI.

But even in the landscape of neuroscience and cognitive psychology, an obvious gap has been identified between the micro-level details of neural functioning and the macro-level phenomena observed in human cognition. At one end of the spectrum lies a good understanding of low-level neuronal operations. There is substantial knowledge about the inner workings of individual neurons and the functions of different brain regions. At the other end, cognitive psychology has identified and described high-level mechanisms that support complex mental processes such as memory, attention, and decision making. However, an important problem persists: the connection between the activity of neurons and the emergence of higher-order cognitive functions is still poorly understood.

In the following sections, we will investigate several perspectives concerning the characteristics of brain and mind function, spanning from the fundamental principles at lower levels to the more complex operations at higher levels of cognition. Our primary focus lies in describing the theoretical principles rather than exploring practical applications and computational optimizations, which often overshadow fundamental research.

It is also important to note that our aim is not to provide an exhaustive description of each concept, as the landscape is vast and diverse. The interested reader is encouraged to further explore these aspects in more depth. Many surveys tend to narrow their focus on specific domains or subfields. Our intention is to provide a more comprehensive perspective by addressing a broader spectrum of issues and trying to offer a more holistic approach of this complex journey towards understanding the human brain and possibly achieving AGI.

## 2. The Biological Neuron

We will begin by describing the behavior of a single biological neuron, which is the fundamental building block of the nervous system that transmits and processes information. In Section 3, we will present some computational models developed in this regard.

### 2.1. The Structure of the Neuron

The main components of a neuron are the cell body, the dendrites and the axon (Figure 2.1). The *cell body*, also known as the soma, is the central part of the neuron. Within it the nucleus resides, which contains the genetic material necessary for the functioning of the neuron, as well as various cellular organelles that support its metabolic processes.

*Dendrites*, often likened to tree branches, extend from the cell body and serve as the input receivers of the neuron. These dendritic extensions receive signals from other neurons in the form of electrical impulses.

The critical decision-making point within the neuron is the *axon hillock*, located at the point where the axon originates from the cell body. Here, the electrical inputs from the dendrites are integrated. If the cumulative strength of these inputs exceeds a certain threshold, it triggers the initiation of an electrical impulse, commonly known as an *action potential* or spike. The action potential is an "all-or-nothing" event, meaning that it either occurs in full strength or it does not occur at all. Once an action potential is generated at the axon hillock, it becomes irreversible and travels unidirectionally along the axon.

The *axon* is a long, cable-like structure responsible for transmitting the action potential signal to other neurons, often over considerable distances. This process ensures the propagation of information within the nervous system. At the end of the axon, there are specialized structures called *axon terminal buttons* (or boutons), which transmit the axon signal to the dendrites of neighboring neurons. This transmission is usually achieved through chemical *synapses*, where the axon terminal releases neurotransmitters, which in turn initiate electrical signals in the dendrites of the receiving neuron, continuing the chain of neural communication.

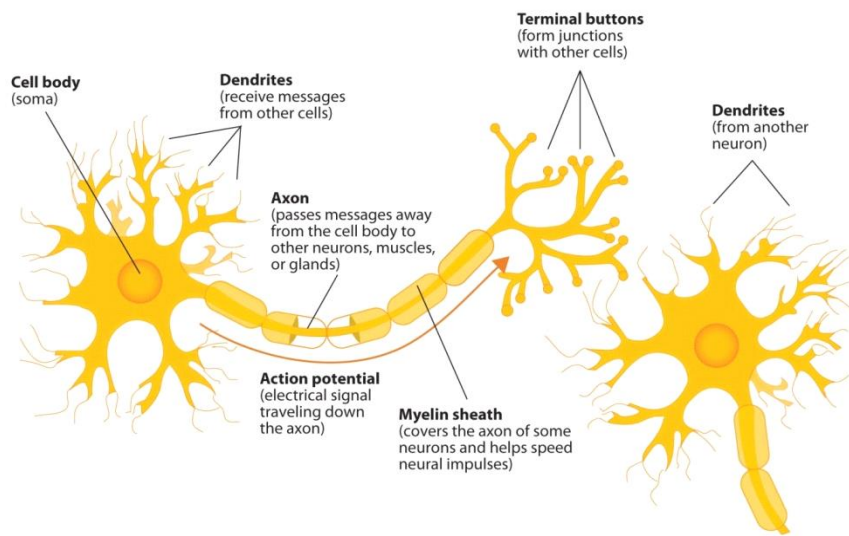

**Figure 2.1.** The components of a neuron
(Walinga, 2014)

It is important to mention that this is the typical description of neuronal signal transmission, but in the brain variations exist in both neuron structure and connection type. For example, as exceptions, some neurons have multiple axons or lack an axon altogether, and some synapses are electrical rather than chemical.

In general, a neuron can be thought of as a *detector* which responds to specific stimuli by evaluating the specific conditions it is wired to detect. Much like a detector in electronic systems, it recognizes and reacts to particular input patterns. The mathematical modeling of the activation of a neuron (the "all-or-nothing" response) closely resembles the functioning of a binary detector, where a certain threshold of input must be reached to trigger a response.

A neuron is an individual cell that generates and transmits electrical signals, while a *nerve* is a bundle of many axons held together by connective tissue, serving as a conduit for transmitting these signals between various parts of the body and the central nervous system.

Comparing neurons with other types of cells, we can mention that neurons are some of the longest-lived cells in the body. Many neurons are present throughout a person's lifetime and are not typically replaced. In contrast, most other cells in the body have a limited lifespan and are continually replaced through cell division. Also, neurons do not undergo cell division to reproduce. Thus, they have limited regenerative capacity; if they are damaged or die, they are usually not replaced through cell division but rather through processes like neural plasticity (adjustments in the structure of the connections between neurons) or by other neurons assuming some of the lost functions. The creation of new neurons (neurogenesis) does occur, but to a limited extent. Furthermore, neurons have high metabolic and energy demands due to their continuous signaling activity and the maintenance of ion gradients across their cell membranes, as we will see in the following sections. On average, the brain consumes about 20% of the body's total energy, although it represents only about 2% of the total body weight.

**2.2. The Structure of the Chemical Synapse**

We will next describe the basic structure and operation of a chemical synapse (Figure 2.2). A synapse is a microscopic junction where communication between neurons occurs. It serves as the bridge through which impulses are transmitted from one neuron to another.

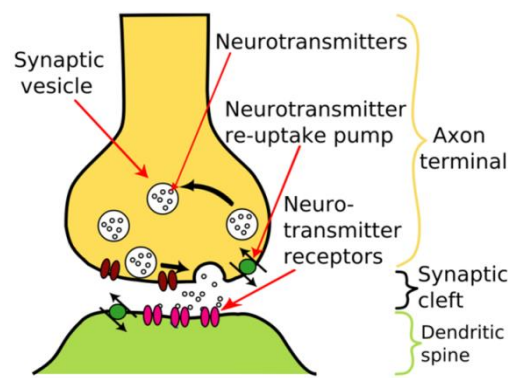

**Figure 2.2.** A chemical synapse
Adapted after (Nrets, 2008)

The transmitting neuron, known as the *presynaptic neuron*, and the receiving neuron, the *postsynaptic neuron*, are separated by a small gap called the *synaptic cleft*. When an action potential reaches the terminal button of the presynaptic neuron axon, it triggers the release of *neurotransmitters* from specialized structures called *synaptic vesicles*. The released neurotransmitters diffuse across the synaptic cleft. On the postsynaptic neuron membrane, there are receptor sites specifically designed to bind with the neurotransmitters. These receptor sites are like locks, and the neurotransmitters act as keys.

When the neurotransmitters bind to their receptors, they trigger a change in the postsynaptic neuron membrane potential. This change can be either *excitatory*, making it more likely for the postsynaptic neuron to fire an action potential, or *inhibitory*, making it less likely. The combined effect of these excitatory and inhibitory signals received by the postsynaptic neuron determines whether it will generate its own action potential and continue transmitting the neural impulse down the neural pathway.

After neurotransmitters are released into the synaptic cleft and have transmitted their signal from one neuron to another, they are cleared from the synapse to terminate the signal and prevent continuous stimulation. *Neurotransmitter reuptake pumps* are specific to certain neurotransmitters and are responsible for reabsorbing them back into the presynaptic neuron that released them.

## 2.3. The Formation of the Action Potential

Finally, we will make a simplified description of the formation of the action potential (AP) in a neuron (Figure 2.3).

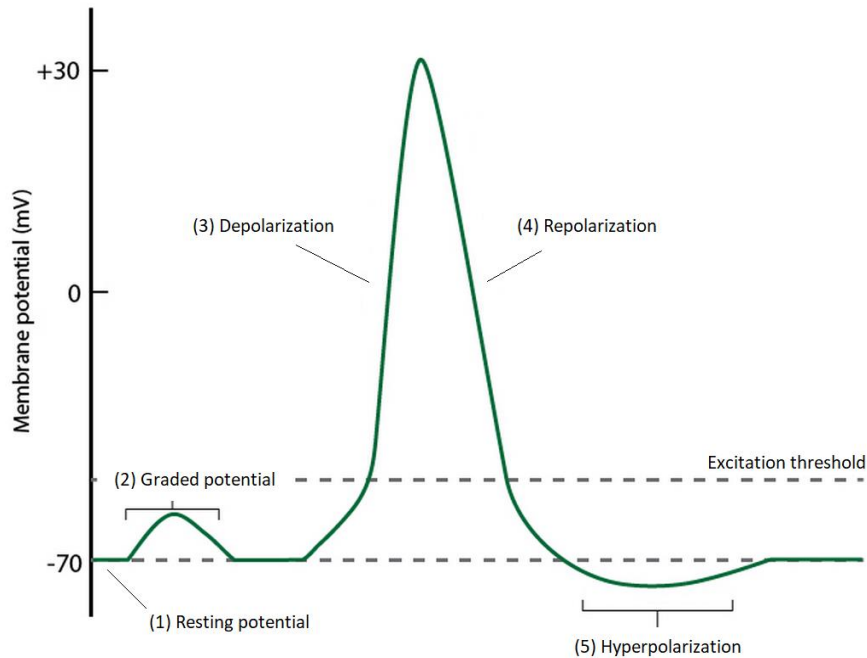

**Figure 2.3.** The phases of an action potential
Adapted after (Westin, 2023) and (Brigham Young University, 2023)

The resting and firing of a neuron depend on the interaction between electrical and chemical forces, involving various types of ions. In our description, we will focus only on the roles of *sodium* ($Na^+$) and *potassium* ($K^+$) ions.

In the resting state of the neuron, the majority of $Na^+$ ions are located outside the cell, while most $K^+$ ions are found inside the cell. The concentration of $Na^+$ outside the cell is approximately 10 times higher than inside, whereas the concentration of $K^+$ inside the cell is roughly 25 times greater than outside (Henley, 2021). This concentration imbalance (or gradient) creates a strong diffusion force, where the ions tend to form a more uniform distribution. However, the diffusion force is counteracted by the electrical force. When the neuron is in a resting state, it achieves an equilibrium when the electrical and chemical (diffusion) forces cancel each other out (O'Reilly et al., 2012). That is why the resting membrane potential of the neuron is about –70 mV (state 1 in Figure 2.3), and not zero.

When the neuron receives incoming signals, $Na^+$ ions enter the cell through *ligand-gated $Na^+$ channels*. These channels open in response to the binding of specific neurotransmitters. The influx of positive ions causes a positive change in the membrane potential. If the membrane potential remains below the activation threshold, a situation called a *graded potential* (stage 2 in Figure 2.3), the neuron does not generate an AP. Afterwards, the neuron membrane potential returns to its resting state, primarily due to the diffusion of positive ions, such as $K^+$, out of the cell until equilibrium is reached.

However, if the membrane potential exceeds the activation threshold, which is about –55 mV, it triggers the opening of *voltage-gated $Na^+$ channels*. These channels open only in response to a change in the electrical potential (voltage) of the cell membrane. Consequently, $Na^+$ ions rapidly flow into the cell, reversing the electrical charge of the membrane. This process is called *depolarization* (stage 3 in Figure 2.3), and it initiates an AP. Depolarization propagates along the cell membrane, enabling the transmission of electrical signals across the axon. As a result, the membrane potential reaches a positive value, around 30-40 mV.

At this point, *voltage-gated $K^+$ channels* open. Since $K^+$ ions are more concentrated inside the neuron, they flow down their electrochemical gradient to exit the cell. This outward movement of positively charged ions repolarizes the neuron membrane, a process called *repolarization* (stage 4 in Figure 2.3).

The movement of ions depends on their conductance through the ion channels in the neuron cell membrane (conductance is the reciprocal of electrical resistance: $G = 1 / R$). While the $Na^+$ conductance varies in a shape very similar to the AP itself in Figure 2.3, the $K^+$ conductance begins to increase later, near the AP peak, reaches its maximum level after the halfway point of the repolarization phase, and then decreases, but it does not reach its minimum value when the membrane attains its resting potential.

The voltage-gated $K^+$ channels remain open for a slightly longer period of time, creating the *hyperpolarization* phase (stage 5 in Figure 2.3), caused by the excessive outflow of $K^+$ ions. In addition, the neuron also has *K+ leak channels* that allow a continuous, passive outflow of K+ ions from the cell, which amplifies the efflux of positive ions. Now the neuron enters a *refractory* period, which makes it more difficult to generate another AP because a larger stimulus would be needed to reach the threshold. Hyperpolarization allows the neuron to recover before reacting to another signal. An excess of APs can be detrimental, leading to conditions like epileptic seizures or cardiac arrhythmia, while too few APs can result in pathologies such as muscle weakness, paralysis, or sensory and cognitive impairment. Therefore, hyperpolarization plays an essential role in helping the cell to maintain homeostasis (i.e., ensuring a stable and relatively constant internal environment).

Afterwards, the neuron membrane potential returns again to its resting state, primarily due to the inward diffusion of positive ions such as $Na^+$ through the membrane until electrochemical equilibrium is reached.

Although the presented mechanisms rely on the influx of $Na^+$ and efflux of $K^+$ ions, a single AP flow is small enough that it does not significantly alter intracellular concentrations of ions, and thus the equilibrium potential. However, on the long-term, multiple APs could gradually affect ion concentrations. The main means used to maintain the resting state are the *sodium-potassium pumps*. A $Na^+/K^+$ pump is an enzyme present in the membrane of all animal cells that has the role to expel $Na^+$ ions from the cell and to admit $K^+$ ions. In a functional cycle, it exchanges 3 $Na^+$ ions for 2 $K^+$ ions (Byrne, 2023). This is an active process that requires energy.

Thus, the small depletions that occur in $K^+$ and $Na^+$ concentrations after the APs are restored by the $Na^+/K^+$ pump. However, this replenishment is not immediately required for another AP. The ion gradients within a neuron are sufficient to generate approximately 10,000 APs without relying on the $Na^+/K^+$ pump.

A complete AP cycle has a short duration, between 1-5 ms, which can vary depending on the type of neuron and across different animal species. Pyramidal neurons in the cerebral cortex, which play a critical role in higher cognitive functions, can fire APs at rates ranging from a few spikes per second to more than 100 spikes per second, depending on the task and context.

The process presented above refers to an *excitatory synapse*, which enhances the likelihood of the postsynaptic neuron firing. In the brain, there are also *inhibitory synapses*, with an opposite role. This type of synapses operate through different mechanisms. For example, the primary neurotransmitter used in excitatory synapses is *glutamate*, while *gamma-aminobutyric acid* (GABA) serves as the primary neurotransmitter in inhibitory synapses. These two neurotransmitters are the most prevalent in the brain, with over half of all brain synapses using glutamate and approximately one-third using GABA (Genetic Science Learning Center, 2013).

The key mechanism in inhibitory synapses involves the movement of chloride ions ($Cl^-$) into the cell. This influx of negatively charged ions makes the interior of the neuron more negative (hyperpolarized), further away from the threshold required to initiate an AP. As a result, it becomes less likely for the neuron to reach its firing threshold.

## 2.4. Excitation and Inhibition

Although in multilayer perceptrons (MLP) the connection weights (that correspond to *synaptic efficacy* in biological neurons) can be positive or negative, in the brain neurons obey *Dale's principle* (Eccles, 1986), which states that a neuron cannot have both inhibitory and excitatory synapses leading from it (or that it can release either excitatory or inhibitory neurotransmitters, but not both simultaneously). Thus, we can consider that a biological neuron can be either excitatory or inhibitory.

Excitation is responsible for the transmission of signals across neurons, supporting information propagation. Excitatory neurons establish long-range connections between different cortical areas, and learning predominantly occurs at the synapses between them.

On the other hand, inhibition controls neural activity by preventing excessive firing and maintaining a balanced level of network activity. It is critical for temporal precision, synchronizing neural firing, enhancing contrast between signals, regulating network oscillations, and adjusting the sensitivity of neurons to inputs.

In the human cerebral cortex, the percentage of excitatory neurons is about 80-85% (Bratenberg, 1989; O'Reilly & Munakata, 2000; Nowak, Sanchez-Vives & McCormick, 2007), therefore the ratio between the numbers of excitatory and inhibitory neurons is about 4:1.

## 2.5. The Complexity of the Brain

The human brain is a very complex organ, with a wide range of neural mechanisms. At the core of this complexity is the sheer number of neurons and synapses. An adult brain is estimated to contain 86 ± 8 billion neurons (Azevedo et al., 2019). The neurons are connected by approximately 100 trillion synapses. On average, each neuron has about 10,000 dendrites to receive input from other neurons, and an axon connects to about 10,000 neurons to transmit its output (White, 1989; Abeles, 1991; Braitenberg & Schüz, 1998).

Furthermore, the brain is composed of various types of neurons, each with unique characteristics and functions. The axons can be several millimeters long in cortical neurons, but the longest neurons in the human body are those in the sciatic nerve, where axons can reach lengths of up to 60 cm.

There are more than 100 neurotransmitters (Eliasmith, 2015) that play an important role in neural communication. Beside glutamate and GABA mentioned earlier, some of the most well-known neurotransmitters include dopamine, serotonin, acetylcholine and adrenaline, each with distinct functions and effects on behavior and cognition.

# 3. Neuron Models

The most common type of artificial neuron encountered today is the perceptron, which serves as a fundamental building block in various forms of MLP neural networks. This is also called the *point neuron* model. Its output is expressed as:

$$y = f\left(\sum_{i=1}^{n} w_i \cdot x_i + b\right) \tag{3.1}$$

where $y$ represents the output, $f$ is a nonlinear activation function, **w** and **x** denote the weights and inputs, respectively, and $b$ is the bias term. The activation function may be sigmoidal (e.g., the hyperbolic tangent), piecewise linear (e.g., the rectified linear unit, ReLU), or may involve further variations.

However, other models exist, which strive for a closer resemblance to biological neurons. The behavior of biological neurons has been a subject of great interest in computational neuroscience, and researchers have tried to create computational models that mimic the function of real neurons. In this section, we describe several such models.

## 3.1. Hodgkin-Huxley Model

The foundational mathematical framework that underlies contemporary biophysically-oriented neural modeling was established by Hodgkin and Huxley (1952). They conducted a series of experiments on the squid giant axon, notable for its very large diameter of approximately 0.5 mm. In comparison, most axons in the squid nervous system and in the nervous systems of other species are much thinner, e.g., the typical neuron diameter in humans is 1-2 μm.

They represented the electrical properties of a neuron in terms of an equivalent electrical circuit. A capacitor denotes the cell membrane, two variable resistors represent the voltage-gated Na$^+$ and K$^+$ channels, a fixed resistor symbolizes a small leakage current due to Cl$^-$ ions, and three batteries represent the electrochemical potentials created by the differing intracellular and extracellular ion concentrations.

The basic equation of the model is:

$$C\frac{dV}{dt} = I_i + I_s \tag{3.2}$$

where $I_s$ is the stimulus current and $I_i$ is the ionic current, represented as a sum of three currents – Na$^+$, K$^+$ and the leak:

$$I_i = -G_K(V - V_K) - G_{Na}(V - V_{Na}) - G_l(V - V_l) \tag{3.3}$$

$$I_i = -\bar{g}_K n^4 (V - V_K) - \bar{g}_{Na} m^3 h (V - V_{Na}) - \bar{g}_l (V - V_l) \tag{3.4}$$

The gating variables $n$, $m$, and $h$ are also modeled by additional equations:

$$\frac{df}{dt} = \frac{f_\infty(V) - f}{\tau_f(V)}, \quad f \in \{n, m, h\} \tag{3.5}$$

$$\tau_f(V) = \frac{1}{\alpha_f(V) + \beta_f(V)} \tag{3.6}$$

$$f_\infty(V) = \alpha_f(V) \cdot \tau_f(V) \tag{3.7}$$

We included these equations to emphasize the complexity of this widely recognized mathematical model for the behavior of neurons. However, we will not delve into a discussion about the meaning and specific values of the parameters involved. They are briefly presented in Table 3.1 (Nelson, 2004; Coombes, 2010; Wells, 2010).

**Table 3.1.** Parameters of the Hodgkin-Huxley model

| | |
|---|---|
| $C$ | the cell capacitance |
| $V$ | the membrane potential |
| $I_s$ | the stimulus current |
| $I_i$ | the ionic current |
| $G_K, G_{Na}, G_l$ | the conductances of the $K^+$, $Na^+$ and leak channels |
| $\bar{g}_K, \bar{g}_{Na}, \bar{g}_l$ | the maximal values of conductance of the $K^+$, $Na^+$ and leak channels |
| $V_K, V_{Na}, V_l$ | reversal potentials |
| $n, m$ | activation variables |
| $h$ | inactivation variable |
| $f$ | a generic notation for $n$, $m$ and $h$, so that there are actually 3 equations for each of the equations (3.5, 3.6, 3.7) |
| $f_\infty(V)$ | the steady-state open probabilities for the three gates, related to the fraction of channels in the open state |
| $\tau_f(V)$ | time constants |
| $\alpha_f(V), \beta_f(V)$ | rate constants |

The values of some of these parameters are approximated so that the equations fit experimental data. We should mention that equations (3.5, 3.6, 3.7) are slightly different from those in the original paper (Hodgkin & Huxley, 1952), because the differential equations are solved at present with numerical techniques such as the Euler method, while in the 1950s the authors had to use a more primitive method for manual calculation (Wells, 2010). Also, the AP values used in their study for squid are a little different from those measured in human neurons (presented in Figure 2.3).

When no external stimulus is present, the dynamics of the Hodgkin-Huxley model reach a stable state in the resting potential. That is why, in order to produce an AP, a stimulus current $I_s$ must be added. An example of how a Hodgkin-Huxley neuron performs in the presence of a step input current is shown in Figure 3.1.

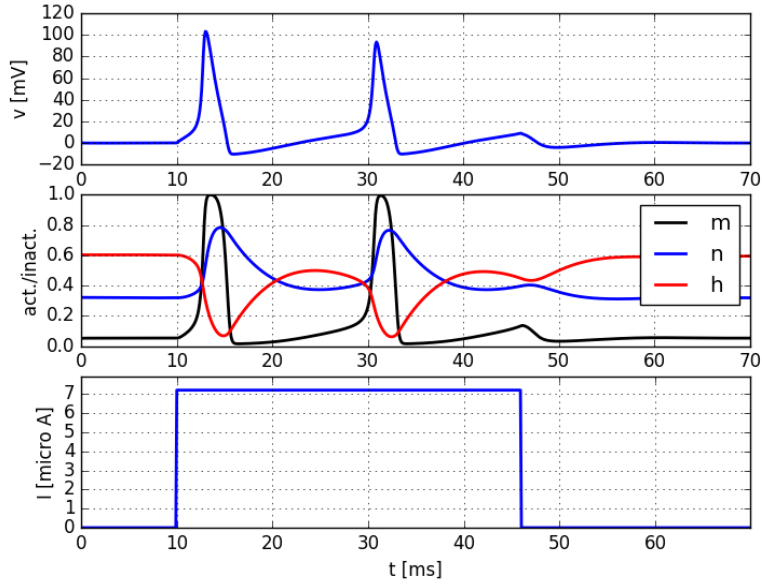

**Figure 3.1.** The activation of a Hodgkin-Huxley neuron model
of the squid axon for a step input current
(EPFL-LCN, 2016)

**3.2. Izhikevich Model**

As we could see, the Hodgkin-Huxley model is quite complex. Eugene Izhikevich (2003) devised a much simpler model that can nevertheless reproduce a large range of behaviors found in biological neurons. It consists of only two differential equations and a non-linear test condition:

$$\frac{dV}{dt} = 0.04V^2 + 5V + 140 - U + I_s \qquad (3.8)$$

$$\frac{dU}{dt} = a(bV - U) \qquad (3.9)$$

if $V \geq \theta$, then $V = c$ and $U = U + d$ \qquad (3.10)

$V$ and $U$ are conceived as dimensionless variables; however, with the coefficients in the first equation found by fitting experimental data with the dynamics of a cortical neuron, $V$ can be interpreted as the membrane potential of the neuron and $U$ as a variable that accounts for the effects of the ionic currents and gives negative feedback to $V$. The model also provides suitable values for the parameters $a$, $b$, $c$, and $d$. Usually, the threshold $\theta = 30$.

Beside its potential for modeling a great variety of behaviors, the main advantage of this model is its computational efficiency. The author reports that tens of thousands of spiking neurons can be simulated in real time with 1 ms resolution on a 1 GHz computer (Izhikevich, 2003).

## 3.3. Integrate and Fire (IF) and Leaky Integrate and Fire (LIF) Neurons

An even simpler model is the *leaky integrate and fire* (LIF) model (Koch & Segev, 1998). The neuron behaves like an integrator: its membrane potential increases when it receives an external stimulus or input signal *I*. The membrane potential decays over time, therefore when no stimuli are present, the voltage slowly returns to the resting potential $V_r$. When the voltage *V* reaches the threshold, the neuron produces a spike (an action potential), and then *V* is reset to $V_r$. By analogy to the electrical circuit considered in Section 3.1, the LIF model is based on a differential equation of the following form (where $\tau$ is the membrane time constant, which has a similar role as the reciprocal of a learning rate):

$$\tau \frac{dV}{dt} = -(V(t) - V_r) + I(t) \tag{3.11}$$

However, software simulations usually employ discrete time dynamics. That is why an equivalent formulation of the LIF model is (Jin et al., 2022):

$$V(t+1) = \alpha \cdot V(t) + I(t) \tag{3.12}$$

$$\text{if } V(t) \geq 1, \text{ then spike and } V(t+1) = 0 \tag{3.13}$$

In this equation, *V* is the membrane potential or voltage, *I* is the input current or stimulus, and $\alpha$ is the leak or decay factor, which determines how quickly the membrane potential goes to 0 over time. One can notice that equations (3.12, 3.13) model a normalized activity, where the resting potential is 0 and the threshold where a spike is produced is 1.

Optionally, after a spike, one can keep *V* = 0 for *n* more time steps, simulating the refractory period where the neuron cannot produce another spike. In equation (3.12), $\alpha \in (0, 1)$ is usually closer to 1 (e.g., 0.9), and the stimuli $I \in [0, 1]$ are usually closer to 0, to allow multiple time steps until spiking (e.g., 0.1 or 0.2).

While less biologically detailed than other models, LIF provides a computationally efficient way to simulate spiking neural networks.

A even simpler model is *integrate and fire* (IF), which is similar to LIF but does not include the decay factor:

$$V(t+1) = V(t) + I(t) \tag{3.14}$$

$$\text{if } V(t) \geq 1, \text{ then spike and } V(t+1) = 0 \tag{3.15}$$

## 3.4. Spike Frequency Adaptation

*Spike frequency adaptation* (SFA) is a physiological phenomenon observed in neurons, i.e., the tendency of the firing rate of a neuron to decrease over time in response to a sustained input stimulation. In other words, as a neuron receives a continuous input, its firing rate gradually decreases, resulting in fewer APs being generated over time. This phenomenon is also known as *accommodation*, where the neuron becomes less responsive to rapid changes in input and is more sensitive to changes occurring over longer timescales.

When used in conjunction with LIF neurons, the SFA idea translates in increasing the threshold of the neuron following a spike and decreasing the threshold to its normal value (such as 1) when no spikes occur. The following equations describe this process. Instead of 1, $\theta(t)$ is used as a dynamic threshold. $\beta$ and $\gamma$ are parameters that control how the threshold increases and decreases.

$$V(t+1) = \alpha \cdot V(t) + I(t) \tag{3.16}$$

$$\text{if } V(t) \geq \theta(t), \text{ then spike and } V(t+1) = 0 \tag{3.17}$$

$$\theta(t) = 1 + \beta \cdot a(t) \tag{3.18}$$

$$a(t+1) = \gamma \cdot a(t) + (1-\gamma) \cdot s(t) \tag{3.19}$$

$$s(t) = 1 \text{ if a spike occurred at time step } t, \text{ and } 0 \text{ otherwise} \tag{3.20}$$

Figure 3.2 shows a comparison between the LIF neuron and a LIF neuron with SFA.

From a computational point of view, SFA can have various functional implications, e.g., it can contribute to regulating the balance between excitatory and inhibitory signals in spiking neural networks (presented in Section 4).

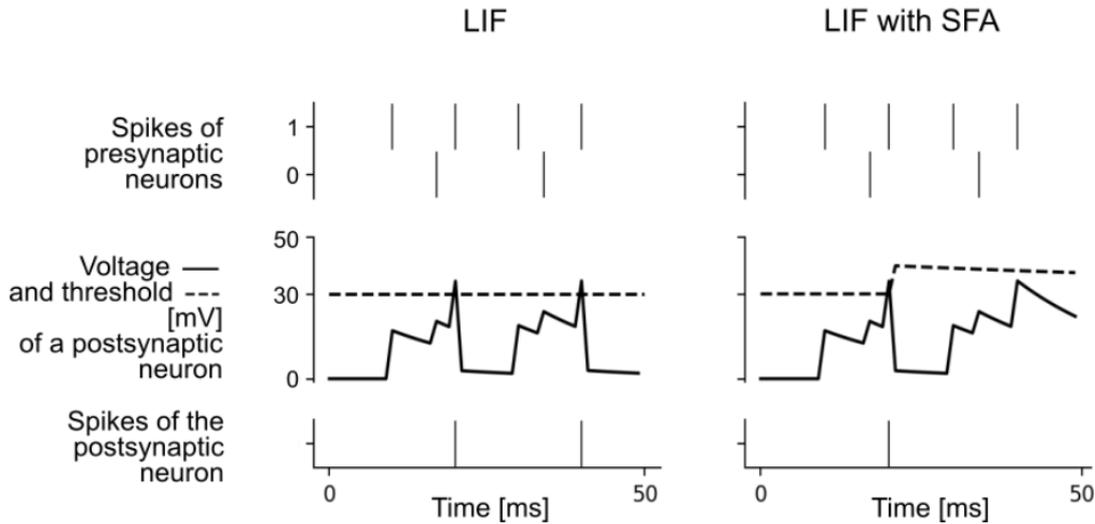

**Figure 3.2.** Neuron dynamics for a leaky integrate and fire neuron with and without spike frequency adaptation
(Kraišniković, Maass & Legenstein, 2021)

## 3.5. The Compartmental Neuron

In many models, including the multilayer perceptron, synapses are treated in a uniform manner, with no consideration for specific dendrite type or dendritic processing. However, in contrast to this conventional approach, the compartmental neuron model recognizes the importance of distinguishing between different types of dendrites, namely *proximal (basal)* and *distal (apical)*. In Section 10.1, we will see how this differentiation is applied in the context of hierarchical temporal memory, a model inspired from neuroscience.

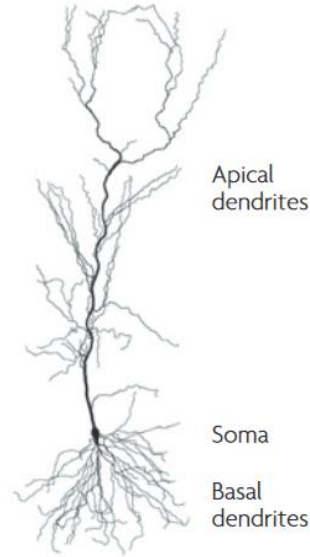

**Figure 3.3.** Pyramidal neuron structure and domains of synaptic input
(Spruston, 2008)

Proximal (near) dendrites are close to the neuron cell body, serving as the primary sites for receiving incoming signals from neighboring neurons. Distal (remote) dendrites extend further away from the cell body, often into more remote regions of the neuron receptive field.

In (Schubert & Gros, 2021), a two-compartment model is presented, which accounts for the nonlinear interactions between proximal and distal inputs. The main motivation of the study is that some dynamical properties of biological pyramidal neuronal cells in the cortex cannot be reproduced using a simple point neuron model (Spruston et al., 1995; Häusser et al., 2000). When synaptic inputs from proximal and distal sources coincide, the neuron exhibits a high spiking activity, which is larger than the maximum firing rate possible only through proximal synaptic input. This indicates a form of temporal coincidence detection between distal and proximal stimuli.

The neuron model presented here is a discrete-time rate-encoding model, i.e., its output signifies the spiking frequency of the neuron at a certain time step. Actually, this means that its output is a real-valued number, just like the output of a classical MLP neuron. The model assumes two different input variables $I_p$ (total proximal input) and $I_d$ (total distal input). They can be positive or negative, again like in the case of an MLP neuron. The output of the compartmental neuron is:

$$y = \alpha \cdot \sigma(I_p) \cdot (1 - \sigma(I_d)) + \sigma(I_d) \cdot \sigma(I_p + 1) \tag{3.21}$$

where $y$, $I_p$ and $I_d$ are in fact functions of time (e.g., $y = y(t)$), $\alpha \in (0,1)$ and $\sigma$ is a sigmoid function:

$$\sigma(x) = \frac{1}{1 + e^{-4x}} \tag{3.22}$$

The paper also uses some thresholds in equation (3.21), but since their values are also provided (e.g., some thresholds are set to 0, another to –1), those values are directly included, to simplify the expression. This equation is itself a simplified version of a phenomenological model described in (Shai et al., 2015).

According to this equation, there are two distinct regions of neural activation in the ($I_p$, $I_d$) plane, as shown in Figure 3.4. When both input currents $I_p$ and $I_d$ are large, the second term dominates, which leads to $y \approx 1$. A medium activity plateau appears when $I_p$ is positive and $I_d$ is negative. As such, the compartment model distinguishes neurons with a lower activity level (e.g., at $\alpha = 0.3$), and strongly bursting neurons, where the maximum firing rate is 1. In the medium activity plateau, neurons process the proximal inputs in the absence of distal stimulation. The distal current acts as a modulator. In the figure, the region with the maximal value of 1 appears in the top right quadrant, where both types of inputs are active, and the region with the value $\alpha = 0.3$ appears in the bottom right quadrant. This means that only 30% of the maximal firing rate can be achieved when only the proximal inputs are active.

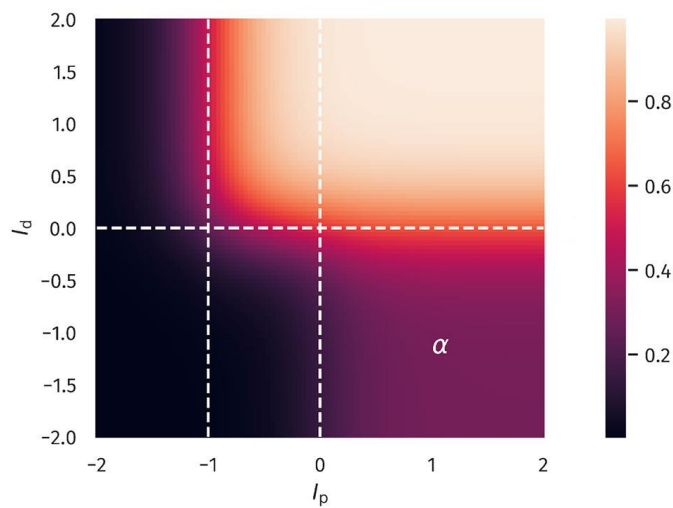

**Figure 3.4.** Firing rate as a function of proximal and distal inputs
(Schubert & Gros, 2021)

## 4. Spiking Neural Networks

*Spiking neural networks* (SNN) are designed to closely mimic the behavior of real neurons, emphasizing the temporal aspect of neural signaling. Unlike traditional artificial neural networks such as the multilayer perceptrons, where activations are typically continuous values, SNNs use discrete events (spikes) to represent neuron activations. The timing and rate of these spikes allow the SNNs to capture precise temporal dynamics in the data.

The building block of SNNs is the spiking neuron, which accumulates input signals over time and emits a spike when a certain threshold is reached. Synaptic connections between neurons are characterized by weights that determine the influence of the activity of one neuron on another. When a presynaptic neurons emits a spike, this impulse (e.g., normalized to 1) is multiplied by the synaptic weight connecting it to the postsynaptic neuron. The weighted input is then integrated over time by the postsynaptic neuron. When its membrane potential reaches a certain threshold (e.g., also 1), it emits a spike, in turn, and resets its membrane potential (e.g., to 0). This spike is then transmitted to other neurons through their corresponding synaptic weights, influencing their membrane potentials, and so on.

## 4.1. Neural Coding

SNN take deeper inspiration from biological neurons than MLPs. One important aspect of neuronal communication is the use of spikes. Another key feature of biological neurons is sparsity. Neurons spend the majority of their time at rest; only a small subset of neurons are active at any given moment. A related concept is "static suppression" (or "event-driven processing"). For example, in the sensory systems, neurons are more responsive to changes in input than to static, unchanging signals. This responsiveness to dynamic information allows the brain to prioritize and focus on relevant stimuli while filtering out constant, background input (Eshraghian et al., 2023). These principles lead to the way in which signals are encoded and processed in SNNs.

### *4.1.1. Rate Coding*

One encoding mechanism is *rate coding*, where an input stimulus $x$ is converted into a firing rate or spike count. The higher the intensity of the input, the more spikes are generated. In this approach, the intensity of a sensory input corresponds to the firing rate of neurons. For example, brighter light results in a higher firing rate, while darker input leads to a lower rate.

The set of spikes generated by a neuron in a certain time interval is called a *spike train*. In rate coding, since the firing rate of the neuron is known (proportional to $x$) and constant, the resulting spike train follows a Poisson distribution, which is used to describe the number of events that occur in a fixed interval of time.

Let $k$ be the total number of spikes in a time interval and $T$ the length of the time interval. Then the mean firing rate $r = k / T$. There are two commonly used methods for generating Poisson spike trains (Heeger, 2000). The first is based on the idea that the probability of a spike occurring in a short subinterval $\Delta t$ is approximately equal to $r \cdot \Delta t$. Thus, the total time interval is divided into a set of short intervals (bins), and a uniform random number $u_i \in [0,1]$ is generated for each bin. If $u_i \leq r \cdot \Delta t$, we consider that a spike occurs in bin $i$.

However, a disadvantage of this method is the time discretization. An alternative is to compute the intervals between spikes using the exponential distribution characteristic to Poisson processes, whose probability density function is:

$$f(t) = r \cdot e^{-r \cdot t} \qquad (4.1)$$

In practice, one can generate uniform random numbers $u_i \in [0,1]$ and transform them into the desired exponential distribution using the formula:

$$e_i = -(1/r) \cdot \ln(1 - u_i) \qquad (4.2)$$

Unlike the discrete spikes characteristic of SNNs, the firing rate is a real-valued number, which can be compared to the output of the activation function in an MLP neuron. For a single neuron, the two are not equivalent, because the spike trains in SNNs contain temporal information that is not present in the instantaneous activations of MLP neurons. However, when seen in a statistical context, for example, by analyzing populations of SNN neurons, the activation of an MLP neuron can be seen as representing the average firing rate of a group of spiking neurons.

*4.1.2. Temporal Coding*

Another neural coding approach is *temporal coding*. Here, the intensity of the input stimulus determines *when* a spike occurs. For example, the *time to first spike* mechanism encodes a larger stimulus as an earlier spike.

Neurons within the sensory systems exhibit a considerable capacity to respond to stimuli across a wide dynamic range. Therefore, a logarithmic relationship between spike times and input feature intensity is often used. We will also encounter this logarithmic relationship in Section 11.1 dedicated to the representation of numerosities (number-related quantities) in the brain.

Temporal coding serves as a neural strategy that mitigates some shortcomings of rate coding. It enables the brain to rapidly and precisely decode complex sensory input, which is essential in scenarios that require fast reaction times. There are cases in sensory processing, for example, where rate coding would be too slow to provide the necessary speed observed in human subjects. Such cases include: sound localization, where brief stimuli demand rapid responses, visual encoding, especially edge detection in high-contrast situations, and discrimination of similar stimuli in gustatory and olfactory systems. In addition, the energy consumption of the brain is estimated to be lower than what would be necessary if only rate coding were used.

While temporal coding schemes are more complex than rate coding, we can mention, e.g., in a simplified form, the idea proposed by (Park et al., 2020) in a model named T2FSNN (time to first spike coding for deep SNNs), with an encoding and a decoding phase.

The encoding phase converts the dynamic input (or membrane potential) $V_i$ into a spike time $s_i$:

$$s_i = \lceil -\tau \cdot \ln(V_i(r-1)) + \delta \rceil \tag{4.3}$$

where $i$ is the index of a neuron, $\tau$ is a time constant, $r$ is a reference time defined as the start time of the following decoding (or fire) phase, and $\delta$ is a time delay. The model assumes a certain number of layers and the signals propagate from layer to layer. $\tau$ and $\delta$ are trainable parameters for each layer, but for simplicity, they can have the same values for all layers.

In the decoding phase, the postsynaptic neurons integrate the information encoded as single spikes in the previous phase:

$$z_j = \sum_i w_{ij} \cdot e^{-(s_i - r - \delta)/\tau}, \tag{4.4}$$

where **w** are the synaptic weights. The neurons are IF neurons, and thus they integrate each spike at the corresponding time, in succession, and beside the weights, earlier spikes have a greater impact due to the exponential transformation.

Temporal coding can be used for classification tasks (Comșa et al., 2020), based on the class whose corresponding output neuron spikes first. In a biological context, the winning neuron can suppress neighboring neurons through lateral inhibition. In machine learning, the spike times of inhibited neurons can suggest alternative predictions, in the order of subsequent spikes.

The related literature also includes the method of *reverse coding* (Zhang et al., 2019), where the stronger the input stimulus is, the later the corresponding neuron fires a spike.

*4.1.3. Burst Coding*

Rate coding and temporal coding are two extremes of the neural coding process. However, there are other types of coding as well. For example, in *burst coding* (Park et al., 2019), considering a normalized stimulus $x \in [0,1]$, the number of spikes a neuron produces in a time interval *T* is:

$$n_s = \lceil n_{max} \cdot x \rceil \tag{4.5}$$

where $n_{max}$ is the maximum number of spikes, which is usually no more than 5 (Buzsáki, 2012). These spikes are then equally distributed in the specified interval with an inter-spike interval (Guo et al, 2021):

$$ISI(x) = \begin{cases} \lceil T \cdot (1-x) \rceil & \text{if } n_s > 1 \\ T & \text{otherwise} \end{cases} \tag{4.6}$$

Therefore, the greater the *x* stimulus, the higher the number of spikes, but unlike in rate coding, they are equally distributed. Bursting also results in spike trains that do not follow a Poisson distribution. The advantage of burst coding is that it is more reliable in the presence of noise than the single-spike temporal encoding.

*4.1.4. Comparison*

Rate coding and temporal coding offer distinct advantages. Rate coding provides error tolerance through the presence of multiple spikes over time. This redundancy reduces the impact of missed events. Furthermore, rate coding generates more spikes, which enhances the gradient signal for learning when gradient-based algorithms are used. On the other hand, temporal coding leads to better power efficiency by generating fewer spikes. This reduces dynamic power dissipation in specialized hardware and minimizes memory access frequency due to sparsity. Temporal coding is particularly effective in scenarios requiring quick responses, as a single spike efficiently represents information, making it ideal for tasks with time constraints. This aligns with observations in biology, where the brain optimizes for efficiency, supported by low average firing rates in neurons.

**4.2. The Computational Power of SNNs**

Maass (1997) showed that SNNs are computationally more powerful, in terms of the number of neurons needed for some tasks, than MLPs that use a sigmoid activation function. He also demonstrated that SNNs and MLPs are equivalent as universal function approximators.

In order to show the (theoretical) superiority of SNNs, he considers an "element distinctness" function $ED : (\mathbb{R}^+)^n \to \{0, 1\}$ defined as:

$$ED(x_1,...,x_n) = \begin{cases} 1 & \text{if } x_i = x_j \text{ for some } i \neq j \\ 0 & \text{if } |x_i - x_j| \geq 1 \text{ for all } i, j \text{ with } i \neq j \\ \text{arbitrary} & \text{otherwise} \end{cases} \tag{4.7}$$

He then proposes a less intuitive variant of *ED* defined as:

- 1, if there exists $k \geq 1$ such that $x_1, x_2, x_3, x_{3k+1}, x_{3k+2}, x_{3k+3}$ have the same value;
- 0, if every interval of length 1 in $\mathbb{R}^+$ contains the values of at most 3 inputs $x_i$;
- arbitrary, otherwise.

The author proves that this function can be computed by a single spiking neuron using temporal coding, because it needs to fire only when two blocks of three adjacent synapses receive synchronous excitatory potentials. He also discovers a lower bound for the number of hidden units needed by a sigmoidal MLP, which can be over 1000 for $n = 10000$.

This is definitely a great difference and it is interesting that it is found for a version of the general problem of detecting similarity between inputs. From the same category we have the *xor* problem that cannot be learned by a single layer perceptron.

Taking the generality of SNNs further, in (Vineyard et al., 2018) the implementation of several fundamental algorithms using temporal coding is described, e.g., the computation of the minimum, maximum and median values, sorting, nearest neighbor classification, and even a more complex neural model, adaptive resonance theory, which will be described later in Section 15.1.

**4.3. Discussion**

SNNs and MLPs exhibit differences and similarities in terms of their structure, behavior, and applications.

One key distinction is that SNNs explicitly model the timing of neuron spikes, mimicking the temporal dynamics of real neurons, while MLPs process input data as static snapshots without explicit time representation. SNNs can be trained using more biologically plausible learning rules (presented in Section 5), which can adjust the synaptic weights based on the precise timing of spikes, whereas MLPs primarily use gradient descent algorithms such as backpropagation and its modern variants for weight updates. Moreover, SNNs tend to be more energy efficient due to their event-driven processing nature, which only activates neurons when necessary, while MLPs often require continuous computations and thus consume more resources.

However, they also share similarities. Both SNNs and MLPs consist of layers of interconnected neurons and are suitable for tasks such as classification, regression, and pattern recognition. In SNNs, neurons produce output spikes based on the accumulation of membrane potentials and firing thresholds, mirroring MLPs, where neurons compute activations using weighted sums of inputs followed by an activation function. Both models rely on the idea of input accumulation and they both include the idea of a nonlinear transformation of this net input.

SNNs have various strengths and weaknesses that impact their applicability and performance in different tasks.

Discussing their strengths, they exhibit a higher level of biological plausibility compared to traditional neural networks such as MLPs. SNNs are good at processing temporal information and sequences, making them suitable for tasks involving speech recognition, event prediction, and time-series analysis. Their innate capacity to encode timing through spike patterns provides a significant advantage in scenarios where the order and timing of events are critical. SNNs also demonstrate robustness to input noise, relying on patterns of spikes over time rather than exact continuous values.

However, SNNs also exhibit several noteworthy weaknesses that need to be taken into account. First, their complexity is a prominent challenge. Developing and training SNNs is inherently more complex compared to traditional NNs. Scalability may be an issue in large-scale networks due to hardware constraints. The learning and training process for SNNs, particularly when employing biologically plausible learning rules, can be slower and more difficult than the variants of backpropagation used in MLPs. Moreover, SNNs often rely on sparse representations, which may lead to overfitting or underutilization of available resources in certain applications. Last, the interpretability and expressiveness of SNNs pose challenges. The complex relationship between spike patterns and network behavior can make it more difficult to interpret and understand the internal representations and decisions of SNNs.

Although spiking neurons lie at the basis of realistic brain modeling, working directly with them may be too low level if one's goal is to address cognitive capabilities. It may be like working in assembler to program intelligent algorithms. That is why, in Section 7 we will describe models that operate on the level of groups of neurons, called "cell assemblies".

## 5. Learning Rules

In this section, we will present several learning rules for neural networks, which are more biologically plausible than the classic backpropagation algorithm and its recent extensions. Backpropagation is not considered to be biologically plausible because error propagation occurs through a different mechanism than neuronal activation. In an equivalent biological setting, a global error signal that provides information about the actual performance of the network compared to the desired output would need to be propagated from the dendrite of the postsynaptic neuron to the axon of the presynaptic neuron. In reality, such an explicit global error feedback that is propagated backwards through the network to adjust the weights does not exists, although the idea of change following an error is indeed present. Moreover, the actual quantities encoded by the gradients have no biological counterpart.

In general, learning in the brain relies on more localized learning rules, where neurons adjust their connections based on the activity patterns of their neighbors.

The simple delta rule used in a single layer perceptron can be seen as biologically plausible because the change of weights depends on the local error. It is the propagation of gradients through the hidden layers of an MLP that is less plausible.

Biological synapses also exhibit various forms of plasticity that include the timing of neuronal spikes, which are more complex and often involve multiple factors beyond simple gradient-based adjustments.

**5.1. Hebbian Learning**

*Hebbian learning* describes a basic learning principle based on synaptic plasticity, i.e., the ability of the strength of connections (synapses) between neurons to change in response to neural activity. Named after Donald Hebb, it is often summarized by the famous phrase: "cells (neurons) that fire together, wire together".

Hebb's rule was actually stated as follows: "When an axon of cell *A* is near enough to excite cell *B* and repeatedly or persistently takes part in firing it, some growth process or metabolic change takes place in one or both cells such that *A*'s efficiency, as one of the cells firing *B*, is increased" (Hebb, 1949). In simpler terms, if two neurons are frequently active at the same time, the synapse between them strengthens. If neuron *A* consistently fires just before neuron *B*, the synapse from *A* to *B* becomes stronger, making it more likely for *A*'s activity to trigger *B*'s activity in the future.

This principle underlies the idea that learning and memory formation in the brain are a result of strengthening the connections between neurons that are frequently activated together.

From the computational point of view, this learning rule is usually unsupervised, involving networks with the structure shown in Figure 5.1. We present here its application for neurons with static behavior, such as those in MLPs, not spiking neurons. These neurons are typically simple, with linear activation functions.

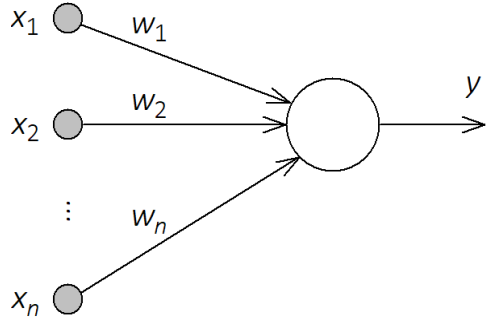

**Figure 5.1.** A simple network to illustrate Hebbian learning

Assuming only one output *y*, its value is computed as follows:

$$y = \sum_{i=1}^{n} x_i \cdot w_i \quad (5.1)$$

For each connection, after the presentation of a training instance, the weight update according to Hebb's rule is:

$$\Delta w_i = \alpha \cdot x_i \cdot y \quad (5.2)$$

where *α* is the *learning rate*. Of course, this equation is applied for all the instances in the training set.

Although simple, such a model is capable of discovering correlations in its input data. Assuming $\alpha = 1/m$, where *m* is the number of training instances, we can express $\Delta w_i$ after a training epoch in terms of an average function over the whole training set using equations (5.1, 5.2):

$$\Delta w_i = \frac{1}{m} \cdot \sum_{s=1}^{m} x_{si} \cdot \sum_{j=1}^{n} x_{sj} \cdot w_j = \sum_{j=1}^{n} \langle x_i \cdot x_j \rangle \cdot w_j \quad (5.3)$$

Therefore, the weight change is a function of the correlations between the inputs. The correlation matrix **C** is defined as:

$$\mathbf{C}_{ij} = \langle x_i \cdot x_j \rangle \quad (5.4)$$

Using the equivalent expression for the weight update:

$$\Delta w_i = \sum_{j=1}^{n} C_{ij} \cdot w_j \qquad (5.5)$$

the weight vector after epoch *n* can be expressed as (Gerstner et al., 2014):

$$\mathbf{w}(n) = (\mathbf{1} + \mathbf{C}) \cdot \mathbf{w}(n-1) = (\mathbf{1} + \mathbf{C})^n \cdot \mathbf{w}(0) \qquad (5.6)$$

where **1** is the identity matrix.

An interesting point is that the weight vector can also be expressed using the eigenvectors $\mathbf{e}_k$ and the eigenvalues $\lambda_k$ of **C**:

$$\mathbf{w}(n) = \sum_k \mathbf{a}_k(n) \cdot \mathbf{e}_k = \sum_k (1 + \lambda_k)^n \cdot \mathbf{a}_k(0) \cdot \mathbf{e}_k \qquad (5.7)$$

Since the correlation matrix is positive semi-definite, $\lambda_k \in \mathbb{R}^+$, and thus the weight vector is growing exponentially, but the growth is dominated by the eigenvector with the largest eigenvalue, i.e., the first principal component:

$$\lim_{n \to \infty} \mathbf{w}(n) = (1 + \lambda_1)^n \cdot \mathbf{a}_1(0) \cdot \mathbf{e}_1 \qquad (5.8)$$

The output *y*, according to equation (5.1), can be interpreted as the projection of the input **x** on the direction **w**. Since **w** is proportional to $\mathbf{e}_1$, *y* is also proportional to the projection on the first principal component. Therefore, because the neuron can eventually project the inputs on $\mathbf{e}_1$, a neuron using Hebb's rule can extract the *first principal component* of the input data.

It is quite interesting that such a simple model can identify the most relevant direction of variation in the input data. However, the input data should have a zero mean, but this can be easily achieved by extracting the average values from the actual data.

The main drawback of this method is that the weight vector increases exponentially. This issue is addressed by the extensions presented next.

## 5.2. Extensions of Hebb's Rule

### 5.2.1. Oja's Rule

To avoid the weights becoming infinitely large, Erkki Oja proposed a solution called *subtractive normalization* (Oja, 1982). It incorporates a competitive or homeostatic term that limits the growth of synaptic weights. This term helps to ensure that the synaptic weights converge to a stable range over time. The corresponding weight update equation is:

$$\Delta w_i = \alpha \cdot x_i \cdot y - y^2 \cdot w_{ij} \qquad (5.9)$$

This rule is a form of weight normalization, i.e., when one element of the weight vector increases, the other elements automatically decrease. In this case, the network may forget old associations: if a training instance is not presented frequently, it may be forgotten.

Because it uses the same correlation term as Hebb's rule, Oja's rule also computes the first principal component of the input data (Hertz, Krogh & Palmer, 1991).

### 5.2.2. Sanger's Rule

This rule (Sanger, 1989) can be considered an extension of Oja's rule, in the sense that it can determine the largest (or all) principal components, which converge successively. Considering the network in Figure 5.2, it projects the $n$-dimensional input onto an $m$-dimensional representation, where $m \leq n$, but usually $m \ll n$. The weights represent $m$ vectors of size $n$. The outputs of the network are computed linearly:

$$y_j = \sum_{i=1}^{n} x_i \cdot w_{ij} \tag{5.10}$$

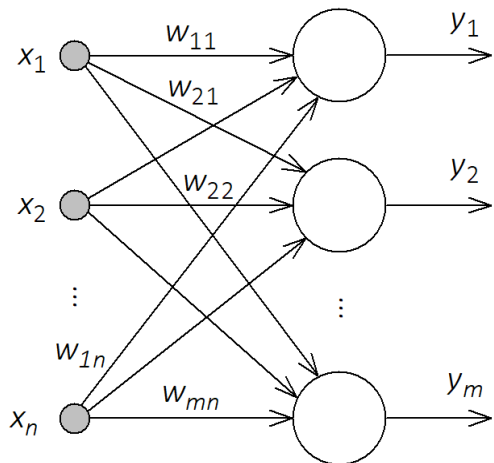

**Figure 5.2.** A simple network with multiple outputs

The learning rule is:

$$\Delta w_{ij} = \alpha \cdot x_i \cdot y_j - y_j \cdot \sum_{k=1}^{j} y_k \cdot w_{ik} \tag{5.11}$$

The first output neuron is equivalent to the Oja model which learns the first principal component. Because of the subtraction involving the sum of 1 to $j$, the second neuron is forced to find another principal component, and so on. The consequence is that the weights converge to an ordered set of $m$ principal components (i.e., the eigenvectors of the correlation matrix **C**).

This highlights the power of these kind of simple models to extract important information about the input data distribution using only Hebbian associations.

## 5.2.3. The ABCD Rule

This is a generalized version of Hebb's rule, which takes into account not only the product between the input and the output, but also the input and the output values separately, together with a bias term:

$$\Delta w_{ij} = \alpha \cdot \left( A \cdot x_i \cdot y_j + B \cdot x_i + C \cdot y_j + D \right) \quad (5.12)$$

In this equation, *A* (the correlation term), *B* (the presynaptic term), *C* (the postsynaptic term) and *D* (the bias or a constant) are parameters whose optimal values can be found, e.g., using an evolutionary algorithm. This rule was used for reinforcement learning tasks where the evolved Hebbian rules were sufficient for an agent to learn to navigate a 2D environment and to learn to walk in a 3D environment (Niv et al., 2001; Niv et al., 2002).

## 5.2.4. BCM Rule

Another solution for the instability of the Hebb's rule is suggested by the BCM rule (Bienenstock, Cooper & Munro, 1982), which uses a sliding threshold:

$$\Delta w_i = \alpha \cdot x_i \cdot y \cdot (y - \theta) \quad (5.13)$$

In this equation, $\theta$ is the average value of $y^2$ for several past time steps or presented training instances, and it controls the strengthening or weakening of the synapses. The BCM rule avoids the infinite increase of weights by allowing the threshold to vary in such a way that $\theta$ can grow faster than *y*. Its effect can be seen in Figure 5.3.

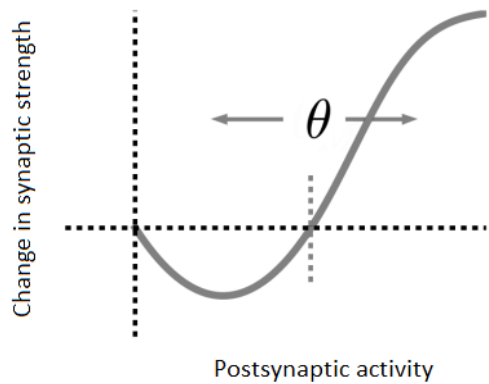

**Figure 5.3.** The effect of the threshold when the presynaptic activity is constant

The shape of the function in the figure is similar to the synaptic changes observed, e.g., in the visual cortex (Intrator & Gold, 1993).

## 5.3. Spike-Timing Dependent Plasticity

In this section, we will take into account the relative time of spiking of the presynaptic and postsynaptic neurons. In biological systems, neurons are unlikely to fire exactly at the same time.

Moreover, a synapse is strengthened only when the activation of the presynaptic neuron leads to the activation of the postsynaptic neuron, i.e., there is a causal link between the two activations.

The *spike-timing dependent plasticity* (STDP) rule (Markram et al., 1997; Bi & Poo, 1998) posits that the timing difference between presynaptic and postsynaptic spikes determines whether a synapse strengthens or weakens. When a presynaptic neuron fires just before a postsynaptic neuron, the synapse is strengthened, and further enhances the signal transmission. Conversely, if it is the postsynaptic neuron that fires before the presynaptic neuron, the synapse weakens, and thus reduces signal transmission.

This behaviors correspond to two important processes related to synaptic plasticity: *long-term potentiation* (LTP), meaning a persistent strengthening of synapses, and *long-term depression* (LTD), meaning a persistent weakening of synapses. According to STDP, repeated presynaptic spike arrival a few milliseconds prior to postsynaptic spikes leads to the LTP of the synapse, while repeated presynaptic spike arrival after postsynaptic spikes leads to the LTD of the synapse.

To formalize these ideas, let $\Delta t$ be the difference between the spiking time of the postsynaptic neuron $j$ and the spiking time of the presynaptic neuron $i$:

$$\Delta t = t_j - t_i \tag{5.14}$$

Then:

$$\Delta w_{ij} = \begin{cases} \alpha_p \cdot e^{-\Delta t/\tau_p} & \text{if } \Delta t > 0 \\ -\alpha_n \cdot e^{\Delta t/\tau_n} & \text{if } \Delta t < 0 \end{cases} \tag{5.15}$$

where $\alpha_p$ (with $p$ from "positive") and $\alpha_n$ (with $n$ from "negative") control the maximum and minimum values of the weight change, and $\tau_p$ and $\tau_n$ are time constants that control the shape of the function.

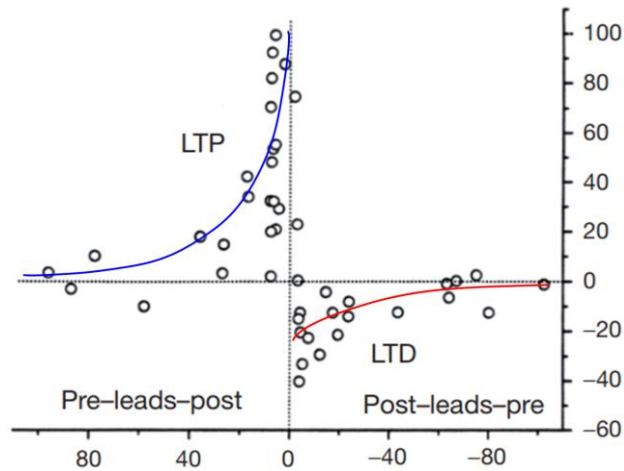

**Figure 5.4.** Experimental evidence of STDP

The *X* axis shows the difference between presynaptic and postsynaptic spike timing in milliseconds. For example, in the left half the presynaptic spike occurs before the postsynaptic spike. The *Y* axis shows the change of the excitatory postsynaptic potential amplitude in percentages. Each small circle refers to a neuron presented with a certain spiking delay. Adapted after (Bi & Poo, 1998) and (Shulz & Feldman, 2013)

Figure 5.4 shows the effect of spiking time difference on the change of the postsynaptic potential amplitude, following the experimental data obtained by Bi and Poo (1998) using neurons in a hippocampal cell culture. In the brain, several variants of STDP were actually observed (e.g., Hebbian or anti-Hebbian), depending on the types of the neurons involved in the pair that communicates through the synapse (e.g., excitatory or inhibitory) (Shulz & Feldman, 2013).

One can see that the maximum LTP and LTD effects are obtained when the two neurons are activated within a short interval. If the presynaptic neuron fires just before the postsynaptic neuron, the synapse strengthening is maximal.

In a way, Hebb's rule can be seen as an approximation of STDP, considering that the neurons fire at the same time and the presynaptic neuron has a direct effect on the activation of the postsynaptic neuron. Several variants of anti-Hebbian equations for rate-based (non-spiking) neurons were proposed to account for the LTD phenomenon. For example, the synapse is weakened when the presynaptic neuron is active and the postsynaptic neuron is not.

From the biological point of view, the presynaptic neural activity involves the release of the excitatory neurotransmitter glutamate. The corresponding postsynaptic activity is related to a depolarized membrane potential. These two conditions open the *n-methyl-d-aspartate* (NMDA) channel that allows calcium ions ($Ca^{2+}$) to enter the synapse. This mechanism occurs at dendrite level. In the end, after several complex chemical reactions, the synaptic efficacy changes. A high concentration of calcium leads to LTP while a lower concentration leads to LTD (O'Reilly & Munakata, 2000). We will revisit the NMDA-related processes in Section 10.

However, the strengthening of synapses is also influenced by the neurotransmitter dopamine. Dopamine-producing neurons are activated when people expect to receive a reward. It is not the reward itself, but the expectation of a reward that most strongly influences emotional reactions and memories. The process of reward learning takes place when individuals encounter unexpected outcomes. If a reward exceeds their expectations, dopamine signaling increases. Conversely, if a reward is less than anticipated, dopamine signaling decreases. But correctly predicting a reward does not significantly alter dopamine signaling because there is no new learning involved (Halber, 2018).

## 5.4. Other Learning Algorithms

In general, many algorithms were proposed for training SNNs. Some use variants of STDP while others draw inspiration from backpropagation to design algorithms for spiking neurons, e.g., using surrogate gradients, where the discontinuous impulse functions are approximated with differentiable functions. A review of such methods is given in (Han & Lee, 2021).

One algorithm that attempts to implement the idea of backpropagation in a biologically plausible manner is *GeneRec* (Generalized Recirculation) (O'Reilly, 1998), also used in the LEABRA (local error-driven and associative, biologically realistic algorithm) cognitive architecture, introduced in (O'Reilly, 1996). It is inspired by the recirculation algorithm (Hinton & McClelland, 1987) and the recurrent backpropagation algorithm (Almeida, 1987; Pineda, 1987).

GeneRec assumes bidirectional connections: from the inputs to the outputs and vice versa (which can also be viewed as bottom-up and top-down). Learning occurs in two phases. In the *plus phase*, the desired value is "clamped" on the output layer. Thus, both the input values and the desired values are presented to the network. Since the network is recurrent, the activations of the neurons in the hidden layer depend on both. In the *minus phase*, only the inputs are presented. Following several assumptions and approximations, the suggested learning rate is:

$$\Delta w_{ij} = \alpha \cdot x_i^- \cdot \left( y_j^+ - y_j^- \right) \tag{5.16}$$

where $x_i$ is the activation of the presynaptic neuron and $y_j$ is the activation of the postsynaptic neuron.

As one can notice, equation (5.16) has the form of the delta rule. The difference between the two phases of activation states is an indication of the contribution of a unit to the overall error signal. The bidirectional connectivity propagates both parts of this signal throughout the network, therefore learning can be based on the difference. Moreover, the activation states are local to the synapse where the weight changes must occur.

# 6. Neuronal Ensembles (Cell Assemblies)[1]

Donald Hebb introduced the concept of *cell assemblies* (CA), also named *neuronal ensembles*, to describe small groups of sparsely distributed neurons that become active in response to relevant stimuli. They represent groups of interconnected neurons that work together to perform specific functions within the brain, spanning from simple sensory processing to complex cognitive tasks. The ensembles serve as functional units that process information and can be considered to be the building blocks of neural information processing in the brain.

The *cell assembly hypothesis* (Hebb, 1949) posits that a CA serves as the neural representation of a concept. This hypothesis has gained support from biological, theoretical, and simulation data over time. CAs can classify sensory stimuli, allowing individuals to identify objects, but they can also be activated without direct sensory input, such as when thinking about a concept.

Since they consist of groups of neurons, they are an intermediate structure larger than individual neurons but smaller than the entire brain. Thus, they can provide an organizing principle for studying the mind and brain, and may bridge the gap between psychology and neurobiology.

CAs represent concepts using sets of neurons with elevated firing rates (*population coding*). They consist of a relatively small set of neurons encoding each concept, with estimated sizes ranging from $10^3$ to $10^7$ neurons per CA. While the brain is generally composed of neurons that constantly fire at a low rate (e.g., 1 Hz), CA neurons fire at a high rate (e.g., 100 Hz).

Neurons within a cell assembly exhibit self-sustaining persistent activity, known as *reverberation*, after the initial stimulation. They have strong mutual synaptic connections and when a sufficient subset is ignited, e.g., by mentioning a specific concept, they initiate a neural cascade that persists after the initial stimulus presentation ends. This sustained activation characterizes a short-term memory (STM), which can last for several seconds.

The set of neurons within a CA is formed through processes such as synaptic weight changes, neural growth, neural death, and synaptic growth. The high mutual synaptic strengths arise as a consequence of Hebbian learning. These neurons have previously fired together in response to earlier stimuli, resulting in the strengthening of synaptic connections between them. Consequently, CAs typically form through the repetitive presentation of similar stimuli, and therefore they can also be considered to represent a long-term memory (LTM).

---

[1] This section uses information and ideas from: (Huyck, 2007), (Huyck & Passmore, 2013), (Carrillo-Reid, 2022), and (Tilley, Miller & Freedman, 2023).

Correlated neural firing, or *synchrony*, is observed when neurons that respond to the same stimulus or cause the same action fire together. Synchrony is considered a signature of CAs working together coherently. Studies showed that synchronized neural firing can be a mechanism for binding perceptual elements and for coordinating activity across brain areas.

CAs can overlap, with individual neurons participating in multiple CAs. Some neurons may be more central or critical to a CA than others. Also, CAs are active entities, influencing other assemblies through excitation or inhibition. They can recruit new neurons, lose neurons, or split into separate CAs. In addition, cell assemblies are dynamic and flexible, capable of reconfiguring themselves based on experience. This plasticity allows the brain to adapt or learn in a changing environment.

Associations between individual CAs can support symbol grounding (discussed in Section 17.4), the process by which symbols gain meaning. Learned CAs can be derived directly from environmental stimuli.

While there is extensive evidence of cell assemblies persistence, linking this evidence with simulations and psychological theories predicting memory strength and duration remains a challenge. Existing CA models either persist indefinitely or cease after a relatively short time, lacking a robust link between memory strength and neural firing. Fast CA formation is also an issue, especially when distinguishing between semantic and episodic memories, where semantic memories may require repeated presentations for CA formation. The dynamics of CA formation can be complex, potentially involving phenomena such as neural avalanches that obey power law dynamics in neural connections.

Each neuronal ensemble is specialized for a particular function or task. For example, in the visual cortex, there are ensembles dedicated to recognizing faces, while others are responsible for detecting motion. In the hippocampus, neuronal ensembles are responsible for encoding and retrieving memories. When recalling past events or learning new information, specific ensembles become active and synchronize their activity patterns. Neuronal ensembles are involved in sensory processing as well, e.g., in the visual cortex, different ensembles respond to specific visual features like edges, colors, or motion. The collective firing patterns of neurons in these ensembles allow the perception and interpretation of complex visual scenes.

Traditional memory research focused on identifying where memories are stored in terms of which cells or brain regions represent memory traces. These physical manifestations of memories are called "memory engrams". Recent experiments suggest that the interaction and reactivation of specific neuronal groups, rather than the localization of memories in specific regions, may be the primary mechanism underlying the brain's ability to remember past experiences and imagine future actions. The interaction between neuronal ensembles is essential for forming sequential activity patterns into memories and for composing complex behaviors through the reactivation of memories.

When classic artificial neural networks are trained on a new task after learning an initial one, their performance on the first task may often drop significantly. This issue is known as *catastrophic forgetting* and has been a persistent challenge in the field of NNs. Continual learning refers to the ability of an NN to learn multiple tasks sequentially without forgetting the knowledge acquired from previous tasks. Unlike artificial neural networks, the human brain can sequentially learn and remember multiple tasks without experiencing catastrophic forgetting. The mechanisms responsible for this capability are not fully understood, but are actively investigated in neuroscience.

One hypothesis suggests that the brain restricts the plasticity of synaptic connections that are important for a memory once it has been formed. New connections are made for new tasks, while connections associated with prior tasks undergo a reduction in plasticity to preserve the acquired knowledge.

Another proposed mechanism is that the brain encodes memories in distinct neuronal ensembles, where different sets of neurons are used to encode memories in different contexts. This approach reduces interference between memories and prevents catastrophic forgetting.

## 7. Cell Assembly Models

In this section, we will present the contributions of several papers that use the concept of cell assemblies for either computational tasks, or neuroscience research.

Papadimitriou et al. (2020) introduced a computational model called *assembly calculus*, aiming to bridge the gap between the level of spiking neurons and that of the whole brain. They identified a set of operations on CAs, demonstrating their feasibility through both analytical methods and simulations.

The model focuses on assemblies of excitatory neurons, which may underlie higher human cognitive functions such as reasoning and planning, as well as syntactic language processing consistent with experimental results. It defines several high-level processes that can be translated into neuron operations, much like high-level programming languages are compiled into machine code.

Assembly calculus adopts a simplified but realistic representation of neurons and synapses. Several areas are used, symbolizing distinct parts of the cortex. Each area has a population of excitatory neurons connected by random synapses, forming recurrent loops. There are also random connections between different areas.

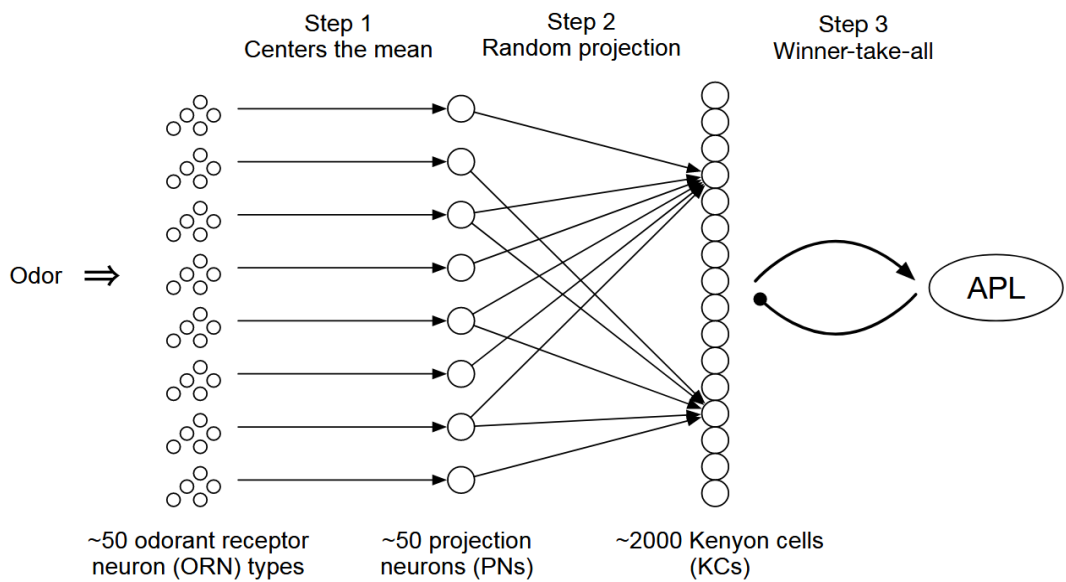

**Figure 7.1.** Smell recognition by the fruit fly
APL (anterior paired lateral) are neurons that provides feedback inhibition to Kenyon cells
(Dasgupta, Stevens & Navlakha, 2017)

It employs Hebbian learning while also incorporating a slower time-scale renormalization of synaptic weights. This adjustment aims to maintain a relatively stable sum of presynaptic weights for each neuron, resembling a homeostatic process (or forgetting). To circumvent the need for the explicit modeling of inhibition, a fixed number of neurons in any given area fire at any given time, i.e., those with the highest levels of activation. This is a form of *k winners take all* (kWTA) function. While kWTA is biologically implausible due to its requirement for global information about activation states and sorting mechanisms, it offers a computationally effective approximation of inhibitory dynamics. In this paper, this concept is referred to as *capping*. It is also used, e.g., in the LEABRA cognitive architecture (O'Reilly, 1996).

An example of kWTA is how the fruit fly (Drosophila melanogaster) recognizes smells, shown in Figure 7.1. It has about 50 receptors for odorants (smell molecules). The sensory neurons are connected one-to-one to 50 other neurons, which in turn randomly connect to 2000 neurons (Kenyon cells). At this stage, because of inhibition, only about 100 Kenyon cells are activated and express the recognition of the smell. This process can be modeled as kWTA with 100 winners.

What is interesting about the random projection and cap operation in assembly calculus is that a projection can be seen as a lower dimensional representation of the source assembly. For example, if two different inputs create overlapping representations in area *A*, the corresponding projections in area *B* will also overlap. This can lead, e.g., to the possibility of an approximate nearest neighbor search in a lower dimensional space in area *B* instead of the original space in area *A*.

The assembly calculus computational model is generic, not relying on pre-existing neural circuits designed for particular tasks. Functionality, including assembly operations, emerges from the intrinsic randomness within the network and the selection of neurons with high input values. This functionality is a consequence of simple events such as the repeated firing of an assembly.

The main operations are presented as follows.

In the association cortex, it is assumed that assemblies form through the *projection* of a neuronal population from one area to another. For example, let us consider two areas *A* and *B*. The *k* most excited neurons in *A* fire in *B*. The *k* most excited neurons in *B* fire to other neurons in *B*. The repeated firing from *A* to *B* in combination with the recursive firing from *B* back to itself increases the overlap of firing neurons in *B* in consecutive rounds. So projection replicates an assembly in a downstream area, ensuring that the newly formed assembly activates whenever the original one does. Assemblies are densely intraconnected, signifying that neurons within the same assembly are considerably more likely to be connected compared to those located in different assemblies. Despite the computational complexity of finding dense subgraphs in sparse graphs, the existence of assemblies can be explained by the creation of assemblies through projection, leading to increased synaptic connectivity.

*Reciprocal projection* represents an elaborated form of projection. Much like in the case of standard projection, when assembly *x* in area *A* fires, it triggers a set of neurons, denoted as $y_1$, in area *B* to become active. The main difference is the presence of synaptic connections from area *B* back to area *A*, in addition to the existing connectivity from *A* to *B*. In the next step, assembly *x* slightly shifts, becoming $x_1$, while $y_1$ becomes $y_2$. This process continues iteratively until convergence is achieved. The resulting assembly *y* exhibits strong synaptic connectivity both to and from assembly *x*. This contrasts with the regular projection, where synaptic connectivity is unidirectional, going only from *x* to *y*.

*Pattern completion* occurs when the firing of a small subset of cells within an assembly triggers the entire assembly to fire with a high probability. Because of the recurrent connections, a small part of an assembly can trigger the activation of the large majority of neurons in that assembly.

Through *association*, two existing CAs converge as a result of repeated simultaneous activation. This co-firing strengthens the connections between their neurons, enhancing the overlap and boosting the interconnectedness of the corresponding assemblies. This operation expresses the idea of an experiment using electrodes connected to the hippocampus of human subjects (Ison, Quian-Quiroga & Fried, 2015). They presented familiar place images and observed distinct neuron activations for each place. Similarly, images of famous people elicited unique neuronal firing patterns. When they simultaneously displayed pairs of person-place images, specific neurons responded, some associated with places, and some with individuals. Upon repeated exposure to these combined stimuli, previously distinct neuronal assemblies began to overlap. Neurons initially responsive to only one stimulus (a person) started firing alongside those associated with the other (the image of a place), highlighting how associative relationships between distinct stimuli can modify neural assembly configurations over time.

The *merge* operation is the most complex and sophisticated operation. It can be viewed as a double reciprocal projection. As assemblies $x$ and $y$ repeatedly fire in areas $B$ and $C$, respectively, a new assembly $z$ is eventually formed in area $A$, which also alters the original assemblies $x$ and $y$. In the resulting assemblies, strong bidirectional synaptic connections appear between $x$ and $z$, as well as between $y$ and $z$. The merge operation has an important role in creating hierarchies and is postulated to contribute to the syntactic processing of language and hierarchical thought processes. This was applied, e.g., to design a cortical architecture emphasizing syntactical aspects in language processing.

(Dabagia, Vempala & Papadimitriou, 2022) demonstrates that the assembly calculus framework can be used for classification. The paper proves that new assemblies can reliably form in response to inputs from different classes and remain distinguishable, particularly when these classes are reasonably well separated.

(D'Amore et al., 2022) empirically shows that complex planning strategies, such as heuristics for solving tasks in the blocks world, can be effectively implemented within assembly calculus, and suggests an approximation strategy for solving such problems. However, it also highlights the constraints in the representations required for planning that involve long chains of strongly connected assemblies; specifically, there are limitations on the reliable implementation of long chains. The idea is that a stack of blocks is modeled more or less like a linked list, connected to a head node. When the top block is removed, its association with the head node is slowly forgotten by weight renormalization (homeostasis) and a new association is learned between the block and its new neighboring node.

(Xie, Li & Rangamani, 2023) discusses the use of two assembly calculus operations, i.e., project for creating assemblies from stimuli and reciprocal-project for implementing variable binding, the process by which concrete objects (variables) are bound to the abstract structure determined by their relations, or in other words, how items that are encoded by distinct CAs can be combined to be used in various settings. It resembles the substitution of a variable with a fact in predicate logic (an important concept on which we will focus later in Section 13). The study shows that reciprocal-project has a lower capacity than project, but suggests a skip connection method that increases capacity and offers opportunities for exploring hierarchical models using neuronal assemblies.

(Huyck, 2007) shows that highly recurrent neural networks can learn reverberating circuits in the form of CAs. They have the ability to classify their inputs, and the paper explores their potential to learn hierarchical classes. A simulator based on spiking fatiguing leaky integrators is used for modeling, and learning is achieved using a form of Hebbian learning. Overlapping CAs allow neurons to participate in multiple CAs, enhancing the classification capabilities of the network. NNs containing CAs can learn and adapt based on sensory information, with CAs forming, growing, modifying, and splitting through unsupervised learning. Such networks act as classifiers, with CAs representing classes and being activated when an instance belongs to a class.

In the example in the paper, the network is divided into neurons that recognize 10 types of animal features, including those specific to an animal class (e.g., features 0 to 2 for dogs, cats and rats), and shared features for all mammals (features 3 and 4). Other features relate to tails and eating habits. Neurons associated with these features are activated based on the presence of specific features in an animal instance. The simulations demonstrate the emergence of hierarchical classification: the neural network recognizes four classes: *Dog*, *Cat*, *Rat*, and *Mammal* (the superclass). This is achieved by forming overlapping CAs. For example, *Dog* and *Rat* share features with *Cat*, while *Mammal* encompasses features common to most of its subclasses.

One advantage of such hierarchies is the ability to perform default reasoning. The trained network can deduce properties of a new class based on the features of its superclass. An example is given where the network, when presented with an instance of a *Bear*, uses default reasoning to infer that bears are omnivores based on the features of the *Mammal* superclass.

*SHRUTI* (Ajjanagadde & Shastri, 1991; Shastri, 1999) is a biologically plausible neural model of reflexive reasoning. It demonstrates that systematic inference with respect to a large body of general as well as specific knowledge can be the spontaneous and natural outcome of a neural system. Humans possess the innate ability to make a variety of inferences quickly and efficiently, resembling reflex responses. Cognitive science and computational neuroscience face the challenge of explaining how a network of neuron-like elements can represent extensive knowledge and perform rapid inferences. The SHRUTI model shows how a neural network can encode semantic and episodic facts, rules, and knowledge, enabling a wide range of rapid inferences.

Relational structures are represented by clusters of cells, and inference involves rhythmic activity propagation over these cell clusters with synchronous firing of appropriate cells.

Inference involving relational knowledge corresponds to the transient propagation of rhythmic activity across these focal clusters. SHRUTI offers a detailed computational account of how synchronous activity can be used to represent and process high-level conceptual knowledge and inference.

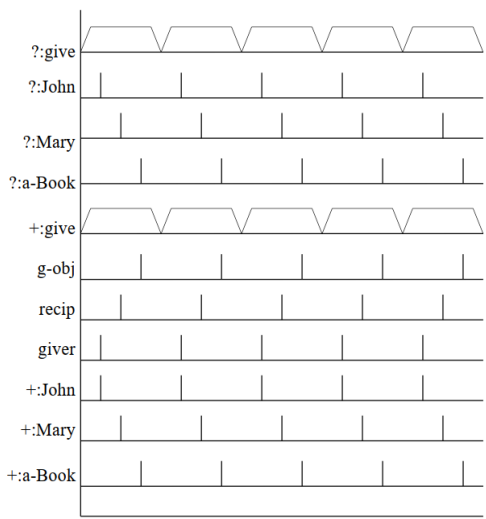

**Figure 7.2.** An example of synchronous neural firing in SHRUTI
(Shastri, 1999)

In SHRUTI, "understanding" corresponds to the coherent activity along closed loops of neural circuits. The model represents the binding between conceptual roles and the entities fulfilling those roles in a given situation through the synchronous firing of appropriate cells. Facts in LTM are encoded as temporal pattern-matching circuits that detect coincidences and coincidence failures in the ongoing flow of rhythmic activity. For example, the answer to the question "Does Mary own a book?" based on the knowledge base containing the assertion give(John, Mary, a-Book) causes a pattern of activations of the kind presented in Figure 7.2. The bindings are expressed by the synchronous activation of the role and the entity nodes. Each spike represents the synchronous firing of cells in the ensemble of the appropriate node.

One limitation of the SHRUTI model is that the concepts, although encoded in a neural fashion, are not grounded, but arbitrary, just like in classic symbolic representations.

(Tetzlaff et al., 2015) proposes that the combination of synaptic plasticity in the Hebbian sense and the slower process of synaptic scaling achieves two key goals: the formation of CAs and the enhancement of neural dynamics diversity. Synaptic scaling is a process that adjusts the strength of connections between neurons to maintain a balance in activity levels. It weakens synapses when activity is high and strengthens them when activity is low, ensuring stable neural function. The study introduces a mechanism based on the interaction between neuronal and synaptic processes on different time scales. It aims to enable the self-organized formation of CAs in the long term while performing nonlinear calculations in the short term. The main motivation is that the formation of CAs in the brain needs to be adaptive to learn and adapt to new information, but it should also be stable and not change rapidly. This suggests that synaptic plasticity, which enables learning, must be stabilized by processes that act on longer time scales, such as synaptic scaling. The analytical results suggest that the combination of synaptic plasticity and scaling is well-suited to achieve the coexistence of multiple CAs, because this is challenging to achieve with other synaptic plasticity rules or mechanisms such as inhibition or short-term plasticity.

(Buzsáki, 2010) suggests that understanding CAs in the brain should focus on how they affect downstream "reader-actuator" mechanisms. In other words, the significance of a group of active neurons, or a presumed CA, can only be determined by considering the explicit outputs they generate. To identify meaningful assemblies, it is important to examine synchrony among these neurons, and this synchrony becomes meaningful from the perspective of a reader mechanism capable of integrating events over time. This approach differs from alternative views of CAs that are based solely on the connectivity of neurons without considering their functional impact.

The paper introduces the idea that CAs and their activation sequences can be understood as having a hierarchical neural syntax, analogous to language. This syntactical organization defines relationships among assemblies, facilitating communication and computation in the brain. The temporal sequencing of discrete assemblies generates "neural words" and "sentences", and enables complex cognitive functions such as recalling, thinking, reasoning, and planning.

The dynamic changes in synaptic weights are referred to as "synapsembles", and they are supposed to have an important role in connecting different spiking CAs. Synapsembles serve two important functions. First, they limit the lifetime of neural assemblies to short time frames, typically lasting from subseconds to seconds. This temporal constraint helps in the organization and termination of assembly activity. Second, synapsembles bridge the gap between different CAs even when there is no ongoing spiking activity, by altering the strength of synaptic connections between neurons, thus connecting neuronal words separated by periods of inactivity. This dual role of synapsembles is considered essential for understanding how the brain's functional connectivity evolves over time.

(Huyck, 2020) outlines the proposal for a neural cognitive architecture that aims to closely approximate both neural and cognitive functions of the human brain. It employs spiking neurons as its fundamental units, emphasizing their ability to approximate biological firing behavior and synaptic modification effectively. The proposed architecture can help to elucidate how neural networks generate cognitive functions, to facilitate the integration of various systems, such as rule-based and associative memory systems, and to enable the exploration of neural behavior in the context of cognitive tasks.

(Valiant, 2005) presents the *neuroidal* model, an original perspective on neural systems and their computational capabilities. The central contributions revolve around the development of two primary computational functions: memory formation (a "join" function), and the association between items already represented in the neural system (a "link" function). These functions are designed to be consistent with neural parameters observed in biological systems.

The article emphasizes the need for their implementation without deleterious interference or side effects, ensuring that they do not disrupt previously established neural circuits or result in unintended consequences. Furthermore, it explores the concept of using random graphs as a computational framework for achieving memory and association tasks within neural systems. It demonstrates that random graphs, even with relatively weak synaptic strengths, can be used to create memory formation and association pathways. The identification of their unexpected computational power is a significant contribution. The paper also introduces the concept of chains of nodes for communication within neural networks, extending previous proposals such as "synfire chains". It highlights the necessity for these communication chains to exist ubiquitously in the network, as opposed to being localized to specific areas.

## 8. Dynamic Models

*Neural dynamics* studies the continuously changing states of neuronal firing, the emergence of synchronous patterns, and the collective behavior of neural populations. Such investigations provide good insights into the underlying mechanisms of perception, cognition, and other various neural phenomena. These dynamic processes are essential to understand how the brain manages information, encodes memories, and produces complex behaviors.

In this section, we will present dynamic models in the brain from three perspectives. The first is related to a set of equations describing the activity of large network of neurons. The second refers to a class of computational dynamic models known as "reservoir computing". The third presents some models that exhibit complex behavior that can be analyzed with chaos theory methods, e.g., emphasizing the presence of attractors in the behavior of neural populations.

### 8.1. Wilson-Cowan Model[2]

In the 1970s, Wilson and Cowan published two classic papers that revolutionized theoretical neuroscience. In their first paper (Wilson & Cowan, 1972), they introduced a set of coupled equations offering a coarse-grained description of the dynamics of networks comprising both excitatory and inhibitory neuron populations. Coarse-graining involves the representation of a system at a lower resolution (from a higher point of view); in our case, it is about describing the behavior of groups of neurons rather than individual neurons. The second paper (Wilson & Cowan, 1973) extended their work to include spatial dependencies. These results marked a turning point in the field of computational neuroscience.

---

[2] This section uses information and ideas from: (Wilson & Cowan, 1972), (Wilson & Cowan, 1973), (Kilpatrick, 2013), and (Chow & Karimipanah, 2020).

In the 1960s, many concepts central to the Wilson-Cowan equations had been proposed, but they had not been integrated into a cohesive model. Wilson and Cowan aimed to develop a comprehensive description of neuronal activity that explicitly considered the distinction between excitatory and inhibitory cells. They were inspired by physiological evidence indicating the existence of specific populations of neurons with similar responses to external stimuli.

Previous studies and simulations explored spatially localized neural populations. These studies used various methods, including discrete and continuous time models. However, they all described the state of the population at a certain time using a single variable, such as the fraction of cells becoming active per unit time.

Given the complexity of the brain and the limited access to microscopic details, a statistical approach was considered necessary, involving concepts such as mean field theory. The challenge was to ensure that the lost information is not critical for understanding the phenomenon of interest. Mean field theory, originally developed for magnetism, serves as a coarse-graining method, used to describe and analyze the behavior of complex systems by simplifying the interactions between individual particles or components of a system, thus making it more tractable to study. The key idea behind mean field theory is to approximate the effect of interactions between components by treating them as if they are influenced by an average or "mean" field that is generated by the other components. In other words, each component in the system interacts with an effective, averaged-out field created by all the other particles, rather than considering the specific interactions with each individual entity. This simplification allows one to obtain analytical solutions that provide insights into the collective behavior of the system, even when the individual interactions are highly complex and difficult to model precisely. The approach is analogous to using statistical thermodynamics to relate the low level Brownian motion of particles to high level properties such as temperature, pressure or diffusion.

Wilson and Cowan explicitly considered the interactions between two distinct subpopulations of excitatory and inhibitory neurons, in accordance with Dale's principle. The authors showed that the dynamics described by their equations displayed multi-stability and hysteresis, which could provide a basis for memory, as well as limit cycles with oscillation frequencies modulated by stimulus intensity. A theorem introduced in (Wilson & Cowan, 1972) states that neural populations exhibiting limit cycle behavior for one class of stimuli show simple hysteresis for another class of stimuli.

Hysteresis (Figure 8.1a) refers to a phenomenon where the neural population response exhibits a delay or a non-linear relationship concerning changing input conditions, because its state does not only depend on the current input but also on its past history, i.e., previous input conditions continue to influence its current state. A limit cycle (Figure 8.1b) represents a stable, repetitive oscillatory pattern in the neural population response to certain stimuli. This cyclic behavior often corresponds to rhythmic activities observed in the brain, such as oscillatory patterns in various cognitive processes.

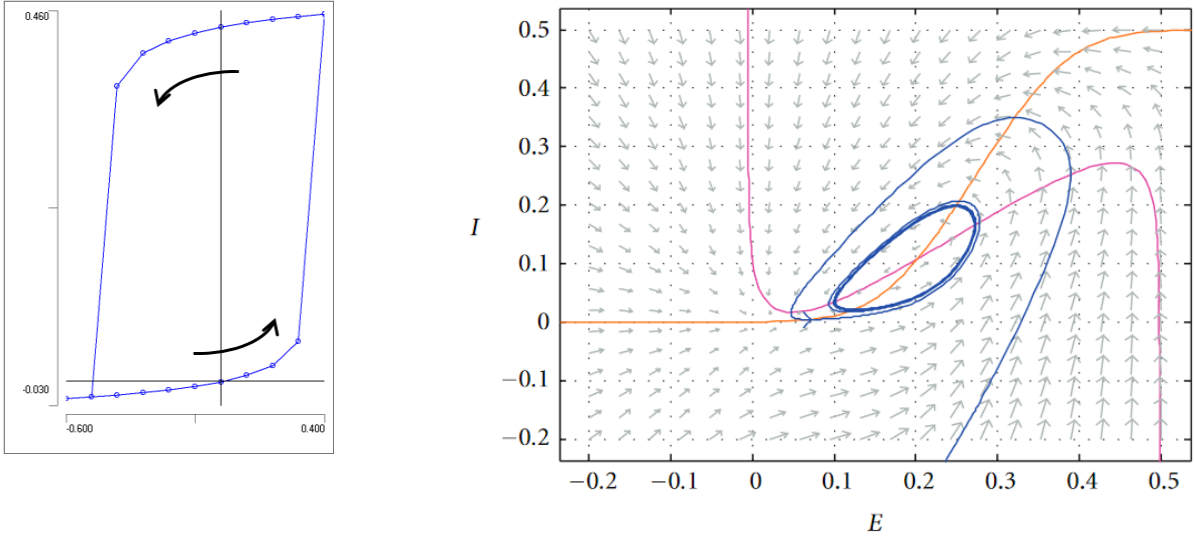

**Figure 8.1.** Hysteresis and limit cycles in the Wilson-Cowan model
a) The *X* axis shows excitation applied to the *E* population, and the *Y* axis shows the *E* activity level.
The arrows show the direction of time (Heitler, 2019)
b) A limit cycle with a fixed point inside (van Drongelen, 2013)

The original Wilson-Cowan equations provide a description of neural activity as continuous variables that reflect the proportion of local neuron populations firing at any given time. They can model many neural phenomena observed in the brain. The activity equations proposed in (Wilson & Cowan, 1972) are:

$$\tau_e \frac{da_e}{dt} = -a_e + (1 - r_e \cdot a_e) \cdot f_e(w_{ee} \cdot a_e - w_{ei} \cdot a_i + I_e) \tag{8.1}$$

$$\tau_i \frac{da_i}{dt} = -a_i + (1 - r_i \cdot a_i) \cdot f_i(w_{ie} \cdot a_e - w_{ii} \cdot a_i + I_i) \tag{8.2}$$

Here, $a_e$, $a_i$, $I_i$, $I_e$ are in fact functions of time. $a_e(t)$ and $a_i(t)$ are the proportion of excitatory and inhibitory cells firing at time *t*. The effects of neurons on their neighbors are modeled with a nonlinear (e.g., sigmoid) function *f* of the presently active proportion of cells:

$$f_e(x) = \frac{1}{1 + e^{-\gamma_e \cdot (x - \theta_e)}} \tag{8.3}$$

$$f_i(x) = \frac{1}{1 + e^{-\gamma_i \cdot (x - \theta_i)}} \tag{8.4}$$

When the synaptic weights among excitatory neurons reaches a sufficiently high level, this leads to the emergence of multiple steady fixed points, characterized by states of high and low excitation. Conversely, if the interconnections among inhibitory groups weaken sufficiently, the system displays limit cycle solutions, which involve a minority of active excitatory cells causing the activation of others, subsequently triggering inhibitory cells to deactivate the entire ensemble, and thereby restarting the cycle. This process represents a straightforward mechanism underlying oscillations in firing rate activity, which generally happen in biological neural networks.

The equations also account for the refractory periods of both populations (when cells are not capable of responding to stimuli after an activation) through the (1 – r · a) factors. *I* represents the cumulative input of external currents received by a population, e.g., from other brain areas.

Introducing spike rate adaptation can result in the emergence of traveling waves of neural activity (Hansel & Sompolinsky, 1998). Also, including short-term plasticity effects that dynamically modulate the synaptic weight functions *w* can lead to similar phenomena (Kilpatrick & Bressloff, 2010).

## 8.2. Reservoir Computing

*Reservoir computing* (RC) is a class of architectures derived from recurrent neural networks. Its main idea is using the dynamics of a fixed, nonlinear system called a *reservoir*, treated like a black box. This layer consists of interconnected neurons that exhibit dynamic behaviors. These neurons process input data over time, and their interactions create complex temporal patterns. The reservoir layer serves as a nonlinear transformation of input data into a high-dimensional space. Unlike conventional NNs that require extensive training on the entire network, reservoir computing only trains a linear *readout* layer that transforms the state of the reservoir into the desired output.

Reservoir computing was independently invented three times: by Peter Dominey when studying the physiology of sequence learning in the primate prefrontal cortex (Dominey, 1995), by Herbert Jaeger as *echo state networks* (ESN) (Jaeger, 2001; Jaeger & Haas, 2004), and by Wolfgang Maass as *liquid state machines* (LSM) (Maass, Natschläger & Markram, 2002). The main difference between the last two is that LSMs are biologically inspired spiking neural networks with integrate and fire neurons, whereas ESNs use a rate-based model where neurons have sigmoidal activation functions. The *reservoir computing* term was later introduced (Verstraeten et al., 2007) as a more general name to account for such architectures.

In 1995, Dominey tried to model the oculomotor circuit of primates, including the ability of conditional behavior, i.e., choosing the correct saccade target from among several possibilities based on learning. He studied how different visual cues could be associated with different targets, and how spatial sequences could be associated with output saccade sequences. The prefrontal cortex circuit was modeled as a reservoir of neurons with recurrent connections, followed by a system that computes the output, which is trained using rewards (a model of the striatum, a component of the basal ganglia). The recurrent connections (both excitatory and inhibitory) were fixed, but produced rich spatiotemporal dynamics. Neural plasticity was used to map the states of the recurrent network with the desired responses in the output layer.

### 8.2.1. Echo State Networks[3]

*Echo state networks* are computational models used, e.g., for the supervised learning of temporal (time series) data. The input data is designated by a vector **u** of size *T*. $\mathbf{u}(t) \in \mathbb{R}^{nu}$, also denoted as $\mathbf{u}_k$, signify the elements of this vector at discrete time steps. The task is to learn a desired (or target) output $\mathbf{y}^d(t) \in \mathbb{R}^{ny}$ for each time step. In practice, the dataset can consist of multiple sequences of varying lengths.

---
[3] This section uses information and ideas from: (Jaeger, 2001), (Lukoševičius, 2012), (Yildiz, Jaeger & Kiebel, 2012), (Schaetti, Salomon & Couturier, 2016), and (Nakajima & Fischer, 2021).

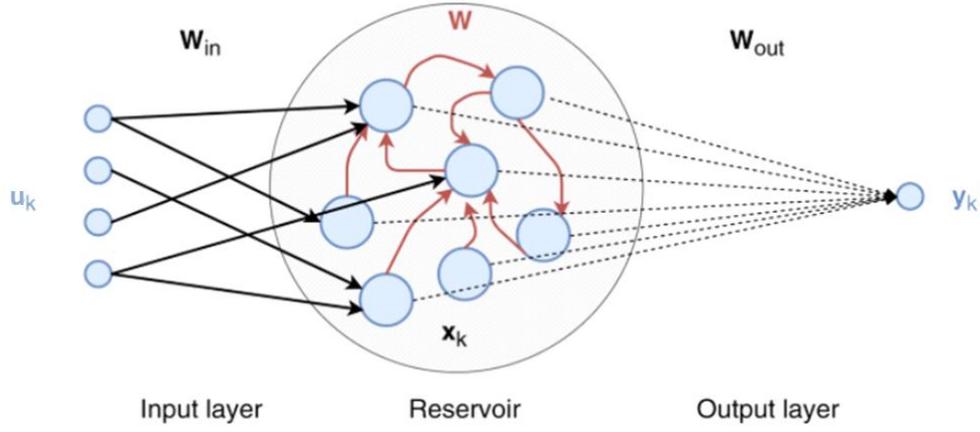

**Figure 8.2.** The architecture of an echo state network
(Verzelli, Alippi & Livi, 2019)

The architecture of an ESN is presented in Figure 8.2. It has an input layer characterized by the weights $\mathbf{W}_{in}$, the reservoir layer with recurrent connections with the weights $\mathbf{W}$ and the states of the units $\mathbf{x}$ (where $\mathbf{x}(t) \in \mathbb{R}^{nx}$), and the output layer with the weights $\mathbf{W}_{out}$ and the resulting output states $\mathbf{y}$. It is also possible to have feedback weights $\mathbf{W}_{fb}$ between the output units $\mathbf{y}_k$ and the reservoir units $\mathbf{x}_k$. They are more rarely used, so in the following equations we will omit them. However, they may be necessary for difficult problems, and feedback ensures universal computational capabilities for reservoirs (Maass, Joshi & Sontag, 2006).

The number of units in the reservoir can be large, e.g., 10000. Similar to a kernel in support vector machines (SVM), the reservoir projects the data in a high-dimensional space, and the bigger this space is, the easier it may be to find a linear combination of features to approximate the desired output.

The $\mathbf{W}_{in}$ and $\mathbf{W}$ matrices are randomly generated, while $\mathbf{W}_{out}$ is learned (or, in fact, computed). $\mathbf{W}$ may be sparse (but not necessarily), while $\mathbf{W}_{in}$ is usually dense.

Given an input $\mathbf{u}(t)$, the states of the reservoir units are updated as follows:

$$\tilde{\mathbf{x}}(t) = \tanh\left(\mathbf{W}_{in} \cdot \mathbf{u}(t) + \mathbf{W} \cdot \mathbf{x}(t-1)\right) \tag{8.5}$$

$$\mathbf{x}(t) = (1-\alpha) \cdot \mathbf{x}(t-1) + \alpha \cdot \tilde{\mathbf{x}}(t) \tag{8.6}$$

where $\alpha$ is a user-defined parameter called the *leaking rate*.

The output of the *readout* layer is:

$$\mathbf{y}(t) = \mathbf{W}_{out} \cdot [\mathbf{u}(t); \mathbf{x}(t)] \tag{8.7}$$

where "[ · ; · ]" represents vertical concatenation.

The units in the reservoir and in the output also have a bias value, which we have omitted in the equations for simplicity. The bias can be thought of as an additional input with a constant value of 1 and a corresponding column in the weight matrices. The explicit equations containing the bias are given, e.g., in (Lukoševičius, 2012).

The training procedure is the following. After the reservoir weights are generated, the inputs $\mathbf{u}(t)$ are presented and the reservoirs states $\mathbf{x}(t)$ are computed and stored. Then, the readout weights are computed by linear regression such that the mean squared error between $\mathbf{y}$ and $\mathbf{y}^d$ is minimized:

$$E(\mathbf{y}, \mathbf{y}^d) = \frac{1}{ny} \sum_{i=1}^{ny} \frac{1}{T} \sum_{t=1}^{T} \left( y_i(t) - y_i^d(t) \right)^2 \tag{8.8}$$

This can be done iteratively by gradient descent, or by matrix computations. Because $\mathbf{y} = \mathbf{W}_{out} \cdot \mathbf{x}$, one can use the pseudoinverse method to compute $\mathbf{W}_{out}$:

$$\mathbf{W}_{out} = \mathbf{y}^d \cdot \mathbf{x}^T \cdot \left( \mathbf{x} \cdot \mathbf{x}^T \right)^{-1} \tag{8.9}$$

In addition, regularization is often used to improve the generalization capabilities:

$$\mathbf{W}_{out} = \mathbf{y}^d \cdot \mathbf{x}^T \cdot \left( \mathbf{x} \cdot \mathbf{x}^T + \lambda \cdot \mathbf{I} \right)^{-1} \tag{8.10}$$

where $\mathbf{I}$ is the identity matrix and $\lambda$ is the regularization parameter.

An important concept related to ESNs is the *echo state property* (ESP), which is a condition of asymptotic *convergence* of the reservoir *state*, in the presence of input.

In general, if the elements of $\mathbf{W}$ are small, the input signal is attenuated, and when they are larger, increasingly irregular behaviors develop due to the recurrent connections in the reservoir. As these weights increase, oscillations may first appear, and then chaotic behavior. Therefore, the reservoir weights can be scaled as necessary, so that the reservoir should be able to learn properly.

In order to define the ESP, it is required for the functions that generate $\mathbf{u}$ and $\mathbf{x}$ to be defined on compact sets. We can simply consider here that this condition is met because $\mathbf{x}$ is generated from the hyperbolic tangent function which is bounded, and usually the input data $\mathbf{u}$ is normalized before training.

The echo state property states that, when a reservoir network is given an *infinite* sequence of inputs and begins from different initial conditions due to the stochastic nature of its weight matrix $\mathbf{W}$, resulting in different states $\mathbf{x}_1$ and $\mathbf{x}_2$, the reservoir states eventually become equal: $\mathbf{x}_1 = \mathbf{x}_2$.

Therefore, the state $\mathbf{x}$ of the reservoir can be understood of an "echo" of its input history, hence the name of the model. In other words, for a sequence of inputs that is long enough, the state of the reservoir is uniquely determined by the input history, and this state can be subsequently mapped into the desired output.

The most commonly employed technique to ensure that an ESN exhibits the ESP is through the *spectral radius*, defined as the maximum absolute eigenvalue of the matrix $\mathbf{W}$:

$$\lambda_{max} = \max_i |\lambda_i| \tag{8.11}$$

The reservoir weight matrix can then be scaled (before training) depending on this value:

$$\mathbf{W} = \frac{k \cdot \mathbf{W}}{\lambda_{max}} \tag{8.12}$$

When $k = 1$, the spectral radius of **W** becomes less than 1, a situation also noted $\rho(\mathbf{W}) < 1$.

While many researchers consider that this is a sufficient condition for the ESP, this is only true when the reservoir runs without inputs, i.e., $\mathbf{u}(t) = \mathbf{0}$. For typical input-driven systems, it is in fact a myth that $\rho(\mathbf{W}) < 1$ guarantees the ESP. This condition is neither sufficient nor necessary for the echo state property. Although this criterion may serve as an initial attempt, in general, spectral radii greater than 1 could potentially yield better results.

### 8.2.2. Liquid State Machines

*Liquid state machines* are similar to ESNs, with the same architecture, but they were designed with a stronger focus on biological plausibility. As mentioned earlier, LSMs use IF neurons and their dynamics are based on the spike trains they generate.

The "liquid" part corresponds to the reservoir. The term derives from an analogy with observing wave patterns formed by stones randomly thrown into water. After a certain number of time steps, distinct wave patterns emerge. These unique wave patterns contain important information about the features of the stones that were thrown. By analyzing them, it becomes feasible to draw conclusions about the properties of various stones. Similarly, the reservoir dynamics and resultant states encapsulate essential information about the input data, and this enables the possibility of meaningful computations based on these states. This is the principle behind the readout layer in both ESNs and LSMs.

### 8.2.3. Other Issues and Models

(Carroll, 2020) discusses the computational capacity of reservoir computers concerning their behavior at the "edge of stability". Previously, cellular automata were found to exhibit their highest computational capacity at the "edge of chaos", which marks a transition from order to chaos. Reservoir computers were expected to perform optimally at the "edge of chaos", but the paper challenges this assumption, suggesting that the term "edge of stability" may be more accurate because not all reservoir computers exhibit chaotic behavior. Still, the computational capacity of reservoir computers may not always peak at the "edge of stability". Instead, various factors, such as generalized synchronization with their input systems and the match between reservoir signals and the task at hand, play important roles in determining their performance.

(Gauthier et al., 2021) points out that standard RC methods rely on randomly generated parameter matrices and involve numerous meta-parameters to be optimized, for which there are few clear selection guidelines. It mentions that *nonlinear vector auto-regression* (NVAR) can achieve similar performance as reservoir computing, but with some key advantages that can lead to the "next generation reservoir computers". NVAR does not use a reservoir, does not rely on random matrices, requires fewer meta-parameters, and produces more interpretable results. It excels in benchmark tasks designed for reservoir computing and needs even less training data and time. Linear activation nodes in an RC combined with a feature vector composed of weighted nonlinear functions of reservoir nodes are as powerful as a universal approximator for dynamical systems. This approach is mathematically equivalent to an NVAR method, which is more efficient and straightforward than traditional RC while achieving similar results.

Classic RC methods have a fundamental limitation in addressing hierarchical modeling. Thus, they are not able to comprehensively capture the multi-scale compositional nature of some data. This is caused by the lack of bidirectional interactions between higher level modules and their corresponding lower level modules (Jaeger, 2021). Therefore, *deep reservoir computing* models were proposed. For example, (Gallicchio & Micheli, 2021) introduces the *deep echo state network* (DeepESN), an extension of the ESN model. Unlike the traditional ESN with a single reservoir layer, DeepESN has a hierarchical structure of stacked recurrent layers. The input layer is connected only to the first reservoir layer. Then, the $(n-1)^{th}$ layer is only connected to the $n^{th}$ layer. The global state of the network is the union of states across all layers. One significant characteristic of DeepESN is its stability and memory capabilities. Deeper architectures tend to exhibit longer memory spans and richer dynamics, even without training recurrent connections. It also has the ability to capture multiple time scales and represent multiple frequencies within sequential data. This hierarchical representation allows it to effectively process and encode complex temporal structures inherent in various types of time-series data. Moreover, its adherence to the echo state property ensures its stability and efficiency in handling temporal information.

In general, the advantage of reservoir computing methods is the very simple training procedure, since most of the weights are assigned only once, at random. Yet they are able to capture complex dynamics over time and are able to model the properties of dynamical systems.

## 8.3. Attractor-Based Models and Other Dynamic Models of the Brain

*Hopfield networks* (Hopfield, 1982) are perhaps well known in the AI community. Therefore, we will only briefly present their characteristics, together with some extensions. This type of recurrent network is typically used for associative memory tasks and pattern recognition. The structure of a Hopfield network comprises a single layer of interconnected neurons, each fully connected to every other neuron in the network, excluding a direct connection to itself. The activation of neurons is binary, which can be seen as an approximation of the on/off states of real neurons. The learning mechanism is Hebbian learning.

After training, due to its recurrent nature, a test instance makes the network converge to an attractor that should ideally represent the correct, possibly denoised version of the instance, i.e., one previously learned. The convergence process is equivalent to the minimization of a quantity called *energy* :

$$E(\mathbf{x}) = -\frac{1}{2}\mathbf{x}\mathbf{W}\mathbf{x}^T + \mathbf{b}\mathbf{x}^T = -\frac{1}{2}\sum_{j=1}^{n}\sum_{i=1}^{n}w_{ij}x_ix_i + \sum_{i=1}^{n}b_ix_i \qquad (8.13)$$

which is defined in terms of the weights **w** and biases **b** of the network.

It should be stated that equation (8.13) has the same form as the energy in the Ising model of ferromagnetism or spin glass in physics. Similarly, in a Hopfield network, the lowest energy is achieved when the activation of each neuron aligns with the "field" of weighted activations of its neighbors. When the network is run with an input pattern, its neurons may "flip" to achieve a better alignment until the minimum energy is reached.

One limitation is that the network may converge to incorrect patterns (false memories), particularly when the network needs to store a large number of patterns or when the stored patterns are highly correlated. Using statistical mechanics methods, researchers were able to assess the average storage capacity of Hopfield networks. Thus, if one attempts to store a number of instances larger than 14% of the network size, this results in an uncontrollable number of retrieval errors (Amit, Gutfreund & Sompolinsky, 1987). This storage limit was also computed for other variants of the architecture, e.g., in a sparsely connected network the ratio is larger, around 64% (Derrida, Gardner & Zippelius, 1987).

Hopfield networks can also be used for optimization. The user needs to set proper values for the weights **w** and biases **b**, and then the network is run starting from a random network state. In the minimum energy configuration, the network state **x** expresses the solution of the optimization problem. This is a process similar to the adiabatic quantum optimization technique used by the *D-Wave* quantum computer (Boixo et al., 2014).

The *Boltzmann machine* (Sherrington & Kirkpatrick, 1975; Ackley, Hinton & Sejnowski, 1985) is a probabilistic extension of Hopfield networks. While Hopfield networks use binary activations, Boltzmann machines incorporate probabilistic activations computed with the sigmoid function. The *restricted Boltzmann machine* (RBM) (Smolensky, 1986) is a variant of the Boltzmann machine, specially designed for efficient training and inference. RBMs introduce constraints on the connections between visible and hidden units, forming a bipartite, fully connected graph. This constraint leads to a simplified training process that can be accomplished through techniques such as contrastive divergence.

RBMs serve as fundamental building blocks in *deep belief networks* (DBN) (Hinton, 2009), constructed by stacking multiple layers of RBMs. The introduction of DBNs marked a turning point in the deep learning field, laying the foundation for the current advancements. DBNs combine the power of unsupervised learning with the expressive capacity of deep neural architectures. Thus, they can learn hierarchical data representations by capturing complex features at multiple levels. The typical training process for a DBN involves two primary phases: pre-training, where each layer of the network is trained in an unsupervised layer-by-layer manner, and fine tuning, which employs a supervised approach such as backpropagation.

(Khona & Fiete, 2022) is a review that discusses the concept of attractors in the context of dynamical neural circuits. Attractors represent specific states to which the system tends to evolve over time, and the paper covers various aspects related to their definition, properties, and mechanisms.

As indicated before, attractors are sets of states within the state space of a dynamical system, where all nearby states tend to converge. These states can be stationary, periodic, or chaotic, depending on the behavior of the system. Since the states of neural circuits are difficult to define, considering the complexities arising from the interconnected subcircuits and the influence of external inputs, the authors suggest to simplify this definition by using the time-varying spike rates of neurons as states for modeling neural circuits. Attractors have various forms, such as discrete attractors, which are stable fixed points, and continuous attractors, which are formed through patterns of neural activity. Continuous attractors are formed when network weights exhibit a continuous symmetry, such as translational or rotational invariance. Pattern formation, driven by local excitation and inhibition, plays a critical role in creating stationary continuous attractors. Large networks with strong asymmetric weights can exhibit chaotic dynamics: lower-dimensional but highly structured attractors. Networks with a dominance of inhibitory synapses may have a single attractor at zero activity, with transient deviations from it in response to perturbations.

Regarding their main uses, attractor networks provide stable internal states that can be used for reproducible representation of inputs. These representations can be achieved through a feedforward learning process, where external states are associated with internal attractor states. Attractor networks also exhibit two forms of memory, one in the structure of the weights and the other in the ability to maintain persistent activity in a stationary attractor state. Attractor states act as robust representations because they perform denoising: when noisy or corrupted inputs are presented, attractor dynamics drive the system into the nearest attractor state, eliminating most noise, which is especially important for memory maintenance. Attractor networks can also be used for decision-making processes, integrating positive and negative evidence to make decisions based on evidence accumulation (which we will address in Section 11.2). Winner-take-all models implement hybrid analog-discrete computations, allowing for rapid and accurate decision making among multiple options. Also, attractor dynamics are essential for stabilizing long-term behaviors such as sequence generation. Sequences can be generated as low-dimensional limit cycle attractors, with high-dimensional perturbations corrected, and systematic, periodic, or quasiperiodic flow of states along the attractor. However, the small components of noise along the limit cycle attractors may lead to timing variability.

(Zhang, Sun & Saggar, 2022) introduces a new modeling framework that examines the relationship between functional and structural connectivity in the brain, trying to explain how the brain's intrinsic dynamics are linked to cognitive functioning. More specifically, the paper tries to connect the continuously changing dynamics to the static anatomy of the brain. The authors suggest that cross-attractor coordination between brain regions predicts human functional connectivity better than single-attractor dynamics driven by noise. A cross-attractor represents a form of interaction between different attractors. It occurs when different attractors exhibit a degree of synchronization. In this case, the system can switch between attractors in a precise manner, allowing for dynamic interactions between them. Cross-attractor coordination describes how different brain regions or neural states interact with one another, potentially leading to coordinated patterns of neural activity, and this is relevant for understanding how different parts of the brain work together. The paper also discusses the idea that the transitions between attractors impose an energy cost and the modeling framework can help predict such costs associated with cognitive functions and psychiatric disorders.

(Spalla, Cornacchia & Treves, 2021) presents a continuous attractor network model designed to capture the dynamic nature of episodic memory retrieval. What sets this model apart is the introduction of memory-dependent asymmetric components in the synaptic connectivity. It considers that the connectivity between neurons is not symmetric, which is a departure from traditional attractor networks. The asymmetric connectivity is essential for creating dynamic memory retrieval because it can produce a robust shift in the activity of the network. When the network is given an initial cue, the asymmetric connections cause the neural activity to move along the memory attractor, effectively recalling the memory in a dynamic and time-dependent manner. The memory-dependent asymmetric connectivity is assumed to result from a learning phase involving STDP. The model emphasizes the balance between two components: one is symmetric and trajectory-averaged, while the other is asymmetric and trajectory-dependent. This balance is essential for the operation of the model. An interesting feature is that the retrieval speed can vary depending on the sparsity of the network activity. This characteristic aligns with observations of replay in the hippocampus, where memory recall occurs at different speeds. Another result of the paper is the quantification of the storage capacity for dynamic continuous attractors, which is found to be substantial and sometimes higher than static attractors in specific network configurations.

(Pereira & Brunel, 2018) describes the development and analysis of a recurrent neural network model inspired by the attractor NN scenario for memory storage in the association cortex. The authors address the existing gap between theoretical models and experimental data by inferring learning rules and the distribution of stored patterns from recorded data of visual responses. Unlike classical attractor NN models, the retrieval states exhibit graded activity, with firing rate distributions that closely resemble the log-normal distribution. The inferred learning rules are demonstrated to be optimized for storing a large number of attractor states, approaching the maximal storage capacity within a family of unsupervised Hebbian learning rules. Furthermore, the paper highlights the identification of two distinct types of retrieval states within the model: one characterized by constant firing rates over time and another where firing rates fluctuate chaotically. This transition to chaotic dynamics at strong coupling is found to result in highly irregular neural activity while maintaining stable memory storage, revealing a surprising aspect of the network behavior under different parameter conditions.

(Ursino, Magosso & Cuppini, 2009) reports the development of a NN model employing Wilson-Cowan oscillators to simulate the recognition of abstract objects, shedding light on the role of gamma-band synchronization in high-level cognitive functions. Synchronization in the gamma band refers to a phenomenon in which groups of neurons in the brain exhibit coordinated and rhythmic firing patterns in a frequency range between 30 to 80 Hz. Each object is represented by a collection of four features arranged in topological maps of oscillators, connected via excitatory lateral synapses, which implement a similarity mechanism. This network takes into account previous experiences with objects stored in long-range synapses, using timing-dependent Hebbian learning. A two-dimensional lattice of oscillators, which resembles the cortical structure, is employed. The model includes a decision network downstream the oscillator network, which indicates whether the detection of an object was correct when all the features oscillators are synchronized. In the first phase of the process, the stimulus is matched against the stored content, where inter-area synapses represent memory. In the second step, the induced activity is responsible for the utilization of the obtained matching information by the downstream decision network. This is similar to the operations in the "match and utilization model" (Herrmann, Munk & Engel, 2004). The authors find that the balance between sensitivity and specificity depends on the strength of the Hebbian learning, but also on the oscillation frequency. The model seems to be able to reconstruct and segment objects even under conditions involving deteriorated information, noise, and moderate feature correlations.

(Siri et al., 2007) focuses on investigating the effects of Hebbian learning in random recurrent NNs with biologically inspired connectivity. The network architecture is designed to mimic the characteristics of real neural networks, including sparse connections and separate populations of excitatory and inhibitory neurons. The authors consider firing-rate neurons with sigmoid activation functions and discrete-time dynamics where learning occurs on a slower time scale than neuron dynamics, i.e., synaptic weights are kept constant for $T \geq 1$ consecutive steps, which defines a learning epoch, and the weights are updated only after one such epoch. A variant of Hebbian learning rule is used:

$$w_{ij} = \lambda \cdot w_{ij} + s_i \cdot \frac{\alpha}{N} \cdot m_i \cdot m_j \cdot H(m_i) \tag{8.14}$$

where $i$ corresponds to the presynaptic neuron and $j$ to the postsynaptic neuron, $\lambda$ is the forgetting rate, $s_i$ is +1 if $i$ is excitatory and –1 if it is inhibitory, $\alpha$ is the learning rate, $N$ is the total number of neurons, set to 500 in the study, and $H$ is the Heaviside function, i.e., $H(x) = 0$ if $x < 0$ and 1 otherwise.

$m_i$ and $m_j$ represent the history of activities defined as a time average of firing rates:

$$m_i = \frac{1}{T}\sum_{t=1}^{T} a_i(t) - \theta_i \tag{8.15}$$

where $a_i(t)$ is the activation (firing rate) at time step $t$ and $\theta_i$ is a threshold, set to 0.1 in the paper, meaning that a neuron is considered active in a learning epoch if its average firing rate is at least 10% of the maximum value.

The application of Hebbian learning results in significant changes in both the dynamics and structural properties of the neural network. One notable effect is the contraction of the norm of the weight matrix. This implies that Hebbian learning leads to a rewiring of the network as synaptic connections are modified based on correlations that progressively develop between neurons. Within the emerging synaptic structure driven by Hebbian learning, the strongest synapses tend to organize as a small-world network, i.e., where most neighboring nodes can be reached from every other node in a small number of steps, which can enhance the efficiency of information transfer. The application of Hebbian learning leads to the contraction of the spectral radius of the Jacobian matrix, which drives the NN to the "edge of chaos", a state where it exhibits maximal sensitivity to input patterns. The effects observed in the simulations are primarily mediated by the passive forgetting term in the learning rule.

Another interesting result is presented in (Trapp, Echeveste & Gros, 2018), which reports that using a Hebbian plasticity rule called the "flux rule" leads to an asynchronous chaotic state where excitatory and inhibitory activations are balanced, but also the synaptic excitatory and inhibitory weights are balanced. The study explores the possibility of a neural network to achieve balance in both synaptic weights and activities in a fully unsupervised way. It is observed that such a balanced network arises in a self-organized manner, similar to the critical state of cortical dynamics.

The paper compares the effects of different Hebbian learning rules and finds that Hebbian learning leads to a balanced distribution of synaptic weights when the learning rule favors small average membrane potentials. Specifically, the flux rule leads to E-I balance, while, e.g., Oja's rule does not. The multiplicative self-limiting Hebbian (or flux) rule is defined as:

$$\frac{dw_{ij}}{dt} = \alpha_w \cdot G(x_j) \cdot H(x_j) \cdot y_j \tag{8.16}$$

$$G(x_j) = x_0 + x_j \cdot (1 - 2y_j) \tag{8.17}$$

$$H(x_j) = 2y_j - 1 + 2x_j \cdot (1 - y_j) \cdot y_j \tag{8.18}$$

where $i$ corresponds to the presynaptic neuron and $j$ to the postsynaptic neuron, $x_i$ is the membrane potential, $y_i$ is the firing rate, and $\alpha_w$ is the learning (or "adaptation") rate, which is 0.01 in the paper.

The study uses rate-encoding neurons and observes that the E-I balance is characterized by near cancellation of large excitatory and inhibitory inputs. This state is sensitive to small imbalances, leading to spike-like bursts in neural activity, indicating that the activity remains asynchronous. The near cancellation of excitatory and inhibitory inputs stabilizes asynchronous neural activity, and the authors find that this activity is strongly irregular. Despite the irregularity, the network remains stable and balanced, reflecting the dynamics observed in actual neural systems.

# 9. Brain Anatomy[4]

## 9.1. The General Structure of the Brain

Figure 9.1 provides an illustration of the general anatomical structure of the human brain. The outermost layer is the neocortex, a highly convoluted sheet responsible for complex cognitive functions, which is the primary focus of our discussion.

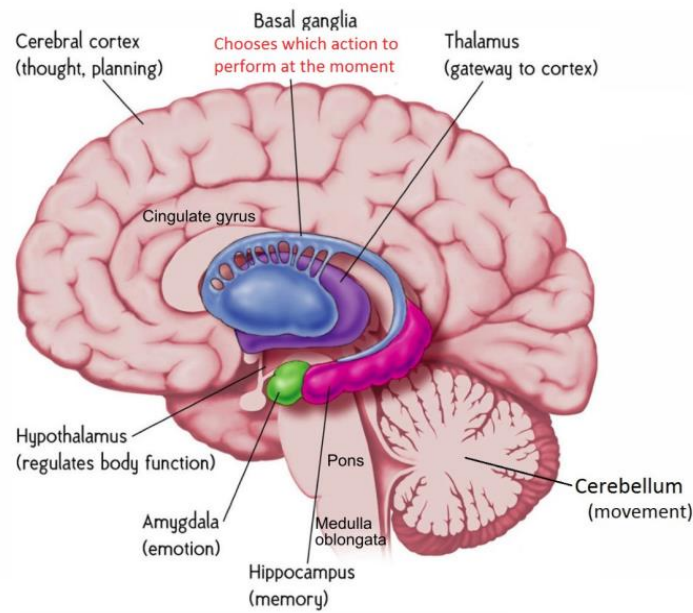

**Figure 9.1.** The general structure of the brain
(O'Reilly et al., 2012)

The *cerebral cortex* is the outermost layer of the brain. It is responsible for processing sensory information, such as touch, vision, and hearing, as well as controlling voluntary muscle movements. It is also involved in higher-order functions like problem solving, decision making, language processing, and emotional regulation. The terms "cerebral cortex" and "neocortex" are often used interchangeably, but they refer to slightly different aspects of the brain's structure. The cerebral cortex is a broader term that encompasses both the neocortex and other regions (e.g., the allocortex). The cerebral cortex is the outer covering of the surfaces of the cerebral hemispheres. In the human brain, it has around 15 billion neurons, is between 2-4 mm thick, and makes up 40% of the brain's mass.

The *neocortex*, on the other hand, is a specific part of the cerebral cortex that is involved in higher-order cognitive functions such as conscious thought, sensory perception, spatial reasoning, and language processing. It is called the "new cortex" because it is considered the most evolutionarily recent addition to the brain and is highly developed in mammals, particularly in humans. The neocortex makes up approximately 90% of the cerebral cortex. It is characterized by its *six-layered structure* and is responsible for many of the complex cognitive processes associated with intelligence and advanced sensory processing.

---

[4] This section uses information and ideas from: (Mountcastle, 1957), (Roberts, 1992), (Saladin, 2010), (O'Reilly et al., 2012), (Halber, 2018), and (Hawkins, 2021).

Beneath the neocortex, there are several brain regions collectively referred to as subcortical brain areas.

The *hippocampus* plays a crucial role in the formation of episodic memories, including everyday memories of events and facts. It is a brain structure involved in memory formation and spatial navigation and plays a central role in converting short-term memories into long-term ones, a process known as *memory consolidation*. The hippocampus also helps the navigation in the environment by creating *cognitive maps*.

The *amygdala* is essential for recognizing emotionally significant stimuli and signaling other brain areas of their presence.

The *cerebellum* primarily coordinates motor functions. It employs an error-driven learning mechanism, enabling the refinement of actions. It contains a significant proportion (80%) of the brain's neurons, but represents only 10% of the brain's total mass.

The *thalamus* serves as the main relay station that acts as a gateway for sensory information on the way to the neocortex. It receives sensory input from various parts of the body, processes it, and then relays it to the appropriate regions of the cerebral cortex, where perception and conscious awareness occur. Additionally, the thalamus plays a role in regulating consciousness, alertness, and sleep-wake cycles. It likely contributes to attention, arousal, and other modulatory processes (that enhance or suppress the activity of other processes). It filters and prioritizes sensory stimuli, helping to focus on relevant information while filtering out distractions.

The *basal ganglia* (BG) is a network of subcortical regions that plays an important role in decision making. It determines whether to execute or inhibit the actions proposed by the cortex, based on prior experiences of rewards and punishments. It possesses the capability to initiate a disinhibitory "go" signal, facilitating the selection of the best action among the potential choices. This action selection process is greatly influenced by reinforcement learning, guided by the presence of dopamine, a neurotransmitter that modulates both learning and the speed of action selection itself. Thus, the basal ganglia act as arbiters, favoring actions that are likely to yield rewards while minimizing the likelihood of punishment.

**9.2. The Cortical Lobes**

The cerebral cortex is divided by a longitudinal fissure into two hemispheres, each containing several lobes, shown in Figure 9.2.

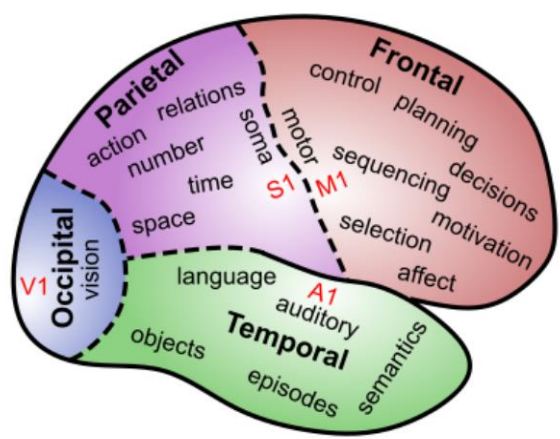

**Figure 9.2.** The cortical lobes
(O'Reilly et al., 2012)

The *frontal lobe* is the largest and most complex of the cerebral lobes, encompassing a hierarchical organization of motor control regions. At its posterior end is the primary motor cortex (M1), which governs basic motor movements. The *prefrontal cortex* (PFC), situated at the front of the lobe, is often considered the brain's executive center. It oversees high-level cognitive functions such as decision making, long-term planning, goal setting, and reasoning. Furthermore, the middle and front regions regulate emotions and motivations.

The *parietal lobe* carries out several functions that tend to operate at a subconscious level. It processes spatial information, forming the "where" pathway for understanding the locations and movements of objects in the environment. It also plays a role in numerical processing, mathematics, and abstract conceptualization. It aids in integrating visual information to guide motor actions, forming the "how" pathway. The primary somatosensory cortex (S1), which is responsible for processing tactile sensations, is located here. Some areas facilitate the translation of information between different frames of reference, such as mapping spatial locations on the body to visual coordinates.

The *temporal lobe* houses the primary auditory cortex (A1) and regions involved in auditory and language processing. It plays an important role in recognizing and classifying visual objects. It also facilitates the translation between visual information and verbal labels, contributing to the ability to name objects and understand their meaning. In addition, the medial temporal lobe (MTL), next to the hippocampus, is essential for the storage and retrieval of episodic memories.

The *occipital lobe* serves as the center for visual processing. It contains the primary visual cortex (V1), which is responsible for initial visual perception and the detection of basic visual features such as edges, shapes, and orientations.

## 9.3. The Visual Cortex

The visual cortex, involved in processing visual information, is organized on six layers.

*Layer 1* is its most superficial layer and consists of horizontal connections, serving as a site for the initial integration of incoming visual inputs.

*Layer 2/3* is where the initial stages of processing occur. It receives inputs from the thalamus and processes basic visual features such as edges, orientations, and basic textures. The neurons in this layer also engage in complex lateral interactions that contribute to feature detection and integration. Layer 2 is more focused on detecting basic visual elements, while layer 3 is involved in integrating these features and forming more sophisticated representations of visual stimuli.

*Layer 4* is specialized for relaying information from the thalamus. It acts as a primary input layer for visual stimuli. In this layer, neurons exhibit a high degree of specificity, responding to distinct features such as orientation and spatial frequency.

*Layer 5* serves as an intermediary layer, conveying processed visual information to higher-order cortical regions. It also plays a role in sending motor commands back to subcortical structures, thus participating in the coordination of motor responses to visual stimuli.

*Layer 6* has two sublayers that contribute to the feedback loop between the visual cortex and subcortical structures, including the thalamus, to modulate sensory processing, and the brainstem, to control the eye movements needed for visual tracking and fixation.

The layered structure of the visual cortex served as a fundamental source of inspiration for convolutional neural networks (CNNs), which excel in image processing and computer vision tasks. The role of the visual cortex in processing visual information in the human brain is analogous to the function of CNNs in computer vision. In the visual cortex, neurons are organized into distinct layers, each responsible for extracting different features from visual stimuli. Early layers capture basic features such as edges and simple shapes, while deeper layers process more complex patterns, textures, and object hierarchies. This hierarchical and layered approach allows the brain to efficiently analyze visual information. CNNs mimic this architecture by using convolutional layers that apply filters to extract low-level features and then stack multiple such layers to progressively capture higher-level features. The hierarchical feature extraction closely aligns with the operation of the visual cortex.

## 9.4. The Function of the Cortical Layers

The neocortex exhibits a consistent six-layer structure, which is uniform across all cortical areas. However, the relative thickness of these layers varies in different cortical areas.

The *input areas*, such as the primary visual cortex, receive sensory information, usually via the thalamus. These areas have a prominent layer 4, where thalamic axons terminate. A specialized type of excitatory neuron called the *stellate* cell, with dense and localized dendrites, plays a role in collecting the input in this layer. These areas neither directly receive sensory input nor directly control motor output. They are characterized by thicker superficial layers 2/3 that contain many *pyramidal* neurons suitable for classification functions.

The *hidden areas* are believed to create abstract classes from sensory inputs and facilitate appropriate behavioral responses based on these high-level classes.

The *output areas* connect directly to muscle control regions and can induce physical movement through electrical stimulation. These areas have thicker deep layers 5/6, which send axonal projections to various subcortical regions.

The layer-wise (laminar) structure of the cortex and the specific functions of different cortical areas collectively suggest that the primary role of the cortex involves processing sensory inputs in diverse ways to extract behaviorally relevant classes, which subsequently drive suitable motor responses.

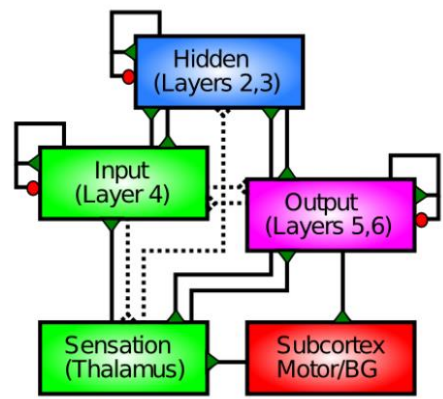

**Figure 9.3.** Connectivity map between cortical layers
The green triangles represent excitation and the red circles represent inhibition
(O'Reilly et al., 2012)

Within the same cortical area, there is extensive lateral connectivity among neurons. Also, the hidden layers within different areas can communicate directly with one another. This interaction across areas plays an important role in forming coherent representations and enabling cognitive functions. Cortical connectivity is largely bidirectional. Areas that send feedforward projections to other areas typically receive reciprocal feedback projections from the same areas. The feedback flow aids in top-down cognitive control, directing attention and resolving sensory input ambiguities.

Many stimuli are ambiguous without additional top-down constraints. Semantic knowledge and context can help the brain resolve ambiguity in favor of a coherent interpretation. For instance, after reading a clue about a Dalmatian in an ambiguous image (Figure 9.4), top-down knowledge helps to perceive the dog.

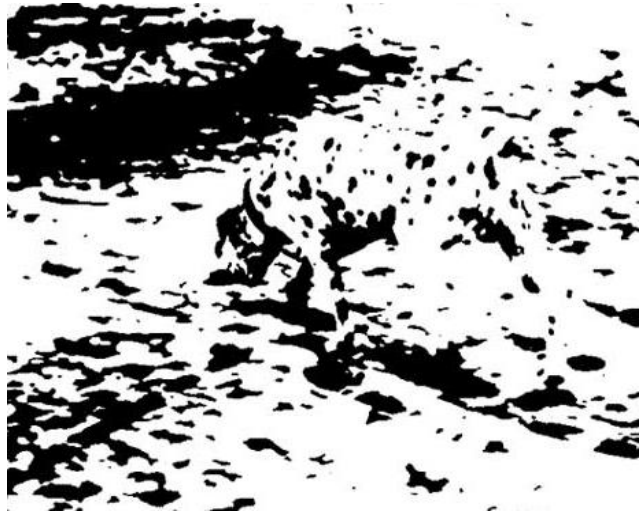

**Figure 9.4.** The Dalmatian dog illusion
(Gregory, 1971)

Bidirectional excitatory dynamics are essential for pattern completion. When given a partial input cue (e.g., "Where did you go last weekend?"), the appropriate memory representation can be partially excited in the brain. Bidirectional excitatory connections enable this partial excitation to propagate through memory circuits, filling in the missing parts of the memory trace. This process is like a system orbiting around the correct memory trace until it converges.

Occasionally, the system may fail to fully converge into the attractor state, leaving memories out of reach. These "tip of the tongue" states can be frustrating, and people may try various strategies to reach the final attractor state to retrieve the memory.

Inhibitory competition is vital because it allows individuals to focus their attention on a select few things at a time, preventing cognitive overload. It ensures that the detectors which are activated are those most excited by a specific input, similar to the concept of "survival of the fittest" in evolution.

Without inhibition, the bidirectional excitatory connections in the cortex would lead to rapid and uncontrolled excitation of all neurons. This is like the high frequency feedback sound produced when a microphone is placed near a speaker playing its own sound. Inhibition functions as a control mechanism, analogous to a thermostat in an air conditioner, maintaining activity at a desired set point and preventing it from becoming too high. Typically, only about 15-25% of neurons in a given area are active at any given time due to inhibition (O'Reilly et al., 2012), forming a *sparse distributed representation*.

Selective attention is a prime example of the role of inhibitory competition. When searching for a specific stimulus in a crowded scene, such as finding a friend in a crowd, people can only focus on a limited number of items at once. The neurons corresponding to the attended items are excited sufficiently to surpass the inhibitory threshold, allowing them to stand out. Inhibition helps to ignore or inhibit the rest of the items. Both bottom-up (sensory driven) and top-down (cognitively driven) factors influence which neurons surpass the threshold. Without inhibition, there would be no mechanism for selecting and focusing on a subset of items.

Individuals with Balint's syndrome, which results from damage to the parietal cortex responsible for spatial attention, exhibit reduced attentional effects. They are often unable to process multiple items simultaneously in a visual display, a condition known as "simultanagnosia", where they cannot recognize objects when multiple items are present in a scene. This is an idea related to the concept of variable binding, which we will discuss in Section 13.

Inhibitory competition also serves as selection pressure during learning. Only the most excited detectors get activated and are reinforced through the learning process. This positive feedback loop enhances the tuning of these detectors to the current inputs, making them more likely to respond to similar inputs in the future. Inhibitory competition leads to the specialization of detectors for specific classes in the environment. Without it, a large percentage of neurons would become trained for each input, resulting in weak detection abilities for everything.

## 9.5. The Language Areas

*Wernicke's area*, located in the left temporal lobe, is primarily associated with language comprehension and processing. It plays an important role in understanding spoken and written language, as well as formulating coherent and meaningful sentences. Damage to this area can result in Wernicke aphasia, a condition where individuals have difficulty with language comprehension and often produce fluent but nonsensical sentences.

*Broca's area*, situated in the left frontal lobe of the brain, is involved in the production of speech and language expression. It coordinates the muscles involved in speech production and enables the formation of grammatically correct and coherent sentences. Damage to this area can lead to Broca aphasia, characterized by difficulty in forming sentences and expressing language fluently, although comprehension remains relatively intact.

## 9.6. The Memory Areas

As mentioned above, the hippocampus plays a fundamental role in memory. It supports the fast learning of new information without causing excessive interference with what was previously learned. This rapid learning ability is valuable, e.g., when remembering names associated with people recently met.

The hippocampus uses sparse representations, which contribute to a phenomenon called *pattern separation*. This process ensures that the neural activity pattern associated with one memory is distinct from patterns linked to other similar memories. By minimizing overlap, pattern separation reduces interference with prior learning.

In addition to pattern separation, *pattern completion* is another critical process during memory retrieval. Pattern completion allows the recovery of memories from partial information, enabling the recall of complete memories even when presented only with fragments or cues.

The learning rate plays a essential role in the learning process. Higher learning rates enable rapid acquisition of new information. In contrast, lower learning rates facilitate the integration of knowledge across various experiences, leading to the development of semantic knowledge. The neocortex generally exhibits a lower learning rate than the hippocampus.

Even with a low learning rate, the neocortex can exhibit measurable effects after a single trial of learning. These effects include priming and familiarity signals that contribute to recognition memory, where one recognizes something as familiar without explicit episodic memory. Medial temporal lobe (MTL) areas are involved in this form of recognition memory.

**9.7. Grid Cells and Place Cells**

*Grid cells* are a type of neurons found in the entorhinal cortex, a region connected to the hippocampus. They display a characteristic hexagonal pattern of activity across space, as shown in Figure 9.5. The entorhinal cortex contains grid cells that represent the body's location relative to the environment. Grid cells fire in a regular pattern that resembles a grid, with multiple grid cells firing at different scales and orientations. The firing pattern of grid cells overlays a spatial grid on the environment, providing a system that the brain can use to estimate distances and navigate accurately.

Grid cells activate at multiple locations in an environment, forming grid-like patterns. Changes in grid cell activity reflect the individual's updated location as he/she moves.

Experimental evidence suggests that entorhinal lesions, which affect grid cells, impair performance on navigation tasks and disrupt the temporal ordering of sequential activations in the hippocampus. This implies a role for grid cells in spatial planning and raises the possibility of a more general role for grid cells in hierarchical planning.

A single grid cell module cannot represent a unique location on its own. To do that, it is necessary to consider the active cells in multiple grid cell modules, each with different tile spacing and/or orientation relative to the environment. The method of using multiple grid cell modules to represent locations has a large representational capacity. The number of locations that can be represented scales exponentially with the number of modules.

*Path integration*, which allows people to keep track of their location during movement, works from any starting location within the network. This provides a form of generalization, as the path integration properties learned for each grid cell module apply to all locations, even those the individual has never visited before.

Grid cell modules "anchor" differently when a person enters a familiar environment, such as a room. This anchoring process ensures that the current location and all possible locations within that environment have unique representations specific to that environment. The set of location representations are related to each other through path integration. Each location representation is unique to a particular environment and will not appear in any other environment. An environment encompasses all locations that the person can move among, including those he/she has not visited yet, but could potentially visit. Some of these locations may be associated with observable landmarks.

Grid cells in the brain are known to use a hexagonally symmetric code to organize spatial representations, as displayed in Figure 9.5. A single grid cell is activated at multiple locations when a person moves in the environment, e.g., the red circles in the figure. Therefore, one grid cell cannot identify a certain location. Different grid cell modules activate with different spacing and orientations e.g., the green circles in the figure. When several grid cells from different modules fire simultaneously, they can precisely identify a particular location.

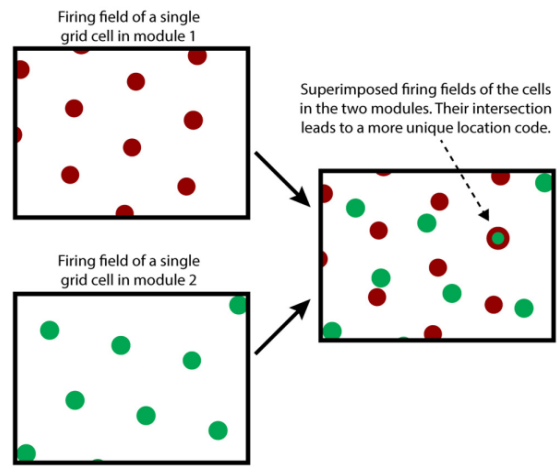

**Figure 9.5.** Location representation using grid cells
Adapted after (Hawkins et al., 2019)

This hexagonal symmetry is reflected in functional magnetic resonance imaging (fMRI) signals during spatial navigation, and may be responsible for the brain's ability to generalize from one environment to another.

(Rodríguez-Domínguez & Caplan, 2019) proposes a descriptive model of grid cells related to a two-dimensional Fourier basis, set in hexagonal coordinates. It suggests that grid cells may perform an inverse Fourier transform of the environment, enabling spatial representation and lossy compression of contextual information. In the hexagonal Fourier model, grid cells appear as specialized filters preserving meaningful information in a compressed form.

*Place cells* are specialized neurons found in the hippocampus that become active when an individual occupies a specific location in his/her environment. Each place cell has a preferred "place field", a specific area in the environment where it becomes most active. As an individual moves through space, different place cells activate, forming a representation of the person's current location. This dynamic activation of place cells creates a kind of neural map of the environment, allowing the brain to keep track of where it is (Hawkins et al., 2019).

*Cognitive maps* are mental representations of the spatial layout of the environment. These maps allow individuals to mentally navigate through space, even when they are not physically present there. Cognitive maps provide a way to plan routes, make decisions about navigation, and recall spatial relationships between different locations. They integrate the information provided by place cells and grid cells, allowing people to mentally navigate and plan routes (Hawkins et al., 2019).

A hypothesis is that the human brain also organizes *concepts* into a cognitive map that allows for the navigation of conceptual relationships in a manner similar to spatial navigation. The hippocampus plays a crucial role in organizing knowledge in the form of cognitive maps. The hippocampus is involved not only in spatial reasoning, but also in non-spatial functions, such as relational memory, which involves understanding relationships and associations between objects and events. These maps are essential for making complex inferences based on sparse observations.

Some researchers proposed a connection between spatial and non-spatial functions, suggesting that they may share a common underlying mechanism. However, this connection remains unclear, and the existing models do not fully account for the diverse types of spatial representations observed in the hippocampus.

## 9.8. The Decision-Making Areas

The prefrontal cortex provides top-down cognitive control over processing in the posterior cortex. It can influence action selection in basal ganglia-motor circuits and also impact attention to task-relevant features in sensory cortex. Another example of top-down control is the *Stroop task*, i.e., a psychological test where subjects are presented with a list of color words (e.g., "red", "blue", "green") written in colored ink. The task requires participants to name the ink color of each word while ignoring the actual word. It demonstrates top-down control because it requires individuals to exert cognitive control and selectively attend to one aspect of the stimulus (the ink color), while inhibiting or ignoring the other aspect (the meaning of the word).

The interaction between the prefrontal cortex and the basal ganglia contributes to the development of a working memory system that allows the brain to hold multiple pieces of information and separately update some of them while maintaining the existing ones.

## 9.9. The Homogeneous Cortical Columns

The "old brain" contains distinct structures, each with specific functions, visibly different in shape and size. In contrast, the neocortex, responsible for various cognitive functions, lacks any obvious divisions and appears as a continuous sheet of cells. The neocortex is very dense with neurons and connections. In 1 mm$^2$, there can be around 100,000 neurons, 500 million synapses, and several kilometers of axons and dendrites. Despite its uniform appearance, the neocortex contains regions responsible for various functions, such as vision, hearing, touch, language, and planning. Surprisingly, even the detailed neural circuits responsible for these functions do not appear significantly different from one another.

The regions within the neocortex are interconnected through bundles of nerve fibers, forming a complex network. Scientists have sought to understand this connectivity to determine the organization of the neocortex. One common interpretation of neocortical organization is a hierarchical flowchart-like structure, where sensory input is processed in steps, gradually extracting complex features until complete objects are perceived. While some evidence supports the hierarchical view, the actual organization of the neocortex does not precisely align with this model. Multiple regions exist at each level, and many connections do not fit a hierarchical scheme. In addition, not all cells in each region act as feature detectors.

Considering the six layer structure mentioned in Section 9.1, most connections between neurons occur vertically, between these layers. The majority of axons traverse between layers, indicating that information primarily moves up and down before being relayed elsewhere.

While some regions have variations in the types and quantities of neurons, these differences are relatively small compared to the overall uniformity. This similarity suggests that what makes regions distinct is not their inherent function but their connectivity. Connecting a region to different sensory inputs, e.g., eyes or ears, leads to different functions, such as vision or hearing. Higher-order functions, such as language, may also arise from the connections between regions.

Vernon Mountcastle (1957) suggested that the fundamental unit of the neocortex is the *cortical column*. These columns, occupying about 1 mm$^2$ on the surface and extending through the 2.5 mm thickness of the neocortex, are repeated throughout the neocortex. Each cortical column processes specific sensory information, and their grouping is defined by the parts of sensory inputs they respond to. Also, their function is flexible, e.g., in blind people the visual areas can adapt to process other sensory inputs, such as hearing or touch. The same structures are used for higher level cognitive functions or learned tasks.

Therefore, the homogenous organization of the cortex into columns may imply a *common brain algorithm* for various cognitive processes.

# 10. Neuroscience Models

## 10.1. Hierarchical Temporal Memory[5]

*Hierarchical temporal memory* (HTM) proposes a neural architecture and a learning theory inspired by the functioning of the neocortex. It relies on a model of the pyramidal neuron and includes algorithms especially fit for learning sequences in an unsupervised manner from continuous streaming data, and detecting anomalies in these streams.

### 10.1.1. The HTM Neuron Model

In the "point neuron" model (equation 3.1), dendrites are passive components that simply pass the received signals on to the neuron body for summation. However, in biological neurons it seems that dendrite branches are active processing elements that can generate nonlinear responses, e.g., in the pyramidal neurons, which are the primary type of excitatory neurons found in the neocortex.

Each neuron has thousands of synapses that act as independent pattern detectors. When any of these patterns is detected by the dendritic segment, it triggers an n-methyl-d-aspartate (NMDA) spike. These are not spikes transmitted through the axon, as described in Section 2.3, but dendritic spikes, highly localized events that occur within a small dendritic segment, typically 10-40 µm in length. NMDA spikes are generated when a small number of synapses are activated synchronously on a specific dendritic segment (8-20, around 15 on average), compared to the total number of thousands of synapses (Brandalise et al., 2016).

Accurate recognition relies on the sparsity of patterns, where only a few neurons are active. In this scenario, a dendrite can detect a specific pattern when all its synapses receive simultaneous activation. Consequently, a single neuron can recognize hundreds of patterns. The recognition of any pattern causes the neuron to depolarize. By establishing more synapses than needed to trigger an NMDA spike, recognition becomes resilient to noise and variations. Even if some cells change or become inactive, the dendrite can still successfully identify the target pattern.

Following Hebbian plasticity, synapses are strengthened when presynaptic input successfully induces postsynaptic activity. The activation of NMDA receptors and subsequent calcium ($Ca^{2+}$) level elevation triggers the biochemical mchanisms responsible for long-term potentiation, the synaptic strengthening process associated with learning and memory.

Dendrites are divided into *basal*, *proximal* and *apical* integration zones (Figure 10.1), each with distinct properties. These zones are considered to play different roles in information processing.

---

[5] This section uses information and ideas from: (Hawkins & Ahmad, 2016), (Cui, Ahmad & Hawkins, 2016), (Ahmad, Taylor & Cui, 2017), (Hawkins et al., 2020), and (Ahmad, Lewis & Lai, 2021).

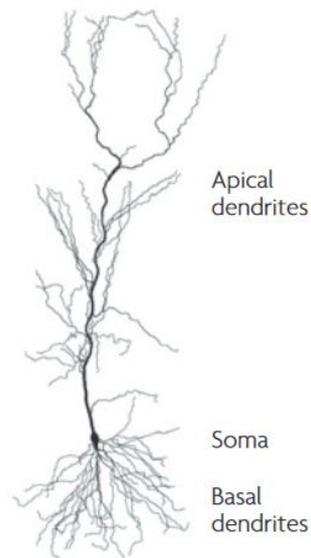

**Figure 10.1.** Pyramidal neuron structure and domains of synaptic input
(Spruston, 2008)

The HTM neuron model (Figure 10.2) handles these types of dendrites in different ways. Proximal dendrites are situated near the cell body and have a direct effect on the generation of the axon action potential. The basal and apical dendrites (also called *distal*) are farther away from the soma and they have an indirect, modulatory effect on the neuron behavior.

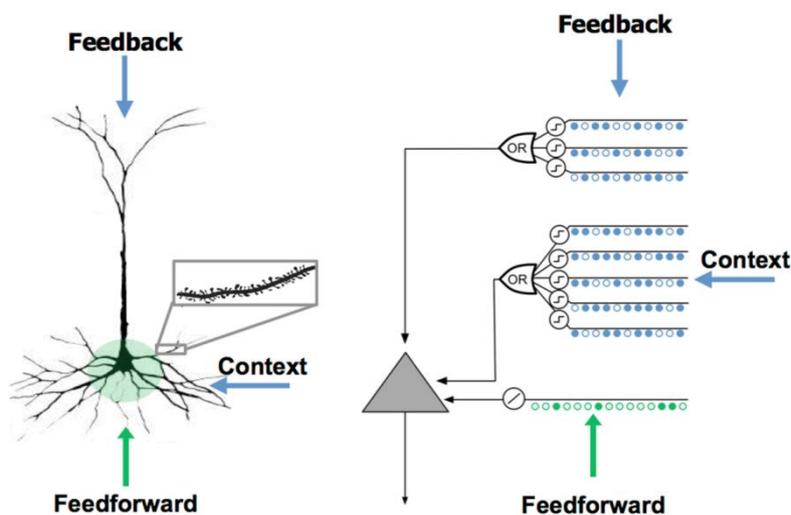

**Figure 10.2.** A comparison between a biological neuron (left) and a HTM neuron (right)
(Hawkins & Ahmad, 2016)

Proximal synapses, typically several hundred, are located on the proximal dendrites and have a relatively large effect at the soma. Their main role is to define the classic receptive field of the neuron. If a subset of proximal synapses are simultaneously activated, they can generate a somatic AP. These synapses are well-suited to recognizing multiple unique feedforward patterns when the inputs are sparsely active. Therefore, the feedforward receptive field of a neuron can be seen as a union of feedforward patterns.

Basal dendrites are responsible for recognizing patterns of cell activity that precede the neuron firing. When a pattern is recognized on a basal dendrite, it generates an NMDA spike. The resulting depolarization, although not sufficient to trigger a somatic AP, represents a prediction that the cell may become active shortly. This subthreshold depolarization may cause the cell to fire earlier than it would without it.

Apical dendrites can also generate NMDA spikes when they recognize a pattern. The depolarization caused by them is thought to create a top-down expectation, which can be considered another form of prediction.

Thus, proximal synapses directly trigger APs, while the patterns identified by the basal and apical dendrites depolarize the cell, and can be thought of as predictive signals. Basal synapses detect contextual patterns predicting the next feedforward input, while apical synapses identify feedback patterns forecasting entire sequences.

Neurons that have been previously depolarized generate spikes quicker than those that have not experienced depolarization. Together with the mechanism of fast local inhibition, this process biases the activation state of the network toward its predictions. This cycle of activation, prediction, and activation forms the basis of sequence memory. HTM is designed to emulate the brain's ability to process and recognize complex patterns, particularly in time-series data.

We will now describe two learning algorithms characteristic of HTM. However, they are presented in a simplified form, with some details being omitted.

*10.1.2. Spatial Pooling Algorithm*

In the *spatial pooling* algorithm, an input is considered to consist of a fixed number of bits. Then a HTM region is initialized, where a predetermined number of columns are allocated to receive this input. For example, a region may be configured as an array of 4096 columns. Spatial pooling operates at the level of these columns, with each column acting as a single computational unit (Figure 10.3). Each of these columns is equipped with an associated dendritic segment, which acts as the bridge connecting it to the input space, as presented in the HTM neuron in Figure 10.2.

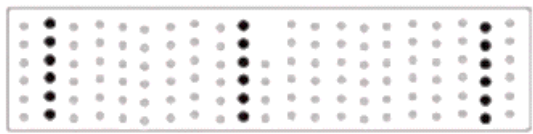

**Figure 10.3.** Example of activated columns in the spatial pooling algorithm
(Hawkins & Ahmad, 2016)

These dendritic segments contain a set of potential synapses, representing a random subset of the input bits. These synapses establish connections with specific bits within the input space. Each synapse is characterized by a *permanence value*, which is conceptually similar to its strength. Synapses can exist in one of three states: *connected* (when the permanence value exceeds a certain threshold), *potential* (when the permanence value falls below the threshold), and *unconnected* (without the ability to establish a connection).

The effect of a synapse is binary, meaning it can either be active or inactive. The permanence values are initially assigned random values centered around the permanence threshold. As a result, some synapses may already be connected if their permanence values exceed the threshold.

Once the initialization is complete, the algorithm begins its operation. For a given input, it evaluates how many connected synapses in each column are associated with active input bits, designating these as active synapses. Then, a small percentage of columns within an *inhibition radius* featuring the highest activations become active, suppressing the other columns within that radius. This phase determines which columns emerge as winners and continue to be active after the inhibition step.

At this stage, a sparse set of active columns remains. Subsequently, the region begins the learning phase with a Hebbian-style learning rule for adjusting the permanence values of the synapses of the active columns. For the winning columns, the permanence values of the synapses aligned with active input bits are increased, while those aligned with inactive input bits are decreased. The permanence values on the synapses of the non-winning columns remain unaltered.

These changes to permanence values may shift the state of some synapses from connected to potential and vice versa. This entire process then repeats.

### 10.1.3. Temporal Memory Algorithm

The *temporal memory* algorithm continues after the spatial pooling algorithm, with the initial state consisting of a set of active columns representing the feedforward input. More formally, we can say that it operates on sequences of *sparse distributed representations* (SDR). The primary function of the temporal memory algorithm is to learn sequences and make predictions.

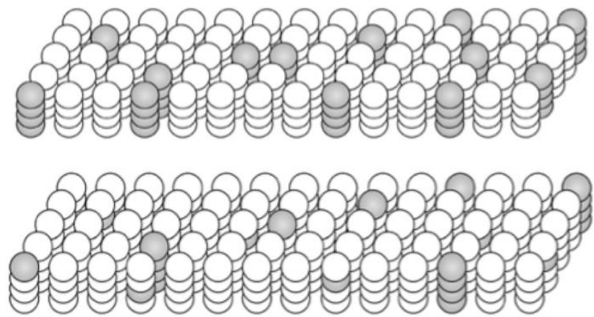

**Figure 10.4.** Example of results after the spatial pooling algorithm (top)
and after the temporal pooling algorithm (bottom)
While the spatial pooling algorithm identifies several columns where all neurons are active, the
temporal memory algorithm leads to the selection of just a few active neurons in the active columns
(Ahmad, Lewis & Lai, 2021)

Within this algorithm, synapses establish connections with dendrite segments, which come in two types: proximal and distal. Proximal dendrite segments form synapses with feedforward inputs, and their active synapses are linearly summed to determine the feedforward activation of a column. In contrast, distal dendrite segments create synapses with cells within the layer. Each cell has multiple distal dendrite segments, and when the sum of the active synapses on a distal segment exceeds a predefined threshold, the associated cell enters the predictive state. The predictive states of the cells are determined through a logical *or* operation across several threshold detectors, considering multiple distal dendrite segments per cell.

Following the spatial pooling phase, the temporal memory algorithm transforms the columnar input representation into a new one that incorporates historical context. This new representation is created by activating a subset of cells within each column, usually selecting only one cell per column.

Here, HTM cells can exist in one of three states: *active* (when influenced by feedforward input), *predictive* (when partially depolarized due to lateral connections with neighboring cells), and *neutral*.

As mentioned above, the temporal memory algorithm picks up where the spatial pooling algorithm leaves off, with active columns representing the feedforward input. Each time step in the algorithm comprises several computations:

- The system receives a set of active columns;
- For each active column, it identifies the cells within the column that have an active distal dendrite segment (i.e., cells that are already in the predictive state from the previous time step) and activates them. If no such cells exist, it activates all the cells in the column. Such a column is said to be "bursting". The set of active cells represents the current input in the context of prior input. This mechanism allows the system to confirm its expectations when the input pattern aligns with predictions. Otherwise, it activates all the cells in the column when the input pattern is unexpected;
- In cases where there is no prior state, and thus no context or prediction, all cells in a column become active when the column is active. This scenario resembles hearing the first note in a song, where without context, it is challenging to predict what will happen next, so all possibilities are kept open. When there is a prior state, but the input does not match the expected pattern, all cells in the active column become active;
- For each active column, the algorithm learns on at least one distal segment. In cases where a column is bursting, it chooses a segment with some active synapses at any permanence level or grows a new segment on the cell with the fewest segments. It increases the permanence on active synapses, decreases it on inactive synapses, and creates new synapses to cells that were previously active;
- Dendrite segments are activated by counting the number of connected synapses corresponding to currently active cells on every dendrite segment in the layer. If the count surpasses a threshold, the dendrite segment is marked as active. These cells with active distal dendrite segments will become predictive cells in the next time step.

The HTM learning algorithms can create different representations for the same instances, depending on the context. Using an example from (Hawkins & Ahmad, 2016), let us consider two input sequences: *ABCD* and *XBCY*. Here, each letter denotes a sparse activation pattern within a group of neuron. After learning these sequences, the network is expected to forecast *D* after *ABC* and *Y* after *XBC*. Hence, the internal representation of *BC* in the two situations must be different. Otherwise, the network cannot correctly predict the next item after *C*.

Figure 10.5 shows how the representations differ in the two situations. Although the representations for *B* and *C* are identical after spatial pooling, the subset of selected neurons after the temporal phase distinguishes the representations based on the sequences that have been presented before. These states are denoted as *B'*, *B''*, *C'*, *C''*, etc. Because *C'* and *C''* are different, they can correctly predict either *D* or *Y*.

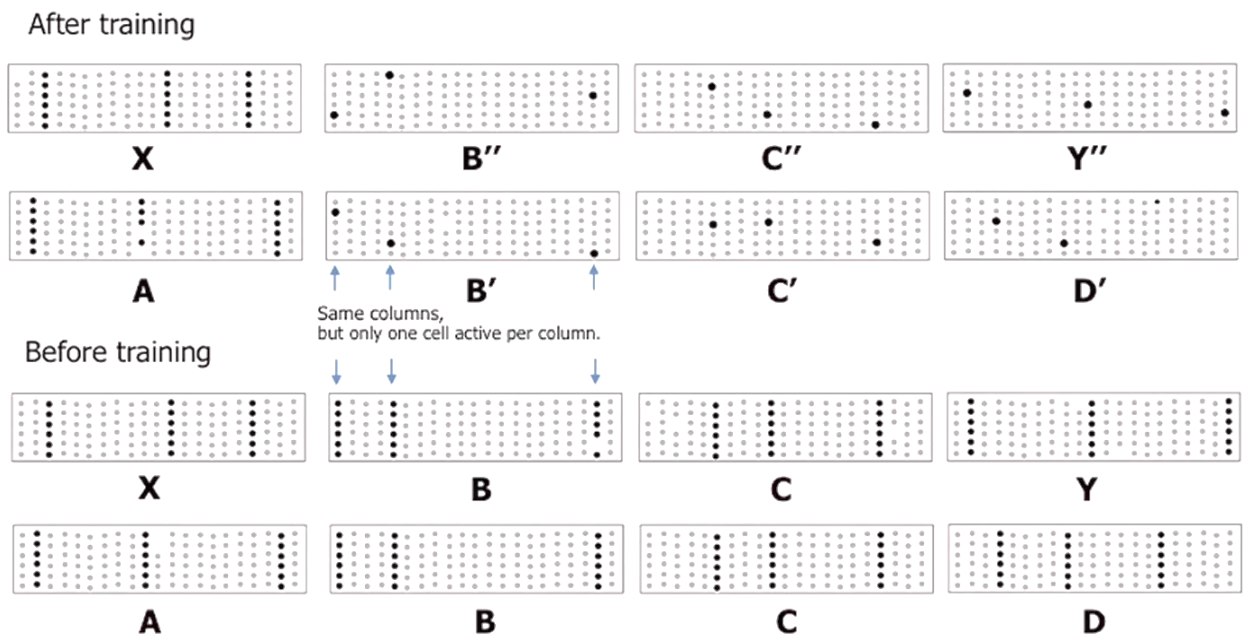

**Figure 10.5.** An example where the same inputs have different representations in different contexts
(Hawkins & Ahmad, 2016)

HTM places a strong emphasis on the temporal aspect of data. It excels at modeling sequences and patterns over time, making it well-suited for tasks involving time series. It can recognize temporal dependencies, the order of events, and anomalies (or outliers). The use of sparse distributed representations, where only a small subset of neurons become active to represent a specific pattern, enhances memory efficiency and robustness, making HTM capable of handling noisy or incomplete data effectively.

HTM is also capable of online learning, meaning it can continuously update its understanding of patterns as new data becomes available. This dynamic learning process makes HTM suitable for real-time applications.

While the name "hierarchical temporal memory" implies a hierarchical structure, it is important to note that such a hierarchy was not actually implemented. The typical descriptions of HTM primarily focus on the two algorithms presented above: spatial and temporal pooling. They are usually discussed in isolation, with no emphasis on the broader hierarchical framework that HTM could potentially create. A possible hierarchical implementation would make use of a structure resembling a tree, with various levels comprising regions. Each level would accommodate multiple regions, and higher levels would feature fewer regions. This hierarchical structure would allow higher-level regions to leverage patterns learned at lower levels, combining them to encode and recall more complex and abstract patterns. However, such an implementation is unlikely, because the HTM model has been superseded by the "thousand brains" model, which will be presented in the next section.

## 10.2. "A Thousand Brains" Theory[6]

In this theory, the neocortex is depicted as generating predictions concerning the sensory input it will receive in the near future from the external world. These predictions are constructed based on a model of the world that the brain has developed through prior experiences. Each column of neurons within the neocortex functions as a mini-brain, autonomously generating predictions. These columnar predictions are subsequently combined through the brain's interconnected network to formulate a collective perception of the world. The term *a thousand brains* alludes to the idea that the neocortex comprises numerous columns of neurons, each of which serves as an independent prediction generator. Each column contains cells that exhibit selectivity to specific patterns or features. For example, in the visual cortex some columns may be specialized in recognizing edges, while others may identify shapes. These specialized predictions are subsequently integrated to yield a comprehensive understanding of the visual scene.

---
[6] This section uses information and ideas from: (Hawkins et al., 2019) and (Hawkins, 2021).

Jeff Howkins (2021) imagines an analogy about five people with different sensory abilities and maps of various towns dropped into an unknown location within one of these towns. They can identify landmarks on their maps, but cannot pinpoint their exact location independently. They each possess maps cut into squares with sensory information about the town: sounds, textures, or images. They start by individually identifying possible towns based on what they perceive. By sharing their lists of potential towns through a voting process, they collectively narrow down their options. If a town appears on everyone's list, they conclude they are in that specific town. The voting mechanism works regardless of the sensory abilities or details of the maps. They only need to agree on the town they all presume. However, they may encounter situations where multiple towns match their collective sensory information. To further refine their location, they share their relative positions to eliminate the towns that do not match the spatial arrangement of identified landmarks. This process of voting among sensory inputs mirrors how the brain's cortical columns merge various sensory inputs into a coherent representation. In addition, the cortical columns may also share information about relative positions, aiding in pinpointing the exact location based on the arrangement of familiar features.

The author *hypothesizes* that the neocortex contains grid cells as well, but they represent the locations of various features relative to objects. He suggests that the mechanisms in the entorhinal cortex and hippocampus used to learn the structure of environments are also used by the neocortex to learn the structure of objects. Therefore, the grid cells in different cortical columns track the location of perceptual features relative to the object being interacted with. Having a representation of location in each cortical column leads to suggestions about how the neocortex represents the compositionality and behaviors of objects. The author proposes that each part of the neocortex learns complete models of objects, and multiple models of the same object are distributed throughout the neocortex.

The concept of *displacement cells* is also introduced, also as a *hypothesis*. Similar to grid cells, displacement cells cannot represent a unique displacement on their own. Instead, they represent relative displacements in a manner analogous to how grid cell modules represent unique locations. The activity of multiple displacement cell modules is used to represent a unique displacement. For example, a single displacement vector can represent the relative position of objects, such as the logo on a coffee cup at a specific location.

Displacement vectors not only represent relative positions, but are also unique to the specific objects being considered. This means that complex objects can be represented by a set of displacement vectors, defining the components of the object and their arrangements relative to one another. The method of representing objects using displacement vectors also allows a hierarchical composition. For example, a logo on a cup can be composed of sub-objects such as letters and graphics. A displacement vector representing the logo implicitly carries information about the sub-objects within it. The concept of displacement vectors also allows for recursive structures. For example, a logo could contain an image of a coffee cup with its own logo. This recursive composition is a fundamental aspect of representing physical objects, language, mathematics, and other forms of intelligent thought.

Sensory processing in the neocortex occurs in two parallel sets of regions known as "what" and "where" pathways. In vision, the "what" pathway is associated with recognizing objects, while the "where" pathway is associated with spatial information, such as reaching for an object. Similar "what" and "where" pathways have been observed in other sensory modalities, suggesting a general principle of cortical organization.

Howkins proposes that a location-based framework for cortical function applies to both "what" and "where" processing. The primary distinction between these regions is that "what" regions represent locations in an allocentric manner (in the location space of objects), while "where" regions represent locations in an egocentric manner (in the location space of the body).

In both "what" and "where" regions, a common operation involves attending to two different locations. Displacement cells determine the movement vector needed to move from the first location to the second location. In a "what" region, this may involve moving a finger from one location on an object to another location on the same object. In contrast, a "where" region calculates how to move from one egocentric location to another within the body, regardless of the presence of objects at those locations.

Thinking is conceptualized as movement through these location spaces.

The author points out that there are many cortical-cortical projections that do not fit the traditional hierarchical processing model. These non-hierarchical connections include long-range projections between regions in the left and right hemispheres and connections between regions in different sensory modalities. He suggests that these connections serve a purpose related to sensor fusion and object recognition. Instead of a single unified model of an object, there are hundreds of models of each object, each based on unique subsets of sensory inputs from different modalities, e.g., vision or touch. Long-range non-hierarchical connections allow these models to quickly reach a consensus on the identity of the perceived object.

The idea that the hierarchical concept representations that appear to be fundamental in cognitive processes may be an emergent effect of non-hierarchical neural representations is an interesting issue, also addressed in the next section.

**10.3. Other Models**

(Pischedda et al., 2017) investigates how the human brain represents and organizes rules used in various cognitive tasks. While some theories suggest that the brain encodes rules at different hierarchical levels within the prefrontal cortex, this research aims to clarify whether rules at distinct levels are indeed represented in separate brain regions. The study relies on functional magnetic resonance imaging (fMRI) to examine whether low-level and high-level rules are encoded differently within the brain. Participants were tasked with applying rule sets containing both lower-level stimulus-response rules and higher-level selection rules.

The findings reveal that there were no significant differences in the brain regions encoding rules from different hierarchical levels. Contrary to some theoretical expectations, it seems that such rules are represented in a similar manner. Hierarchical rules, which encompass both lower-level and higher-level rules, are represented without a significant spatial distinction based on their position in the hierarchy. This means that regardless of whether a rule is a fundamental, low-level rule (dealing with basic sensory stimuli and responses) or a high-level rule (involving more abstract cognitive processes such as decision making and task selection), both types of rules are processed by the same network of brain regions.

While grid cells can represent 2D variables effectively, the question arises regarding their ability to represent higher-dimensional variables. The model presented in (Klukas, Lewis & Fiete, 2020) demonstrates how they can efficiently encode variables with dimensions greater than two while maintaining low-dimensional structure. It achieves this through a combination of low-dimensional random projections and traditional hexagonal grid cell responses.

Grid cells are known for their role in representing spatial information, but recent research suggests they can also represent various cognitive variables beyond spatial location. The firing fields observed in animals such as flying bats or climbing rats can be generated by neurons that combine activity from multiple grid modules. Grid cell responses in 3D environments may not exhibit regular 3D grid patterns, but they do seem to allow localization in all three dimensions. This localization may be influenced by spatial landmarks or formed by conjunctions of grid cells encoding higher-dimensional spaces.

This suggests that grid cells could implement a general circuit capable of generating coding and memory states for high-dimensional variables. The authors propose a coding scheme for high-dimensional variables, considering the structural and dynamical constraints of grid cell responses. They assume that the activity of each grid module remains within a 2D toroidal attractor in neural state space. Modular codes, where neurons are divided into distinct groups for encoding different aspects of a variable, provide an efficient means to represent high-dimensional variables. The multi-module representation of grid cells offers an efficient high-dimensional vector space that can be used for both representing and memorizing arbitrary vectors of higher dimension. The update mechanism of grid cells allows for vector-algebraic operations between stored vectors, essential for vector integration in abstract high-dimensional spaces. The network can represent, store, and perform algebraic operations on abstract vectors of varying dimensions without requiring the reconfiguration of the grid cell network.

(Constantinescu, O'Reilly & Behrens, 2016) explains that the finding of grid-like activity in brain regions associated with both spatial and conceptual tasks raises the possibility of common neural coding mechanisms for storing spatial and conceptual representations. It cites evidence that hippocampal cells encode individual concepts in humans and that rodent grid cells may represent dimensions beyond space, such as time. The study used fMRI to investigate whether humans employ a hexagonally symmetric code when navigating abstract conceptual representations. The task designed for the experiment was analogous to spatial navigation, but involved abstract dimensions.

Subjects learned associations between bird stimuli and Christmas-related symbols. The bird stimuli were unique and varied in two continuous dimensions (neck and legs lengths) but were presented in a one-dimensional (vertical) visual space. This required participants to extract two-dimensional conceptual information from the one-dimensional visual presentation.

Participants watched videos of birds morphing according to specific neck to legs ratios and were instructed to imagine the symbol that would result if the bird continued to morph in the same way. To ensure that the orientation of movement trajectories in bird space was dissociated from visual properties, the trajectories were designed to avoid sharing variance with visual features. In some trials, the subjects had to choose one of three symbols (Figure 10.6).

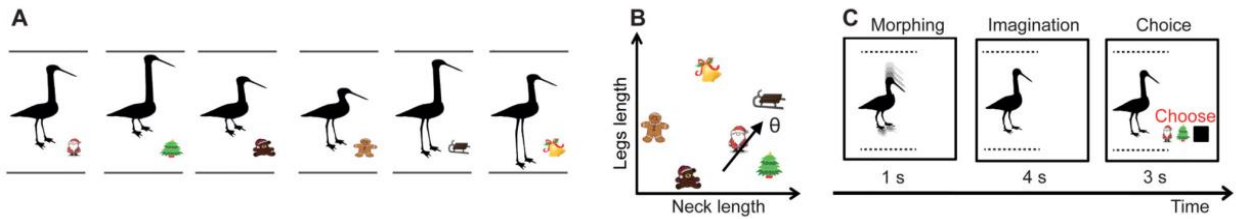

**Figure 10.6.** Experimental design for navigation in abstract space
(Constantinescu, O'Reilly & Behrens, 2016)

The participants achieved a high accuracy rate (about 73%) in predicting symbols based on bird morphs, however, no one reported to conceive the relationships between birds and symbols as a spatial map, i.e., the subjects were not consciously aware that these associations could be organized in a continuous "bird space".

The study found a hexagonal modulation effect in neural activity; this effect was identified across the whole brain and was strongest in specific regions. Therefore, global codes with hexagonal grid-like patterns may be used in the brain to organize non-spatial, conceptual representations.

(Mok & Love, 2019) challenges the idea that conceptual knowledge in the medial temporal lobe (MTL) relies on spatial processing and navigation circuitry, proposing a domain-general learning algorithm as an explanation for findings in both spatial and conceptual domains. Despite superficial dissimilarities in the types of memory supported by the MTL, this paper proposes that the MTL employs a general-purpose algorithm to learn and organize context-relevant information in a useful format, rather than relying on navigation-specific neural circuitry.

Empirical studies with rodents placed in different geometric environments support the model predictions. In this case, the activation of grid cells becomes less grid-like in non-uniform environments, and this decline is particularly noticeable in narrower areas.

A cognitive clustering model (presented in Section 15.7) was trained in a square environment and then transferred to a trapezoid environment. Consequently, the "grid score" of the model (a measure of hexagonal symmetry) decreased in the trapezoid environment, with a more significant decline on the narrow side compared to the wide side of the trapezoid. The clustering model suggests that grid-like representations emerge in spatial navigation tasks due to the relatively uniform distribution of possible inputs in such tasks. In contrast, in conceptual tasks where input sampling is sparser and the overall space is higher-dimensional, representations align more with human conceptual knowledge. Changes in the geometry of the environment, including non-uniformity, are expected to affect clustering and make grid-like patterns less pronounced.

The study underlines that the model, originally proposed for understanding memory and concept learning, also explains spatial navigation phenomena, including place and grid cell-like representations. It suggests that the spatial findings naturally arise as limiting cases of a more general concept learning mechanism. Therefore, grid-like responses should be the exception, not the rule, when encoding abstract spaces. Representational spaces can be high dimensional, with not all dimensions equally relevant, leading to non-grid-like responses in complex environments.

The traditional view of the place cells in the hippocampus as encoding a purely spatial cognitive map is challenged in (Stachenfeld, Botvinick & Gershman, 2017). Instead, the authors propose that the hippocampus primarily encodes a predictive map, which represents expectations about an individual's future state. This means that the firing of place cells is not just related to the current location, but also depends on where the individual is expected to go next.

The predictive function of the hippocampus is formalized within a reinforcement learning (RL) framework, which emphasizes the encoding of expectations and predictions. The *successor representation* (SR) approach is suggested as an intermediary between model-free and model-based learning approaches. It allows for flexible value computation in response to changes in rewards without the computational inefficiency associated with purely model-based methods. The paper suggests that SR can complement model-based planning by extending the range of forward sweeps that prioritize updates. Unlike traditional cognitive maps or model-based RL, this theory posits that a model-free learning approach is used to build a predictive map, which is more adaptable in dynamic environments.

The authors propose that the hippocampus encodes the SR as a rate code across a population of neurons. Each neuron represents a possible future state (e.g., a future spatial position). The firing rate of a neuron encoding a particular state is proportional to the discounted expected number of times that state will be visited under the current policy given the current position. The paper introduces the concept of "SR place fields" or "SR receptive fields", which are regions in the environment where a specific neuron encoding a future state has a high firing rate. In a 2D environment, these SR place fields resemble the traditional circular firing fields of place cells. The firing rate gradually decreases as one moves away from the peak of the field.

Initially, grid cells were thought to represent a Euclidean spatial metric, aiding in path integration (or dead reckoning), i.e., the approximation of the current position by using a previously determined position and estimates of speed, heading, and elapsed time. The proposed predictive map theory suggests that grid fields tend not toward a globally regular grid, but toward a predictive map of the task structure, influenced by global boundaries and the multi-compartment structure. Grid fields become less regular in multi-compartment environments compared to single-compartment rectangular enclosures, because separating barriers between compartments perturb the task topology from an uninterrupted 2D grid.

In (Stoewer et al., 2023), cognitive maps are discussed as representations of memories and experiences and their relationships. These maps are formed and navigated using place and grid cells. The paper introduces the concept of "multi-scale successor representation" as a mathematical principle underlying the computations of place and grid cells. This principle is proposed to be fundamental in how cognitive maps are constructed.

A NN model is presented, which is trained to learn a cognitive map of a semantic space derived from 32 distinct animal species encoded as feature vectors. Through training, the network effectively identifies the similarities among these species and builds a cognitive map of an "animal space".

The NN model implements SRs for a non-spatial navigation task and combines the memory trace theory and cognitive map theory. SR is considered as a way to model the firing patterns of place cells. It involves calculating future reward matrices for states in an environment and using them to build a representation. The developed cognitive map based on SR can be used for navigation through arbitrary cognitive maps and for finding similarities in new inputs and past memories.

The model demonstrates the potential for creating hierarchical cognitive maps with different scales. Fine-grained maps show even distribution of animal vectors in feature space, while coarse-grained maps cluster animals by biological class, e.g., amphibians, mammals, and insects. The model also shows that it can accurately represent completely new or incomplete input by interpolating representations from the cognitive map.

(Whittington et al., 2020) views both spatial and relational memory problems as examples of structural abstraction and generalization. This means that just like different spatial environments share common regularities, allowing for inferences and shortcuts, similar structural regularities enable inferences in non-spatial relational problems. Factorized representations involve separating different aspects of knowledge and recombining them flexibly to represent novel experiences, which is beneficial for learning and making inferences. The authors introduce the *Tolman-Eichenbaum machine* (TEM) as a model that leverages factorization and conjunction to build a relational memory system capable of generalizing structural knowledge in both spatial and non-spatial domains. TEM predicts that although hippocampal remapping may appear random, it actually reflects a preserved structural representation across environments. The authors verify this prediction in place and grid cells, suggesting a unified framework for hippocampal-entorhinal representation, inference, and generalization across various tasks.

The paper describes an unsupervised learning problem involving an agent tasked with predicting the next sensory experience in a sequence derived from probabilistic transitions on graphs. The agent is presented with sequences of sensory observations and information about the relations or actions that caused transitions between adjacent nodes on a graph. Different types of relations exist, such as in a family hierarchy or spatial navigation. When the agent has experienced all possible transitions on a graph, it can store the entire graph in memory and make perfect predictions. However, if the structural properties of the graph are known in advance, perfect predictions can be made even before all transitions have been experienced. This is because understanding the structure allows the agent to infer additional relations and transitions based on known patterns.

For example, in a family hierarchy, knowing the structure allows the agent to make inferences such as "Bob has a daughter, Emily", which immediately leads to other inferences such as "Emily is Alice's granddaughter and Cat's niece", without directly experiencing those transitions. Similarly, in spatial navigation, understanding the structure of 2D graphs enables the agent to place a new node on a coordinate and infer relational information about its connections to other points on the graph.

The problem of sensory prediction is decomposed into two main components: the relational graph structure and the sensory observations. Understanding the relational structure helps in path integration, while relational memories bind sensory observations to locations in the structure. To facilitate generalization, the model separates variables related to abstract location (generalized across maps) from those grounded in sensory experience (specific to a particular map). These variables are represented as populations of units in a neural network.

The primary objective is to learn the NN weights that can represent locations in relational structures and form relational memories. These memories are stored using Hebbian learning and later retrieved. The resulting NN architecture closely mirrors the functional anatomy of the hippocampal formation, since hippocampal representations are formed by combining sensory input and abstract locations. To infer new representations of abstract locations, TEM performs path integration from the previous abstract locations based on the current action/relation. Error accumulation in path integration is corrected using conjunctive representations stored in hippocampal memory. In cases with self-repeating structures, cognitive maps can be organized hierarchically. The model includes multiple parallel streams, each receiving sensory input and having its own representation of abstract locations.

The study highlights the similarity between the TEM learning scheme and the wake-sleep algorithm and Helmholtz machine (Dayan et al., 1995). It suggests that hippocampal replay, which extracts regularities from wake experiences, may involve sampling from a generative model of the environment, as TEM does.

(O'Reilly et al., 2014) reviews the ideas related to the *complementary learning system* (CLS) framework (McClelland, McNaughton & O'Reilly, 1995), which explains why the brain requires two specialized learning and memory systems, and specifies their central properties: the hippocampus as a sparse, pattern-separated system for rapidly learning episodic memories, and the neocortex as a distributed, overlapping system for gradually integrating across episodes to extract latent semantic structure.

The authors consider that catastrophic interference is a consequence of systems using highly overlapping distributed representations, but these can also offer desirable properties such as generalization and inference. As mentioned earlier, catastrophic interference is a phenomenon where, as the network learns new patterns, it adjusts its parameters to accommodate the new data, often overwriting the existing patterns it had learned earlier. Consequently, the network fails to perform well on both tasks simultaneously. Catastrophic interference poses a significant challenge in lifelong learning scenarios where a model must continually adapt to new information without forgetting what it has already learned.

CLS proposes that a structurally distinct system with sparse, non-overlapping representations could complement the highly overlapping system. The hippocampus integrates information from various cortical areas to form a conjunctive representation of events. The hippocampal system employs pattern separation, with very sparse activity levels, to encode new information while preventing interference with existing memories. CLS asserts that the hippocampus encodes information differently from the neocortex to minimize interference, maintaining highly separated representations through sparse activation levels. To achieve separation, the hippocampus employs very sparse activation, such as 0.05%, in contrast to the cortex's roughly 15% activation.

It is believed that the hippocampus replays memories during sleep, allowing the cortex the time to integrate new memories without overwriting older ones. (Winocur, Moscovitch & Bontempi, 2010) proposes a view of consolidation similar to (McClelland, McNaughton & O'Reilly, 1995), emphasizing that the memories consolidated in the neocortex are different from those initially encoded by the hippocampus. The cortex extracts a generalized "gist" representation. Memories are not transferred from the hippocampus to the cortex but rather, the cortex forms its own distributed representation based on hippocampal encoding, capturing similarity structures not initially present. The hippocampus needs to find a balance between encoding (benefitting from pattern separation) and recall (benefitting from pattern completion). The "theta-phase model" suggests that the hippocampus switches between encoding and retrieval modes several times per second, rather than infrequently and strategically. The hippocampus may use these theta-phase dynamics for error-driven learning. It continuously attempts to recall information relevant to the current situation and learns based on the difference between recall and the actual inputs.

(Tomasello et al., 2018) describes a neuro-computational model that simulates semantic learning and symbol grounding in action and perception. The model tries to replicate how the brain learns and associates semantic information with sensory perception and motor actions through the co-activation of neuron populations in different brain areas. The model employs Hebbian learning, resulting in the emergence of distributed cell assembly circuits across various cortical areas. The semantic circuits formed through this learning process exhibit class-specific distributions. For example, circuits related to action words extend into motor areas, while those related to visually descriptive words reach into visual areas. The model identifies certain hub areas with a large number of neurons within the brain, which play an important role in integrating phonological and semantic information. The model aims to explain the existence of both semantic hubs and class-specific brain regions as a consequence of two major factors: neuroanatomical connectivity structure and correlated neuronal activation during language learning.

(Mack, Love & Preston, 2018) discusses the relationship between hippocampal function and concept learning. It highlights the idea that the hippocampus, traditionally associated with memory, also plays an important role in forming and organizing conceptual knowledge. It introduces the *EpCon* (episodes-to-concepts) theoretical model that links episodic memory and concept learning. It suggests that the hippocampus transforms initially-encoded episodic memories into organized conceptual knowledge. While the hippocampus was initially thought to be primarily involved in individual episode encoding and retrieval, recent research suggests a broader role. The hippocampus is involved in building flexible representations that span multiple experiences, are goal-sensitive, and guide decision making. The EpCon model is influenced by the SUSTAIN model of concept learning (Love, Medin & Gureckis, 2004), presented in Section 15.7, which posits that conceptual representations are formed through the interaction of selective attention and memory processes. It comprises several mechanisms that map onto hippocampal functions, including pattern separation and completion, memory integration, and memory-based prediction error. They are guided by attentional biasing. The adaptive nature of EpCon allows for the formation of representations that highlight common features specific to concepts and distinguish between concepts. This process transforms episodic memories into organized conceptual knowledge. The paper also mentions another study (Davis, Love & Preston, 2012) that provides direct evidence for hippocampal involvement in concept formation, suggesting that the hippocampus adapts its representations to capture the nature of new concepts, integrating overlapping experiences for rule-based representations, and using pattern separation to create distinct representations for exceptions.

## 11. Representations in the Brain

### 11.1. Numbers

(Dehaene, 1993) introduces a mathematical theory connecting neuro-biological observations with psychological principles in numerical cognition. The quantity of objects named *numerosity* is represented by the firing patterns of numerosity detectors. Each neuron corresponds to a preferred number, following a Gaussian tuning curve based on the logarithm of the numbers. Decision making relies on Bayesian log-likelihood computations and the accumulation derived from this encoding. The resultant equations accurately model response times and errors in tasks involving number comparison and same-different judgments, aligning closely with behavioral and neural data.

Each quantity is internally represented through a distribution of activations along an internal continuum known as the *mental number line*. A quantity of *n* objects is internally depicted by a Gaussian random variable with a mean $q(n)$ and a standard deviation $w(n)$, which determines the degree of internal variability or noise within the coding system.

$q(n)$ and $w(n)$ obey *Weber's law*, an empirical observation regarding number stimuli, which states that numerosity discrimination depends only on the ratio of the numbers involved, not their absolute values. Assuming that internal variability is the same for all numbers leads to $w(n) = w$. To comply with Weber's law, $q(n) = \log(n)$. Therefore, the probability that a number $n$ is represented by a particular value $x$ of the internal random variable $X$ is:

$$P(X \in [x, x+dx]) = N\left(\frac{x - \log(n)}{w}\right)dx = \frac{e^{-\frac{(x-\log(n))^2}{2w^2}}}{\sqrt{2\pi} \cdot w} dx \qquad (11.1)$$

where $N$ is the normal distribution.

The equation asserts that a given numerical input is represented at different moments by noisy values that tend to cluster around the $\log(n)$ position on the number line.

Each quantity is approximately represented by a specific group of numerosity detector neurons, each attuned to a particular numerosity (Figure 11.1). Considering the capability of people to process a wide range of numbers, a logarithmic arrangement of neural thresholds is plausible. This arrangement implies fewer neurons allocated to larger numerosities, and it aligns with the observation that discerning differences between smaller numbers is easier than distinguishing larger quantities. As numbers increase, the precision of discrimination decreases.

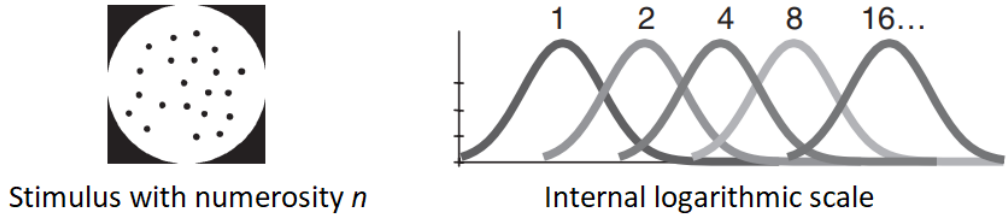

**Figure 11.1.** Coding with numerosity detectors
Adapter after (Dehaene, 1993)

In the Dehaene-Changeux model (1993), the firing rate of a detector neuron that responds preferentially to numerosity $p$, in response stimulus $n$, is:

$$f(n, p) = \alpha \cdot N\left(\frac{\log(n) - \log(p)}{w'}\right) = \alpha \cdot \frac{e^{-\frac{(\log(n)-\log(p))^2}{2w'^2}}}{\sqrt{2\pi} \cdot w'} \qquad (11.2)$$

where $w'$ is the *neural Weber fraction* which defines the degree of coarseness with which neurons encode numerosity and may be different from the *internal Weber fraction w*. The psychological representation based on $w$ can result from averaging the responses of a neural population with the biological parameters $w'$.

An alternative, slightly simpler way of computing the activation of a numerosity detector is based on relative spiking rates (Prather, 2012):

$$f(x) = e^{-\frac{x^2}{2s^2}} \qquad (11.3)$$

$$x = \log_{10} p - \log_{10} n \qquad (11.4)$$

Direct neural recordings identified two types of neural coding related to numbers: *number-selective coding* and *summation coding*. The former was presented above. The latter is related to an accumulator model of number representation, where numerosities are represented by accumulating a number of pulses serially generated, i.e., the number of spikes increases with the numerosity (Meck & Church, 1983).

Psychological studies involving mapping number values to spatial representations, like the number line, showed that small children typically provide logarithmic estimations, while older children and adults tend to produce linear estimations. This may be caused by the narrowing of the tuning functions of neurons which are essential for the accurate coding of numbers. Also, the Weber fraction, which represents the smallest proportional difference that can be distinguished, changes with age, indicating a modification in the underlying tuning functions (Siegler & Booth, 2004).

(Kutter et al., 2022) demonstrates the existence of abstract and notation-independent codes for addition and subtraction within neuronal populations in the medial temporal lobe, which suggests that the brain has the capacity to perform arithmetic operations regardless of the specific notation or symbols used. This research involved recording single-neuron activity while human participants performed addition and subtraction tasks. The study used various notations and visual displays to control for non-numerical factors and found that participants performed these tasks with high accuracy (Figure 11.2). The authors identify rule-selective neurons in the MTL that selectively respond to either addition or subtraction instructions. These neurons are believed to encode arithmetic rules and exhibit a degree of specialization in responding to specific quantitative rules applied to different magnitudes.

The authors identify distinct coding patterns in different MTL regions, i.e., they may have different cognitive functions in arithmetic processing. Static and dynamic codes suggest different cognitive processes at play during arithmetic tasks. Static codes, observed in the hippocampus, involve persistently rule-selective neurons, so this region may be responsible for the actual computation of arithmetic operations, such as adding or subtracting numbers. This relates with the concept of working memory, where information is manipulated and calculated. On the other hand, dynamic codes, seen in the parahippocampal cortex, involve rapidly changing rule information related to short-term memory, possibly holding the arithmetic rules temporarily for calculation.

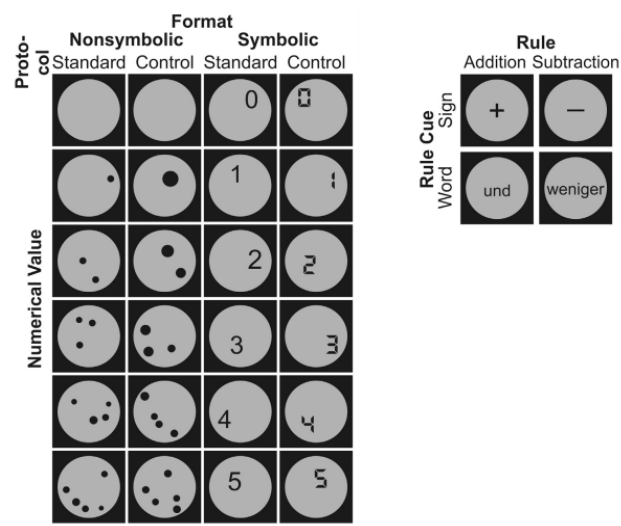

**Figure 11.2.** Examples of input stimuli and tasks representations
The operands can be presented either as digits with various shapes or as a number of points. The operations can be presented either in the standard mathematical notation (+, –) or as written words (add/und, subtract/weniger). The participants need to assess whether the result of an operation is correct or not.
Adapted after (Kutter et al., 2022)

In (Cope et al., 2018), the authors address the question of whether simpler animals such as honey bees possess the cognitive capacity to learn abstract concepts like sameness and difference. These abilities are typically associated with higher-order cognitive functions and believed to rely on complex neural processes in the mammalian neocortex. However, the paper presents a novel neural network model that demonstrates that honey bees can indeed learn these abstract concepts using relatively simple neural structures in their brains.

The model is based on the known neural connections and properties of the honey bee mushroom body, a brain structure involved in sensory processing and learning. It successfully replicates the performance of the bees in various associative learning tasks, including those concerning sameness and difference. This finding challenges the assumption that advanced neural mechanisms are required for abstract concept learning and suggests that the honey bee brain, despite its small size and simplicity compared to mammals, can perform such operations.

## 11.2. Decisions

The *leaky, competing accumulator* (LCA) model (Usher & McClelland, 2001) is a detailed computational model aimed at explaining the progress of choices in time. This model is an extension of classical accumulator models. It is based on the idea that decision making is a gradual process. It assumes that information accumulates over time and that a choice is made based on the relative accumulation of evidence for different response alternatives.

LCA includes randomness; at each time step, the evidence accumulation process is subject to random fluctuations. This stochastic element allows for variability in both decision outcomes and response times. An important feature of the model is the introduction of leakage, also known as decay. This means that over time, the accumulated evidence for each response alternative gradually decreases. The leakage introduces a degree of imperfection into the decision process, as it reflects the idea that evidence may not be perfectly retained.

It also incorporates the principle of competition among response alternatives. As evidence accumulates, response alternatives compete with one another, and the choice is made based on which alternative accumulates the most evidence over time. This competition is achieved through a process of lateral inhibition, which means that the activation of one response alternative inhibits the activation of others.

The model can be represented as a two-layer network. It consists of input units, indicating the external input to the network, and accumulator units, each corresponding to a response alternative. These accumulator units are analogous to populations of neurons and have activation and output variables. A simple nonlinear function (the threshold-linear function) is used, which maps the activations of accumulator units to their outputs. This function is used to determine when a response is triggered and which response alternative is chosen.

The evolution of the activation of units representing the choices is given by the following equations:

$$da_i = \left( I_i - \lambda a_i - \beta \sum_{j \neq i} a_j \right) \frac{dt}{\tau} + \xi_i \sqrt{\frac{dt}{\tau}} \qquad (11.5)$$

$$a_i = \max(a_i, 0) \qquad (11.6)$$

In equation (11.5), $a_i$ is the activation of an accumulator unit, $I_i$ is the input of the accumulator, $\lambda$ is the decay rate that accounts for the leakage of the accumulator, $\beta$ is the weight of the inhibitory influences of the other units, $\tau$ is a convenient time scale, and $\xi_i$ is a Gaussian noise term with zero mean and specified variance.

An example of the effect of these equations is presented in Figure 11.3.

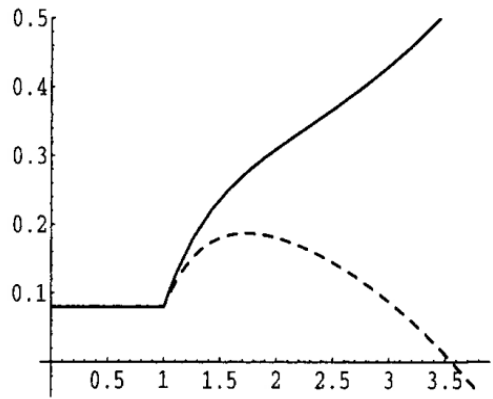

**Figure 11.3.** The dynamics of the accumulator units of two choices
(Usher & McClelland, 2001)

The LCA model also accounts for the delay in processing information from the sensory input to the explicit response. These delays are treated as fixed parameters, and they explain the initial flat portion of reaction time curves observed in experiments.

The model is not limited to specific types of perceptual tasks. It has been applied to a wide range of choice tasks and has shown its utility in explaining various empirical phenomena related to decision making.

(Fang, Cohen & Kincaid, 2010) explores a specific category of dynamical neural networks characterized by lateral inhibition and WTA behavior. The study reveals the existence of WTA behavior in a broad class of competitive NNs and establishes sufficient conditions for achieving WTA equilibrium. Furthermore, rigorous convergence analysis is conducted. The identified conditions for WTA behavior serve as practical guidelines for designing such networks. Once the network enters the WTA region, it rapidly converges to the WTA point. This feature streamlines decision making processes since the winner can be declared as soon as the network enters the WTA region. Additionally, the paper introduces the concept of a self-resetting NN, enabling the network to return to its initial state when not in use, and become ready for new inputs.

**11.3. Actions**

(Tan et al., 2013) discusses research on a part of the temporal lobe of a macaque, a brain area involved in processing visual information related to actions and actors. The authors aim to understand how individual neurons represent actions and actors. They uses a simple encoding called the *snippet-matching* model that assumes that each neuron compares incoming visual input over a single time step of approximately 120 ms to its preferred stimulus, which is a short segment of a specific action.

Contrary to the expectation of finding distinct clusters of neurons specialized for coding actor-invariant or action-invariant information, the study does not identify such clusters; the results suggest that the neural representation scheme is more continuous and distributed. Instead of discrete groups of neurons with specific functions, the visual system appears to adopt a more universal and generalized approach to represent actions and actors. The model uses linear weights to determine the similarity between the incoming input and the preferred stimulus, i.e., the authors test the ability of a linear weighted sum of responses to reproduce the neural response patterns in the neurons, and this simple model provides surprisingly good results.

(Vaidya & Badre, 2022) discusses abstract task representations in two distinct brain networks: the fronto-parietal (FP) network and a network involving the medial temporal lobes (MTL), medial prefrontal cortex, and orbitofrontal cortex (OMPFC). The MTL-OMPFC network is associated with encoding relationships between objects, contextual dependencies, and abstract task information. The FP network is linked to rapid reformatting of task information for cognitive control and action selection.

The MTL-OMPFC network maintains task knowledge in a cognitive map format, enabling evaluation of an individual's position in an abstract task space. In contrast, the FP cortex formats abstract task knowledge as production rules, facilitating action selection.

The knowledge-action dissociation phenomenon is mentioned, where frontal lobe damage can impair the implementation of actions based on abstract rules, despite an intact understanding of the rules. In contrast, MTL damage does not significantly affect performance on cognitive control tasks. Behavioral studies on skill learning tasks suggest that abstract task knowledge can evolve from comparative searches through the declarative memory to more efficient abstract productions, speeding up task implementation.

The FP network is engaged at the start of novel tasks, during task instructions, and when new rules are added to existing tasks, suggesting a rapid reformatting of task information. The MTL-OMPFC network is associated with making inferences and planning actions based on representations of task space. Computational models propose that the entorhinal cortex and hippocampus are involved in building generalizable knowledge that aids learning and inference. The two networks fulfill different roles based on cognitive and behavioral demands. On the one hand, the MTL-OMPFC network focuses on discovering task structure from experience, inferring latent task states and abstract relations. On the other hand, the FP network maintains abstract task representations as state-action contingencies for cognitive control and guided action selection.

## 12. Vector Symbolic Architectures (Hyperdimensional Computing)

*Symbolic* representations represent objects or concepts using symbols. They have a combinatorial structure allowing the creation of a virtually infinite number of expressions, where complex representations are composed of simpler ones. However, their biological implementation is uncertain. *Connectionist* representations encompass NNs and brain-like representations and include two main types. *Localist* representations use a single element for each object and are equivalent to the concept of "grandmother cells" (Quiroga et al., 2005), for which some evidence has been found indeed in the brain. However, the large majority of brain representations do not rely on this form, but rather as *distributed* representations which model information as distributed across many neurons. In its mathematical equivalent, they use vector representations, where each object is represented by a subset of vector components. They offer high representational capacity, direct access to object representation, can work effectively in the presence of noise and uncertainty, and are more neuro-biologically plausible.

One of the challenges especially for classic connectionist representations is the *superposition catastrophe*. For example, let us consider four neurons that are activated as follows: the first in the presence of squares, the second in the presents of circles, the third in the presence of red objects, and the fourth in the presence of blue objects. These neurons would fail to distinguish the simultaneous presentation of a red square and a blue circle from the simultaneous presentation of a blue square and a red circle, because all four units would be activated in both cases. This problem also prevents the representation of hierarchical compositional structures.

The field of *vector symbolic architectures* (VSA), also known as *hyperdimensional computing* (HDC), aims to combine the advantages of distributed and symbolic representations while avoiding their drawbacks. VSAs are connectionist models that can directly implement functions specific to symbolic processing (Kleyko et al., 2023a).

Key characteristics of VSA/HDC include (Thomas, Dasgupta & Rosing, 2021): a single, static mapping from input data to high-dimensional neural representations, the fact that all computations are performed in the high-dimensional space using simple operations such as element-wise additions and products, and the fact that the mapping may be random, and thus the individual elements of the representations have low precision, often taking binary values.

## 12.1. Fundamental Operations

VSAs usually implement two main operations, whose mathematical details distinguish the various representations that have been proposed.

*Superposition* (or bundling) combines multiple *hypervectors* (HVs) into a single HV. It models the simultaneous activation of neural patterns, typically as the disjunction of binary HVs or the addition of real-valued HVs. However, the superposition operation alone can lead to the superposition catastrophe, where information about combinations of initial objects is lost.

*Binding* is the other fundamental operation that combines two HVs in a recoverable way.

In the following sections, we will present several VSAs defined by specific implementations of these operations.

## 12.2. Representation Methods

### 12.2.1. Tensor Product Representation

The *tensor product representation* (TPR) is one of the earliest VSA models (Smolensky, 1990). It uses atomic HVs selected randomly from the Euclidean unit sphere (in general, HVs can be real-valued). Superposition is implemented by tensor addition. The binding operation is a tensor product, resulting in exponential growth in dimensionality as more HVs are bound.

Let us consider an example of using the TPR. We will use four concepts: circle ($C$), square ($S$), red ($R$), and blue ($B$), each represented by a 4D vector:

$$C = \begin{pmatrix} 1 & 0 & 0 & 0 \end{pmatrix}^T$$
$$S = \begin{pmatrix} 0 & 1 & 0 & 0 \end{pmatrix}^T$$

$$R = \begin{pmatrix} 0 & 0 & 1 & 0 \end{pmatrix}^T$$

$$B = \begin{pmatrix} 0 & 0 & 0 & 1 \end{pmatrix}^T$$

These base vectors must be orthogonal. In our example, only one bit of 1 is used to distinguish between the concepts, but in general, in high dimensions, more bits of 1 are used; the representation does not assume a one-hot encoding. An alternative representation can use element values of 1 and –1 and rely on Hadamard matrices, where rows are mutually orthogonal, for example:

$$\mathbf{H}_4 = \begin{pmatrix} 1 & 1 & 1 & 1 \\ 1 & -1 & 1 & -1 \\ 1 & 1 & -1 & -1 \\ 1 & -1 & -1 & 1 \end{pmatrix}$$

In the following, we will use our simple representation proposed in the beginning.
We can now define "a circle and a square" using vector addition:

$$C + S = \begin{pmatrix} 1 & 1 & 0 & 0 \end{pmatrix}^T$$

We can represent composite concepts using binding, which is vector multiplication. For example, "a red circle" is represented as follows (red is the column, and circle is the row):

$$R * C = R \cdot C^T = \begin{pmatrix} 0 & 0 & 0 & 0 \\ 0 & 0 & 0 & 0 \\ 1 & 0 & 0 & 0 \\ 0 & 0 & 0 & 0 \end{pmatrix}$$

We can represent "a red circle and a blue square":

$$B * S = \begin{pmatrix} 0 & 0 & 0 & 0 \\ 0 & 0 & 0 & 0 \\ 0 & 0 & 0 & 0 \\ 0 & 1 & 0 & 0 \end{pmatrix}$$

$$M = R * C + B * S = \begin{pmatrix} 0 & 0 & 0 & 0 \\ 0 & 0 & 0 & 0 \\ 1 & 0 & 0 & 0 \\ 0 & 1 & 0 & 0 \end{pmatrix}$$

The interesting part is that we can now ask questions using this representation.

For example: "What color is the circle?" Answer: "red".

$$c_C = M \cdot C = \begin{pmatrix} 0 & 0 & 1 & 0 \end{pmatrix}^T = R$$

Or: "What shape is the blue object?" Answer: "square".

$$s_B = B^T \cdot M = \begin{pmatrix} 0 & 0 & 1 & 0 \end{pmatrix} = S^T$$

The TPR can be used to express role-filler representations. The roles are the categories or slots that define the aspects of the representation. We can extend our example to have roles such as *Color* and *Shape*. The fillers are the specific pieces of information that occupy those roles: *Red* can be a filler for the *Color* role, and *Circle* can be a filler for the *Shape* role. In this case, we would express the red color as *Color * Red*.

### 12.2.2. Holographic Reduced Representation

The *holographic reduced representation* (HRR) (Plate, 1991) is inspired by the TPR. The elements of the HVs are real-valued and generated from a normal distribution with mean 0 and variance $1/D$, where $D$ is the number of dimensions. For a large $D$, the Euclidean norm is close to 1. Binding is achieved through *circular convolution*, which preserves unit norms but results in HVs that are not similar to the input HVs. Unbinding in HRR involves *circular correlation* and may require a *clean-up* procedure.

The HRR is used in the semantic pointer architecture (SPA) (Eliasmith, 2015) that will be presented in Section 16.4.

The circular convolution of two HVs **x** and **y** is defined as follows:

$$\mathbf{z} = \mathbf{x} * \mathbf{y} \tag{12.1}$$

$$z_i = \sum_{j=0}^{D-1} x_j \cdot y_{(i-j) \bmod D} \tag{12.2}$$

Unlike in the TPR, binding in the HRR creates a vector with the same length as its input vectors, and this increases the consistency of the representation.

A computationally efficient algorithm for computing the circular convolution of two vectors takes advantage of the *discrete Fourier transform* (DFT) and *inverse discrete Fourier transform* (IDFT). In general, Fourier transforms are closely related to convolution because the Fourier transform of a convolution operation is equivalent to a multiplication in the frequency domain. Therefore, circular convolution can be expressed in terms of DFT and IDFT as follows:

$$\mathbf{x} * \mathbf{y} = IDFT\left(DFT(\mathbf{x}) \cdot DFT(\mathbf{y})\right) \tag{12.3}$$

where "·" indicates element-wise multiplication.

In SPA, these operations are implemented efficiently by means of matrix multiplications with matrices that can be pre-computed for a certain $D$.

Unbinding an HV can be done with an operation analogous to a matrix inversion, which can be further simplified using a permutation as follows:

$$\mathbf{x} = \mathbf{z} * \mathbf{y}' \quad (12.4)$$

$$\mathbf{y}' = \begin{pmatrix} y_0 & y_{D-1} & y_{D-2} & \cdots & y_1 \end{pmatrix} \quad (12.5)$$

However, this operation is approximate, therefore the resulting $\mathbf{x}$ vector needs to be compared to the base vectors in order to identify the closest match, which corresponds to the actual result.

Let us consider the same example with shapes and colors, but this time in the HRR:

$$C = \begin{pmatrix} 1.30 & -0.27 & -0.85 & 0.33 & -0.57 \end{pmatrix}^T$$

$$S = \begin{pmatrix} -0.36 & 0.06 & 0.43 & 0.34 & -0.52 \end{pmatrix}^T$$

$$R = \begin{pmatrix} 0.76 & 1.25 & -0.21 & -0.21 & 1.00 \end{pmatrix}^T$$

$$B = \begin{pmatrix} -2.52 & 1.84 & 1.76 & 1.29 & 0.31 \end{pmatrix}^T$$

Using the operations of superposition by addition and binding by circular convolution implemented in a public library (Tulkens, 2019), we compute the representation of "a red circle and a blue square":

$$R * C + B * S = \begin{pmatrix} 1.23 & -0.56 & -3.01 & -2.19 & 4.22 \end{pmatrix}$$

If we unbind *Circle* to obtain its color, the result is:

$$\tilde{R} = \begin{pmatrix} 1.17 & 2.64 & -6.20 & -3.50 & 5.91 \end{pmatrix}$$

Then we compute the cosine similarity $S_c$ between $\tilde{R}$ and the base vectors (whose set is also called the *codebook*):

$$S_c(\mathbf{x}, \mathbf{y}) = \frac{\mathbf{x} \cdot \mathbf{y}}{\|\mathbf{x}\| \cdot \|\mathbf{y}\|} = \frac{\sum x_i y_i}{\sqrt{x_i^2} \cdot \sqrt{y_i^2}} \quad (12.6)$$

and get: $S_c(\tilde{R}, C) = 0.09$, $S_c(\tilde{R}, S) = -0.88$, $S_c(\tilde{R}, R) = 0.70$, and $S_c(\tilde{R}, B) = -0.32$. The maximum similarity is obtained for the *Red* vector, and therefore this is the result after the clean-up procedure.

### 12.2.3. Other Representations

The *Fourier holographic reduced representation* (FHRR) (Plate, 1994), also called *frequency domain holographic reduced representation*, is similar to the HRR but uses complex numbers with unit magnitude for the elements of the HVs. Superposition in the FHRR is component-wise complex addition with optional magnitude normalization. The binding operation is component-wise complex multiplication (the Hadamard product) and unbinding is implemented through binding with an HV conjugate (component-wise angle subtraction modulo $2\pi$).

*Multiply, add, permute* (MAP) (Gayler, 1998) uses real or integer elements in the HVs. Superposition is performed by element-wise addition, while binding and unbinding are performed by element-wise multiplication.

*Binary spatter code* (BSC) (Kanerva, 1996) uses binary elements, superposition is performed by element-wise addition with a limiting threshold of 1, while binding and unbinding are performed by the logical *xor* operation.

Several other representations have been proposed, and the presented ones also have multiple variants.

## 12.3. Analogical Reasoning

An example of analogical reasoning using the concepts of role and fillers in VSAs is proposed by Pentti Kanerva (2009), in the form of answering a question such as "What is the Dollar of Mexico?", i.e., its currency.

This problem can be solved in the following way. The roles of *Country* (C) and *Monetary unit* (M) are encoded as HVs (e.g., with 10,000 elements). The possible filler values, *USA* (U), *Mexico* (E), *Dollar* (D), *Peso* (P) are encoded in a similar way.

A "holistic record" for the information about the USA is:

$$A = C * U + M * D$$

and for Mexico it is:

$$B = C * E + M * P$$

We can find the role played by the Dollar by unbinding: $D * A \approx M$. The result is approximate because only the $M * D$ part of $A$ yields a meaningful result, close to a vector in the codebook, while the $C * U$ part yields noise. A similar operation on $B$ ($D * B$) would also yield noise. Then, we can find the currency of Mexico by unbinding $M * E \approx P$. Therefore, the general formula for finding "the Dollar of Mexico" is:

$$(D * A) * B \approx P.$$

## 12.4. Representing Compositional Structures

Compositional structures are formed from objects, which can be either atomic or compositional. Atomic objects are the basic elements of a compositional structure. More complex compositional objects are constructed from atomic elements and simpler compositional objects. This construction resembles a part-whole hierarchy, where lower-level parts are combined to create higher-level entities. In VSA, compositional structures are transformed into HVs using the HVs of their constituent elements. The superposition and binding operations are applied in this transformation process. The goal is to represent similar compositional structures with similar HVs, and to recover the original representations when needed.

*12.4.1. Representing Symbols*

To transform symbols into HVs, independent and identically distributed (*i.i.d.*) random HVs are often used. The resulting HVs are equivalent to symbolic representations as they behave like symbols: they have maximal similarity with their copies and minimal similarity with other *i.i.d.* random HVs.

*12.3.2. Representing Numeric Values*

Numeric scalars and vectors are commonly encountered in various tasks, and in VSA, it is essential to represent them while preserving similarity between close values and dissimilarity between distant ones. There are three main approaches for transforming numeric vectors into HVs (Kleyko et al., 2023a), mentioned as follows.

In the *compositional approach*, close values of a scalar are represented by similar HVs, with similarity decreasing as the scalar values differ more. Typically, scalars are first normalized into a specified range (e.g., [0, 1]) and then quantized into finite grades or levels. Correlated HVs are generated to represent a limited number of scalar grades, often up to several dozen. Various schemes are used to generate these HVs, including *encoding by concatenation* and *subtractive-additive*. Schemes such as *fractional power encoding* allow for direct exponentiation of complex-valued HVs to represent scalars, without the need for normalization or quantization. HVs for different scalar components are combined using superposition or multiplicative binding to form the compositional HV representing the numeric vector.

The *receptive field approach*, known as *coarse coding*, represents numeric vectors through receptive fields activated by the vector. Various schemes, such as *cerebellar model articulation controller*, *Prager codes*, and *random subspace codes* use randomly placed and sized hyper-rectangles as receptive fields. These approaches can produce binary HVs or real-valued HVs, using for example radial basis functions (RBF). They form a similarity function between the numeric input vectors and these receptive fields.

The *random projection approach* (RP) forms an HV for a numeric vector by multiplying it by an RP matrix. It allows for dimensionality reduction when smaller-dimensional vectors are produced. RP matrices can consist of components from the normal distribution or bipolar and ternary matrices. Depending on the application, results can be, e.g., binarized, to produce sparse HVs. RP matrices can also be used to expand the dimensionality of the original vector. Multiple RP matrices can be employed to contribute to the resultant HVs.

*12.3.3. Representing Sequences*

Sequences can be represented by using the entire previous sequence as a context, which allows one to store repeated elements (Plate, 1995; Eisape et al., 2020), for example:

$$s_{abc} = a + a * b + a * b * c$$

$$s_{de} = d + d * e$$

$$s_{abcde} = s_{abc} + s_{abc} * s_{de}$$

This idea can be used for linguistic representations using roles and fillers, e.g., "The boy saw a dog run":

$$s = r_{action} * f_{saw} + r_{agent} * f_{boy} + r_{object} * (r_{action} * f_{run} + r_{agent} * f_{dog})$$

An alternative is to use a set of fixed HVs to represent each position in the sequence:

$$s_{abc} = p_1 * a + p_2 * b + p_3 * c$$

### 12.3.4. Representing Graphs

In case of graphs, a simple method is to assigned random HVs to each node, and to represent the edges as the binding of HVs of connected nodes. The whole graph is then represented as the superposition of the HVs of all the edges (Kleyko et al., 2023a).

## 12.4. VSA Reviews

(Kleyko et al., 2023a) offers an in-depth review of VSA models and focuses on computational models and transformations of input data types to high-dimensional distributed representations. (Kleyko et al., 2023b) extends the analysis to applications, cognitive computing, architectures, and future directions, offering a holistic view of the scope of VSAs. It delves into applications primarily within machine learning/artificial intelligence while also encompassing diverse application domains, in order to emphasize the potential of VSAs.

(Schlegel, Neubert & Protzel, 2022) provides an overview of various implemented operators, and compares VSAs in terms of bundle capacity, unbinding approximation quality, and impact of combining binding and bundling operations on query answering performance. It evaluates VSAs in visual and language recognition tasks, revealing performance variations based on architecture choices.

# 13. The Neural Binding Problem

## 13.1. Variants of the Neural Binding Problem

(Feldman, 2013) posits that the *neural binding problem* (NBP) comprises several distinct problems, presented as follows.

*General coordination* refers to the challenge of understanding how the brain processes information and perceives unity in objects and activities that occur simultaneously. This binding of perceptual elements relies on attention, either through explicit fixation or implicit activation, which plays an important role in determining what is bound together, noticed, and remembered. Temporal synchrony is a central theme within the NBP, encompassing neural firing, adaptation, and the coordination of different neural circuits. It involves timing considerations (binding by synchrony) and the oscillations in neural signals, especially phase coupling, are crucial components of this synchronization.

*Visual feature binding* focuses on how the brain combines different visual features such as color, shape, size, texture, and motion to perceive objects as coherent wholes. The central question is why people do not confuse objects with similar features, such as a red circle and a blue square, with others such as a blue circle and a red square (the example mentioned in Section 12). The visual system organization in space and time plays an important role in feature binding. Most detailed feature binding occurs during foveal vision, where fixations are inherently coordinated in both space and time. Attention is also a key factor, as it aids in the binding of visual features. Various experiments such as brief presentations, masking, and binocular rivalry often reveal limitations in feature binding under stressed conditions, shedding light on the mechanisms involved. Additionally, the brain may employ multiple smaller combinations of features rather than a single unified representation for feature binding.

The *subjective unity of perception* raises questions about how the brain integrates different visual features, despite their processing by distinct neural circuits. There is a clear contrast between people's subjective experience of a stable, detailed visual world and the absence of a corresponding neural representation for this experience. This discrepancy is referred to as the "explanatory gap" or "the hard problem" in neuroscience.

## 13.2. Variable Binding

Our focus in this review is the neural realization of *variable binding*. This is a process where variables in language or abstract reasoning are linked to specific values or entities for comprehension. For example, in the sentence "he gave it to her before", four out of six words are variables that need to be bound to values to understand the sentence. The challenge in variable binding is that there can be a virtually infinite number of items that could be bound to the variables, making traditional methods inadequate.

One proposed mechanism for variable binding is *temporal phase synchrony*, which relies on synchronizing the timing of neural firing to create bindings. This approach divides neural firing into discrete time slices, where the coincidence of firing represents a binding between variables. This mechanism is used, e.g., by the SHRUTI model (Ajjanagadde & Shastri, 1991; Shastri, 1999), presented in Section 7.

An alternative approach involves the use of *signature propagation* (Browne & Sun, 1999). In this model, each variable in an expression has its own node or group of neurons that represent and transmit a specific signature corresponding to a concept. These signatures essentially act as names for the concepts. However, one of the main challenges with this approach is that it would require a potentially large number of signatures to represent all possible concepts, and it is unclear how the brain could generate and manage such a vast number of signatures.

Another model introduces a *central binding structure* that controls binding (Barrett, Feldman & MacDermed, 2008). This structure allows for the temporary linking of nodes or neurons between different concepts, enabling the system to keep track of specific bindings, even as time slices or signatures spread through the network. The central binding also permits more complex operations, such as the unification of signatures representing the same variable.

Some researchers explored *multiplicative techniques* for dynamic variable binding (Hummel, 2011). These approaches involve using distributed representations where various attributes of concepts are combined in a multiplicative manner, and this allows for flexible dynamic binding of variables.

Another approach involves *crossbar networks* for variable binding (van der Velde & de Kamps, 2006), where the connections between computational nodes can be temporarily enabled or disabled to allow signals to travel between nodes for a specific period of time. This approach attempts to solve the binding problem by creating temporary links between nodes, enabling dynamic variable binding.

(Greff, van Steenkiste & Schmidhuber, 2020) is a review of issues related to the variable binding problem. It proposes a framework that tackles the binding problem and emphasizes the need to create meaningful entities from unstructured sensory inputs, maintain separate representations, and construct inferences, predictions, and behaviors using these entities. This framework draws inspiration from neuroscience and cognitive psychology, aligning neural network research with insights from human cognition. The paper discusses the inability of NNs to dynamically and flexibly combine distributed information – a capability necessary for effective symbol-like entity formation, representation, and relational understanding. The authors propose addressing the binding problem through three key facets: representation, segregation, and composition.

The *representation* involves binding information from distinct symbol-like entities at a representational level. The object representations act as fundamental building blocks for symbolic behavior within neural processing. They should combine neural efficiency with symbol compositionality, encoding information in a manner that retains both the expressiveness of distributed, feature-based internal structures and the integrity of self-contained objects. Object representations encompass various forms of entities, including visual, auditory, abstract, and conceptual elements.

The *segregation* problem concerns the structuring of sensory information into meaningful entities, including dynamically creating object representations. This should enable NNs to acquire context-dependent notions of objects, often in a mostly unsupervised manner. The high variability of objects makes the segregation problem a complex task, but it is essential for successful symbolic information processing.

The *composition* problem revolves around using object representations to construct structured models that support inference, prediction, and planning. These models should use the modularity of objects to achieve systematic, human-like generalization. This requires a flexible mechanism that allows NNs to quickly restructure their information flow to adapt to specific contexts or tasks.

(Do & Hasselmo, 2021) discusses various approaches to understanding how the brain represents and binds different components to form complex structures. The central concept explored in this framework is *conjunctive coding*, where roles and fillers are represented by separate vectors of activity, and their binding is achieved through a weight matrix that combines them.

However, some challenges with conjunctive coding are also discussed. One issue is its potential to fail in preserving the independence of roles and fillers. For instance, in sentences such as "Alice loves Bob" and "Bob loves Alice", conjunctive coding would create separate and dissimilar representations for Alice as the lover and Alice as the beloved, even though they are essentially the same entity. This lack of role-filler independence may hinder the ability to generalize across different contexts.

To address this challenge, the paper introduces the concept of *dynamic binding*. This approach involves creating distinct links from a single node representing Alice to her different roles, with the ability to rapidly create or destroy these links as the context changes. Dynamic binding relies on the temporal synchrony of spiking activity in neurons to represent the relationships between roles and fillers. It allows for role-filler independence and can accommodate various bindings over time.

Temporal synchrony has limitations in cases where units must act simultaneously as both a role and a filler. To overcome this, the idea of *temporal asynchrony* is introduced, where the order of firing preserves causality. This concept is particularly valuable for representing complex structures and hierarchies, such as those found in human language, where elements may need to be combined and recombined in various ways to convey different meanings or nuances. Temporal asynchrony also offers a solution when units need to act simultaneously as both a role and a filler, preserving binding directionality.

(Hayworth, 2012) focuses on the binding problem for visual perception in the context of neuroscience. It introduces a new NN model called the *dynamically partitionable autoassociative network* (DPAAN), which provides a solution that does not rely on precise temporal synchrony among neurons, making it more compatible with established neural models of learning, memory, and pattern recognition.

It builds on the *anatomical binding hypothesis*, which associates symbols with specific patterns of neural firing. However, this involves the challenge of maintaining consistent codes across different neural modules in a biological system, which is implausible. The paper also discusses the challenge of representing multiple visual objects simultaneously, a task that healthy people easily achieve. It suggests the idea of multiple spotlights of attention to allow for independent training on separate objects, drawing on evidence of the brain's ability to focus on multiple entities.

DPANN assumes that there exists a unique central representation for each concept. In this way, different sensory modalities and various modules with different internal representation can "understand" one another through this common representation. DPAAN functions as a "universal translator" between various neural modules in the brain. It establishes a *global stable attractor state* as a common language in which symbols from different modules can be translated and represented.

Instead of relying on identical neural codes for symbols, DPAAN associates symbols with fragments of the global stable attractor state. The synapses can be dynamically partitioned during operation, creating independent auto-associative memory buffers for different modules. This partitioning allows for effective communication and symbol transfer between modules while preventing crosstalk. To maintain consistency and avoid violations of role-filler independence, DPAAN synchronizes the training of associations between neural codes and the universal language. This synchronization occurs when the spotlight modules focus on the same object.

(Kriete et al., 2013) introduces an NN model in which neurons in one part of the prefrontal cortex (area *A*) encode and maintain a pattern of neural activity representing the location or address of information stored in another part of the PFC (area *B*). This mechanism allows representations in area *A* to regulate the use of information in area *B* through the basal ganglia. The paper introduces the concept of *indirection* in neural processing, where neural populations in area *A* serve as pointers to neural populations in area *B*. This indirection mechanism enables the binding of roles and fillers in a flexible and generalizable manner.

Unlike traditional neural representation schemes where each population of neurons encodes specific information content, the indirection model separates the representation of roles and fillers. This separation avoids resource duplication and promotes efficient representation. The indirection model facilitates generalization by allowing different roles to point to the same filler representation. This enables the model to generalize relationships between fillers in different roles.

The paper demonstrates that the proposed architecture can self-organize and learn to implement indirection. Learning occurs through exposure to various role-filler combinations, enabling the system to process a wide range of combinations, even those it has never encountered before.

The architecture of the model also allows for hierarchically nested control, where the output of a group of neurons in the PFC controls the BG gating signal for another group of neurons. This separation of variables and values supports the mechanism of indirection.

## 14. Cognitive Models

(Dehaene, Kerszberg & Changeux, 1998) addresses the modeling of the brain processing in case of tasks assuming conscious effort. The authors suggest the existence of a *global workspace* composed of a distributed set of cortical neurons. These neurons receive inputs from and send outputs to neurons in various cortical areas through long-range excitatory axons. This population of neurons is not restricted to specific brain areas but is distributed among them in variable proportions. The extent to which a particular brain area contributes to the global workspace depends on the fraction of its pyramidal neurons contributing to layers 2 and 3, which are particularly prominent in certain cortical structures. Another computational space consists of a network of functionally specialized processors or modular subsystems. This architecture is presented in Figure 14.1a.

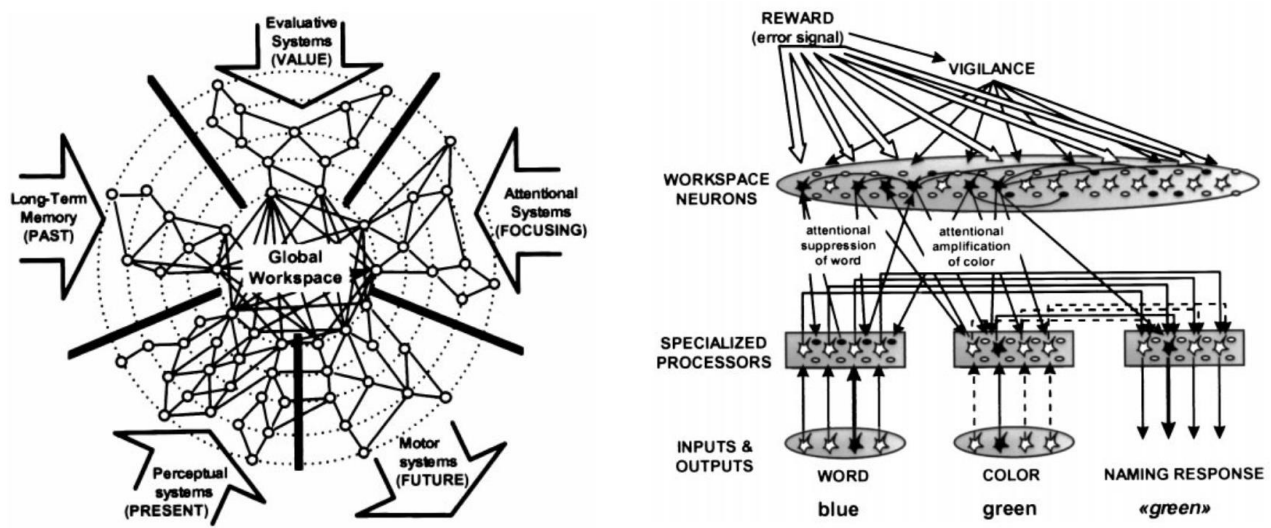

**Figure 14.1.** a) The architecture of the global workspace and several specialized processors;
b) A corresponding simulated network used to solve the Stroop task
(Dehaene, Kerszberg & Changeux, 1998)

The global workspace selectively allows a subset of inputs to access it at any given time. This gating process is mediated by descending projections from workspace neurons to peripheral processor neurons. These projections can amplify or suppress ascending inputs from processing neurons, activating specific processors in the workspace while inhibiting others.

Workspace activity exhibits specific spatio-temporal dynamics. It is characterized by the spontaneous activation of a subset of workspace neurons in a coherent and exclusive manner; only one "workspace representation" can be active at a time. This property distinguishes the global workspace from peripheral processors, where multiple representations can coexist. An active representation in the workspace may remain active autonomously, but can be replaced by another representation if negatively evaluated or if attention shifts. This dynamic property of workspace neurons contributes to generating diversity in thought and cognitive processing, as it constantly projects and tests hypotheses about the external world.

The proposed neuronal architecture (Figure 14.1b) demonstrates the ability to learn the Stroop test without the need for prewired rule-coding units specifically designed for the task. Learning is achieved through realistic neuronal processes.

Peter Gärdenfors (2004) introduces *conceptual spaces* as a framework that blends elements from symbolic and connectionist approaches. It offers an alternative for representing concepts and knowledge. Concepts are represented as regions within multidimensional spaces, i.e., coordinate systems with each dimension corresponding to a specific attribute or quality related to the concept. For example, the "color" concept may be represented in a space with dimensions for hue (e.g., red, green, blue), saturation (intensity of color), and brightness (lightness or darkness).

Concepts are linked to the *quality dimensions* that characterize them. These dimensions encompass a wide range of attributes, from sensory qualities, e.g., taste and smell, to more abstract ones, such as virtue or danger. The conceptual space of taste may have dimensions such as sweetness, bitterness, and saltiness. By positioning concepts within these dimensions, the theory captures how different qualities relate to one another.

A distinctive feature of this theory is its focus on the geometric structure of conceptual spaces. The distances between concepts in these spaces carry semantic meaning and measure *similarity*: close concepts are similar, while distant concepts are dissimilar. Convex regions play a significant role in representing natural categories, with prototypes at the geometric centroid of regions. *Typicality* is measured by the degree of centrality within the region of a concept.

The operations performed in such spaces, such as intersection (common attributes) and blending (combining attributes), mirror how people combine concepts in their thinking. *Intersection* involves finding commonalities between two concepts, for example, the intersection of "bird" and "mammal" may yield "bat". Concept *blending* or combination occurs by intersecting or merging different space regions, for instance, blending the concepts of "stone" and "lion" yields a new concept of "stone lion". This capacity to blend concepts can capture composite or metaphorical meanings.

The meaning of the concept is often determined by the *context* in which it occurs, because some properties cannot be defined independently of others. For example, the "tall" property is connected to the height dimension, but cannot be identified with a particular region in this dimension. A Chihuahua is a dog, but a tall Chihuahua is not a tall dog. Thus, "tall" cannot be identified with a set of tall objects or with a tall region of the height dimension. The solution to the problem is that this property presumes a *contrast class* given by some other property, since things are not tall in themselves but only in relation to a particular class of things.

(Lieto, Chella & Frixione, 2017) considers that conceptual spaces can serve as a *lingua franca* between symbolic and subsymbolic representations. It reminds the challenge of reconciling *compositionality* and *typicality* in conceptual representations. Logic-based representations are compositional but generally incompatible with typicality effects because prototypes cannot always be composed from the individual prototypes of subconcepts. Conceptual spaces, based on geometric representations, offer a more promising approach to dealing with typicality. By representing concepts as convex regions in a suitable conceptual space, typicality can be measured as the distance of an individual point from the center of the region. The intersection of regions represents the conjunction of concepts, allowing for a more intuitive representation of typicality and compositionality. Thus, conceptual spaces can unify and generalize aspects of symbolic and subsymbolic approaches.

## 15. Categorization Models

In this section we will present several categorization models proposed within the field of cognitive psychology (CP). Unlike the classification algorithms in machine learning (ML), their primary goal is not only to model some data, but also match human performance on those data. Still, one can clearly recognize similarities between ML algorithms and the algorithms presented in the following sections.

In general, these algorithms are based on the concepts of examples, prototypes, or rules, which we briefly define as follows.

*Examples* are specific instances or representations of items or events within a category. Examples serve as concrete manifestations of a category and can vary extensively. They are the individual cases or stimuli that a person encounters and associates with a specific category. For example, different breeds of dogs (e.g., Labrador or Poodle) serve as examples within the broader category of "dogs".

*Prototypes* denote the central representation of a category. They encapsulate the most typical or illustrative features of a category and serve as a mental average. Prototypes are formed based on the features or attributes commonly shared among different examples within a category. For example, a typical bird may have features such as wings, beak, feathers, and the ability to fly, serving as a prototype for the category "bird".

*Rules* involve explicit criteria that determine category membership. They are based on specific defining features or characteristics that an object or event must possess to be included within a particular category. The criteria can refer to either inclusion or exclusion in/from a category. For example, a rule for categorizing a car as a "sports car" may involve high horsepower, aerodynamic design, and high-speed capabilities.

Since this review is intended for an AI audience, we will use the ML terminology here, although in cognitive psychology slightly different terms are used, e.g.: categorization (CP) = classification (ML), stimuli (CP) = training instances (ML), transfer stimuli (CP) = testing instances (ML), exemplars (CP) = instances stored in memory.

### 15.1. Adaptive Resonance Theory[7]

*Adaptive resonance theory* (ART) (Grossberg, 1976a; Grossberg, 1976b; Grossberg, 2013) tries to explain how the brain learns, classifies, and predicts in a dynamic environment while maintaining stable memories. It addresses the *stability-plasticity dilemma*, which appears to be solved by the brain's self-organizing nature; it refers to the need to learn rapidly and adaptively without erasing past knowledge. ART attempts to elucidate the neural mechanisms to resolve this dilemma.

ART also focuses on intentional and attentional processes and highlights the importance of top-down expectations in focusing attention on relevant information. It tries to model how resonant brain states emerge when bottom-up and top-down signals align. Such resonant states facilitate rapid learning by directing attention to critical feature patterns necessary for correct classification.

---

[7] This section uses information and ideas from: (da Silva, Elnabarawy & Wunsch, 2019).

Thus, ART allows the learning of infrequent but impactful data. After stabilization, input patterns cause the selection of the classes that offer the best match to top-down expectations, acting as prototypes for the represented class of input patterns. ART can learn both general prototypes and individual exemplars, and their balance is controlled by a user-defined parameter.

Although there are many ART versions (ART1, ART2, Fuzzy ART, Fuzzy min-max, Distributed ART, ARTMAP, Fuzzy ARTMAP, etc.), the basic ART architecture is the one presented in Figure 15.1.

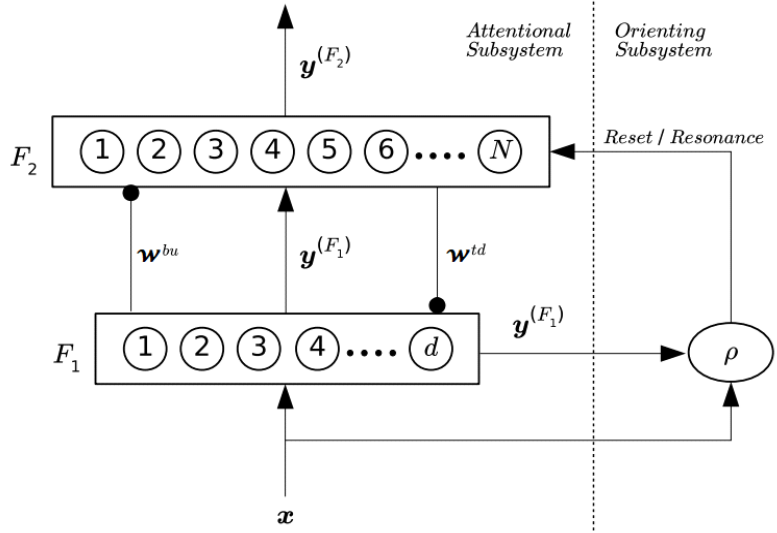

**Figure 15.1.** The architecture of a simple ART model
Adapted after (da Silva, Elnabarawy & Wunsch, 2019)

It has two layers with bidirectional connections and a decision making module.

The input layer is $F_1$, which can also be considered as a short-term memory. It projects the inputs to the $F_2$ layer through the bottom-up weights $\mathbf{w}^{bu}$, which act like a long-term memory. Then, the activations in $F_2$ are propagated back to the $F_1$ layer through the top-down LTM weights $\mathbf{w}^{td}$. Then $F_1$ acts as a comparator between the actual input and the expectation. In some representations, two separate layers appear instead of a single $F_1$, i.e., one for the input and the other for the expectation.

The $F_2$ layer encodes the prototypes created by the input data. When an instance is presented at $F_1$ and propagated to $F_2$, a winner-takes-all competition occurs between the $F_2$ units, and the neuron with the highest activation is the only one selected:

$$j^* = \arg\max_j a_j \tag{15.1}$$

An $F_2$ neuron represents a prototype (category, hypothesis). Its adequacy for the current input is further tested by the extent to which the expectation at $F_1$ matches the input instance. The amount of similarity between the two representation that allows one to assume that there is a match, is controlled by a *vigilance parameter $\rho$*. When a match occurs, this is equivalent to a resonant state in the network between the two layers and the model learns, i.e., the prototype $j^*$ is modified to account for the new input. If the similarity if below the threshold imposed by $\rho$, $j^*$ is inhibited, and the prototype with the next highest activation is selected. If eventually no existing prototype is found to match the input data, a new prototype is created that reflects the current input.

The general ART algorithm has the following steps – adapted from (da Silva, Elnabarawy & Wunsch, 2019):

- Present an input instance;
- Compute the activation values for each prototype;
- Select the prototype with the highest activation value by WTA;
- Evaluate the match between the input and the selected prototype;
- If the match is sufficient, update the prototype based on the input;
- If the match is insufficient, deactivate the prototype;
- Repeat the WTA process with the other prototypes until a matching one is found;
- If no suitable prototype is found, create a new prototype based on the input;
- Generate an output based on the selected or created prototype;
- Process the next input.

### 15.1.1. ART 1

The *ART 1* model (Carpenter & Grossberg, 1987) considers binary inputs and uses the Hamming distance as a similarity measure. The activations of the neurons in $F_2$ for an input $\mathbf{x}$ are:

$$a_j = \left\| \mathbf{x} \wedge \mathbf{w}_j^{bu} \right\|_1 = \sum_{i=1}^{d} x_i w_{ji}^{bu} \qquad (15.2)$$

where $\wedge$ represents the element-wise logical *and* function.

After a neuron $j^*$ is selected after the WTA competition, the $F_2$ activity is:

$$\mathbf{y}_j^{(F_2)} = \begin{cases} 1 & \text{if } j = j^* \\ 0 & \text{otherwise} \end{cases} \qquad (15.3)$$

Then, the $F_1$ activity (regarded as an expectation) is:

$$\mathbf{y}^{(F_1)} = \begin{cases} \mathbf{x} & \text{if } F_2 \text{ is inactive} \\ \mathbf{x} \wedge \mathbf{w}_{j^*}^{td} & \text{otherwise} \end{cases} \qquad (15.4)$$

The neuron with the highest activation is then tested for resonance using a *match function*:

$$m_{j^*} = \frac{\left\| \mathbf{y}^{(F_1)} \right\|_1}{\left\| \mathbf{x} \right\|_1} = \frac{\left\| \mathbf{x} \wedge \mathbf{w}_{j^*}^{td} \right\|_1}{\left\| \mathbf{x} \right\|_1} \qquad (15.5)$$

If $m_{j^*} \geq \rho$ ($\rho \in [0, 1]$), then the corresponding prototype is allowed to learn:

$$\mathbf{w}_{j^*}^{bu}(new) = \frac{l}{l - 1 + \left\| \mathbf{w}_{j^*}^{td}(new) \right\|_1} \mathbf{w}_{j^*}^{td}(new) \qquad (15.6)$$

where $l > 1$ is a user-defined parameter.

If a new prototype is created for input **x**, it is initialized with:

$$\mathbf{w}^{td} = \mathbf{1} \tag{15.7}$$

$$\mathbf{w}^{bu} = \frac{l}{l-1+d}\mathbf{w}^{td} \tag{15.8}$$

### 15.1.2. Fuzzy ART

*Fuzzy ART* (Carpenter, Grossberg & Rosen, 1991) is one of the most popular ART models. It can handle real-valued data and uses fuzzy logic operations. Usually, the inputs are transformed using *complement coding*, where **x** becomes [**x**, **1** – **x**]. In this way, both the presence and the absence of a data attribute is explicitly handled.

The activations of the $F_2$ units are defined as:

$$a_j = \frac{\|\mathbf{x} \wedge \mathbf{w}_j\|_1}{\|\mathbf{w}_j\|_1 + \alpha} \tag{15.9}$$

where **w** is a single weight matrix, $\wedge$ represents the fuzzy logic operation *and* (element-wise minimum), and $\alpha > 0$ is the *choice parameter* that performs a role similar to regularization.

The $F_1$ activations are:

$$\mathbf{y}^{(F_1)} = \begin{cases} \mathbf{x} & \text{if } F_2 \text{ is inactive} \\ \mathbf{x} \wedge \mathbf{w}_{j^*} & \text{otherwise} \end{cases} \tag{15.10}$$

and when the winner node $j^*$ is selected, the $F_2$ activity is given by equation (15.3).

The match function is now:

$$m_{j^*} = \frac{\|\mathbf{y}^{(F_1)}\|_1}{\|\mathbf{x}\|_1} = \frac{\|\mathbf{x} \wedge \mathbf{w}_{j^*}\|_1}{\|\mathbf{x}\|_1} \tag{15.11}$$

When the vigilance criterion is met by prototype $j^*$, it is modified as:

$$\mathbf{w}_{j^*}(new) = (1-\beta) \cdot \mathbf{w}_{j^*}(old) + \beta \cdot \left(\mathbf{x} \wedge \mathbf{w}_{j^*}(old)\right) \tag{15.12}$$

where $\beta \in (0, 1]$ is the *learning parameter*. When $\beta = 1$, the ART model is said to be in the *fast learning* mode. When an new prototype is created to learn an input **x**, it is initialized with **w** = **1**.

In general, one neuron cannot account for a concept, as demonstrated by the evidence for the group coding of many concepts in neural populations. The individual neurons that represent prototypes or classes in ART may be considered to stand for cell assemblies and approximate their behavior, aligning with the goal of neural plausibility.

## 15.2. The Generalized Context Model[8]

The *generalized context model* (GCM) (Nosofsky, 1984; 1986; 2011) is based on the idea that classification relies on comparing new instances with previously encountered instances stored in memory (exemplars). It is an extension of the previous *context model* (Medin & Schaffer, 1978). The main idea resembles the *k-nearest neighbors* algorithm in ML.

However, the GCM uses *multidimensional scaling* (MDS) to create an estimation of the psychological space of the problem where the instances can be represented as points. MDS relies on human subjects to estimate the similarity between pairs of instances and creates a representation (usually 2D or 3D) where the distances between points reflect their original similarities, i.e., similar items are closer and dissimilar ones are farther apart.

GCM also emphasizes that similarity is context-dependent, affected by selective attention weights that modify the space where exemplars are embedded. These weights stretch along relevant dimensions and shrink along irrelevant ones. For example, in the space of flowers defined by a fixed set of attributes, those attributes may have a set of weights when one has the goal of finding beautiful flowers and a different set of weights when the goal is to find medicinal flowers. Also, exemplars usually have varying strengths influenced by factors such as presentation frequency, recency, or feedback during learning. When classifying a test item, highly similar exemplars with high memory strengths are likely to influence the decision more. However, since retrieval is probabilistic, all exemplars can impact classification decisions to some extent.

According to the model, in the test phase an instance $i$ is classified in class $c$ based on the following equation, which implies that the probability of a choice results from the normalization of the values corresponding to the alternatives, also known as *Luce's choice rule* (Luce, 1963):

$$P(c\,|\,i) = \frac{\left(\sum_{j=1}^{m} v_{jc} s_{ij}\right)^{\gamma}}{\sum_{k=1}^{C}\left(\sum_{l=1}^{m} v_{lk} s_{il}\right)^{\gamma}} \tag{15.13}$$

where $m$ is the number of training instances, $C$ is the number of classes, $v_{jc}$ is the memory strength of exemplar $j$ with respect to class $c$, and $s_{ij}$ is the similarity between instance $i$ and exemplar $j$. $\gamma$ is a user-defined positive response-scaling parameter that influences the level of determinism in classification responses. When $\gamma = 1$, responses are probabilistic, matching the relative summed similarities of each class. When $\gamma > 1$, responses become more deterministic, favoring the class with the highest summed similarity. The memory-strength values **v** are often predetermined by experimental design, commonly set based on the relative frequency of each exemplar during training. For example, in typical experiments where instances are equally presented and assigned to single classes, their memory strengths are set to 1, while unassigned instances have a strength of 0 for those classes.

The similarity $s_{ij}$ is computed as:

$$s_{ij} = e^{-\beta \cdot d_{ij}} \tag{15.14}$$

---

[8] This section uses information and ideas from: (Nosofsky, 2011).

$$d_{ij} = \left( \sum_{a=1}^{D} w_a \left| x_{ia} - x_{ja} \right|^r \right)^{1/r} \tag{15.15}$$

In these equations, $d_{ij}$ is the Minkowski distance between the points $\mathbf{x}_i$ and $\mathbf{x}_j$ in the $D$-dimensional psychological space. $\mathbf{w}$ are the attention weights for each dimension; they are positive and their sum is 1. $\beta$ is a user-defined sensitivity parameter that shows how steeply similarity declines with distance. A larger $\beta$ results in a steep similarity decline, resembling a nearest neighbor classifier, while a smaller $\beta$ leads to a moderate decline, allowing multiple stored instances to influence classification.

Parameter $r$ defines the form of the distance metric. For *separable dimension* instances, $r = 1$ (the Manhattan distance); for *integral dimension* instances, $r = 2$ (the Euclidian distance) (Shepard, 1964). Separable dimensions are those that can be attended to and processed independently of one another. For example, in a dataset defined by color and shape, each object is described by two separate categorical dimensions which are unrelated. Integral dimensions are those that cannot be processed independently – the instances are perceived as varying along a single dimension. For example, when 3D objects are rotated along different axes, the viewpoint can be considered a single integral dimension (Soto & Wasserman, 2010).

**15.3. ALCOVE**

The *attention learning covering map* (ALCOVE) (Kruschke, 1992) is an algorithm that combines the exemplar-based idea of GCM with error-driven learning, characteristic, e.g., to neural networks. ALCOVE extends both by adding a learning mechanism, allowing continuous dimensions, and incorporating dimensional attention learning. Its design, inspired by psychological theories, differs from standard NNs using backpropagation, because it aims to model human learning processes rather than to map inputs to desired outputs after extensive training.

ALCOVE operates as a feedforward connectionist network. It also assumes that instances can be represented in a psychological space using MDS. Its general architecture is presented in Figure 15.2.

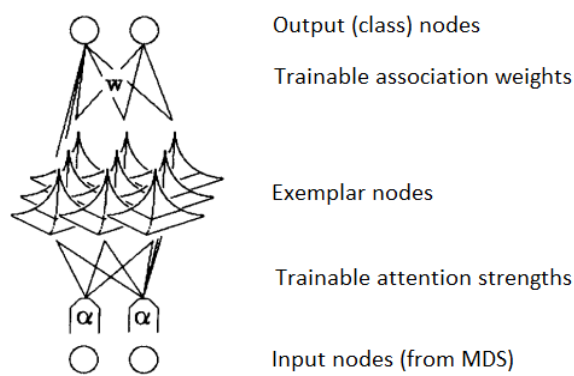

**Figure 15.2.** The architecture of ALCOVE
Adapted from (Kruschke, 1992)

Each input node of the network represents a specific dimension, activated based on the value of the instance on that dimension, exactly like the inputs of an MLP.

The influence of the input nodes on the subsequent hidden layer also depends on the *attention strengths* of each dimension, which signify the relevance of each dimension for the classification task. Initially, these attention strengths are equal on all dimensions but adapt during training to increase on relevant dimensions and decrease on less relevant ones. This process of attention learning gives the first part of the ALCOVE name.

Hidden nodes correspond to positions in the instance space, similar to radial basis function networks. In its basic form, each exemplar corresponds to the position of a hidden node. A more complex version involves random distribution of hidden nodes, creating a covering map of the input space, which gives the last part of the ALCOVE name.

The activation of hidden nodes is determined by the similarity between the input (test) instance and the exemplar corresponding to that hidden node. This similarity computation is similar to the one used in GCM:

$$a_j^{hid} = e^{-\beta \cdot d_{ij}} \qquad (15.16)$$

$$d_{ij} = \sum_{i=1}^{D} \alpha_i \left| h_{ji} - a_i^{in} \right| \qquad (15.17)$$

In equation (15.17), only separable dimensions are assumed here, hence the use of the Manhattan distance. This equation also shows the role of dimensional attention strengths $\alpha_i$, acting as multipliers on dimensions when computing the distance between an input instance and a hidden node.

Each hidden node is linked to output nodes representing class membership. The connection weight between a hidden node and a class node is called the *association weight*. Unlike the GCM, in ALCOVE the association weights **w** are adjusted iteratively through an error-driven learning rule and can assume real values, including negative ones:

$$a_c^{out} = \sum_j w_{cj} a_j^{hid} \qquad (15.18)$$

To assess the model performance, the class activations are translated into response probabilities, using the same choice rule used in the GCM (Luce, 1963):

$$P(c) = \frac{\exp(\phi \cdot a_c^{out})}{\sum_k \exp(\phi \cdot a_k^{out})} \qquad (15.19)$$

In equation (15.19), $P(c)$ should be understood in the same way as in equation (15.13), i.e., $P(c|i)$, the probability of class $c$ for instance $i$.

Thus, output nodes are activated by summing across hidden nodes, weighted by association weights, ultimately yielding response probabilities.

To find the best dimensional attention strengths **α** and association weights **w**, ALCOVE uses gradient descent on the sum squared error, like standard backpropagation, with updates after the presentation of each training instance:

$$E = \frac{1}{2} \sum_k \left(t_k - a_k^{out}\right)^2 \qquad (15.20)$$

$$t_k = \begin{cases} \max\left(1, a_k^{out}\right) & \text{if } i \in k \\ \min\left(-1, a_k^{out}\right) & \text{if } i \notin k \end{cases} \qquad (15.21)$$

By $i \in k$, we mean that the desired (target) class of training instance $i$ is $k$.

Unlike backpropagation for MLPs, which uses the desired output values as targets, ALCOVE uses so-called *humble teacher* values $t_k$. For example, if an instance belongs to a class, the corresponding output node should have an activation of at least 1. If the activation is greater than 1, the difference is not considered an error. The metaphor is related to a teacher that does not mind being outperformed by his/her students.

The adjustments to **α** and **w** are based on gradient descent:

$$\Delta \alpha_i = -\lambda_\alpha \cdot \sum_j \left( \sum_k \left(t_k - a_k^{out}\right) \cdot w_{kj} \right) \cdot a_j^{hid} \cdot c \cdot \left|h_{ji} - a_i^{in}\right| \qquad (15.22)$$

$$\Delta w_{kj} = \lambda_w \cdot \left(t_k - a_k^{out}\right) \cdot a_j^{hid} \qquad (15.23)$$

$\lambda_\alpha$ and $\lambda_w$ are user-defined learning rates. After an update, if $\alpha_i < 0$, then $\alpha_i = 0$, to ensure a valid psychological interpretation.

## 15.4. RULEX

Exemplar-based models allow for flexible data classification, especially in scenarios where rules may be challenging to define. They can also account for graded membership in classes, where some instances are more typical than others. This can be the case with the classification of natural objects defined by many attributes.

At the other end of the abstraction spectrum, rule-based models, such as the *rule plus exception* (RULEX) (Nosofsky, Palmeri & McKinley, 1994; Nosofsky & Palmeri, 1998), try to find concise class representations. People often classify objects and situations based on explicit rules or criteria based on a small number of important features. Another advantage of rules is that they are usually verbalizable (describable in words) and easy to understand. Rule-based models also require less memory and computational resources.

An experiment (Rips, 1989) showed conclusive evidence for rule-based classification in a situation where it is clearly not equivalent to similarity-based classification. Human subjects were asked to classify an object as a pizza or a coin based only on its diameter. They were presented with test instances beyond the normal boundaries that defined the two classes – they were asked to classify an object with a diameter of 7 cm, which was larger than a normal coin and smaller than a normal pizza. The subjects classified such objects as pizzas, because coins are restricted in size but pizzas are not. Even if they assessed the 7 cm object to be more similar to a coin, they classified it as a pizza because they implicitly applied a rule such as "if an object is more than 3 cm in diameter, it cannot be a coin" (Smith & Sloman, 1994).

The RULEX model, first created to handle binary data (Nosofsky, Palmeri & McKinley, 1994), was later extended to handle continuous (real-valued) data as well (Nosofsky & Palmeri, 1998). Based on the decision boundary theory, it establishes partitions within the psychological problem space, where boundaries separate distinct classes. In the simplest case, the algorithm identifies single-dimension rules. For binary data, a rule can represent a set of values that are similar for most instances; for continuous data, the rule is given by a threshold value such that instances can have higher or lower values on that dimension. RULEX also handles intervals, i.e., two threshold values on a single dimension. More complex logical rules can result from combinations of such boundaries using conjunctions and disjunctions along multiple dimensions.

Once a rule is established, exceptions are then stored. For binary data, the exceptions represent the exact matches to an exception instance; for the continuous case, they are based on similarity comparison. In case of rules involving multiple dimensions, exceptions can also be defined as subsets of those dimensions. Similarity is computed in a similar way as by GCM and ALCOVE.

Given a rule $R$ and a set of exceptions to that rule $E(R)$, the probability to use the exception-based classification process for an instance $i$ is:

$$P_{use-ex}(R) = \frac{\sum_{j \in E(R)} s_{ij}}{\sum_{j \in E(R)} s_{ij} + v} \quad (15.24)$$

where $v$ is a user-defined parameter called *exception use criterion*.

The probability of classifying instance $i$ in class $c$ with rule $R$ and its exceptions is:

$$P_R(c \mid i) = P_{use-ex}(R) \cdot P_{exR}(c \mid i) + \left(1 - P_{use-ex}(R)\right) \cdot P_{ruleR}(c \mid i) \quad (15.25)$$

Because several rules $r$ may be available for a problem, the overall probability for the classification in class $c$ is:

$$P(c \mid i) = \sum_r P(r) \cdot P_r(c \mid i) \quad (15.26)$$

## 15.5. ATRIUM

The *attention to rules and items in a unified model* (ATRIUM) (Erickson & Kruschke, 1998) combines the exemplars and rules approaches, implemented in separate modules, together with a competitive gating mechanism that decides, for a given test instance, which of the two separate results should be returned as the output of the model. Its general architecture is presented in Figure 15.3.

In the rule module, the instances activate rule nodes based on their placement relative to 1D rule boundaries. The rule nodes implement a sigmoid function where the midpoint of the sigmoid represents the rule threshold. The rule nodes are linked to class nodes via connections with learned weights.

The exemplar module receives the same input as the rule module; it interprets instances as points in psychological space and activates nearby exemplar nodes more strongly and distant nodes more weakly. The exemplar nodes also connect to class nodes via connections with learned weights. This module is an implementation of ALCOVE, presented in Section 15.3.

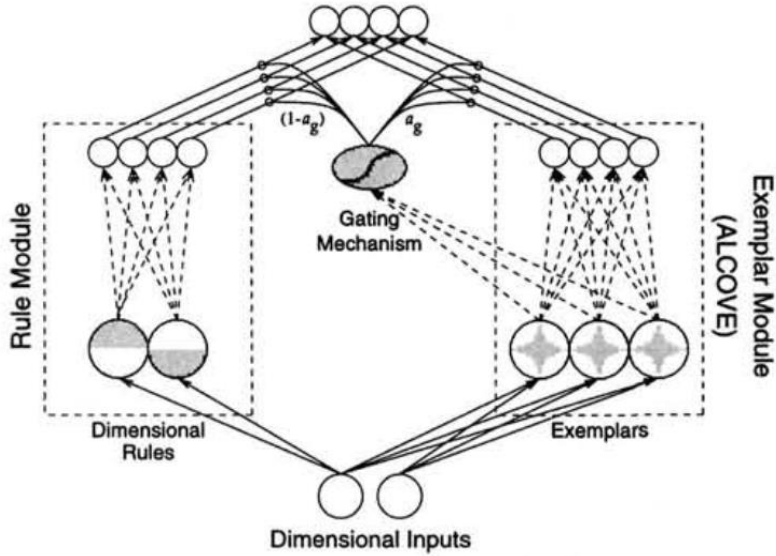

**Figure 15.3.** The architecture of ATRIUM
The dotted lines represent connections with learned weights
(Erickson & Kruschke, 1998)

Each instance is processed in parallel by the two modules. ATRIUM thus implements a so-called *mixture-of-experts* approach, where each expert, i.e., module, learns its own mapping from inputs to outputs using its own form of representation.

The gating mechanism learns to activate a certain module in response to a particular input. The probability of choosing a class $c$ as the output is given by:

$$P(c) = a_g \cdot \frac{\exp(\phi \cdot a_{e_c})}{\sum_k \exp(\phi \cdot a_{e_k})} + (1 - a_g) \cdot \frac{\exp(\phi \cdot a_{r_c})}{\sum_k \exp(\phi \cdot a_{r_k})} \qquad (15.27)$$

where $a_{e_c}$ and $a_{r_c}$ are the activations of the exemplar and rule nodes for class $c$, $\phi$ is a user-defined scaling parameter, and $a_g$ is the activation of the gating node signifying the weight of the exemplar module:

$$a_g = \frac{1}{\exp\left(-\gamma_g \sum_{e_j} w_{g,e_j} a_{e_j} + \beta_g\right)} \qquad (15.28)$$

where $w_{g,e_j}$ is the weight from exemplar node $j$ to the gating node, and $\beta_g$ and $\gamma_g$ are other user-defined parameters.

The mean accuracies of the exemplar and rule modules, respectively (where $t_k$ represent the humble teacher values defined in equation (15.21), and $c_e$ and $c_r$ are other parameters) are defined as:

$$EA = \exp\left(-\frac{1}{2} c_e \cdot \|\mathbf{t}_e - \mathbf{a}_e\|^2\right) \qquad (15.29)$$

$$RA = \exp\left(-\frac{1}{2} c_r \cdot \|\mathbf{t}_r - \mathbf{a}_r\|^2\right) \tag{15.30}$$

The mean accuracy of the whole system and then the total error are defined as:

$$MA = a_g \cdot EA + (1 - a_g) \cdot RA \tag{15.31}$$

$$E = -\log(EA) \tag{15.32}$$

Based on *E* the learning equations are deduced by gradient descent for the connection weights and the attention weights. They are complex formulas that we will omit here, but can be found in the original paper (Erickson & Kruschke, 1998).

## 15.6. COVIS[9]

Unlike the previous models, the *competition between verbal and implicit systems* (COVIS) (Ashby et al., 1998) is more complex, therefore in this section we will provide only a general description, without any equations.

It combines an explicit (verbalizable) module based on a declarative memory with rules and hypothesis testing and an implicit (non-verbalizable) module called the "procedural system" that learns subsymbolic classification decisions with a neural network representation and a form of reinforcement learning. The first system can learn quickly a rather small set of rules, in situations where a simple separation of classes is possible. It tries to model the functionality of the prefrontal cortex. The second system can learn more general patterns, but slowly and incrementally and depends on immediate feedback. It tries to model the functionality of the striatum, the major input region of basal ganglia. The procedural learning system reflects the incremental stimulus-response associations created by dopamine-mediated Hebbian learning.

The declarative memory relies on conscious reasoning, while the non-declarative memory system does not require any conscious processes. These two systems compete to propose a response and therefore COVIS also contains a decision module that selects the winner for each testing instance.

Since COVIS is based on a neurobiologically-constrained architecture, it succeeds in predicting many behavioral and neuroscience data in cognitive psychology experiments.

### *16.6.1. The Explicit System*

This system tries to identify fairly simple rules for classification. In the simplest case, rules are one-dimensional, but more complex rules can be envisioned on more dimensions using logical conjunctions or disjunctions.

This module is implemented using a hybrid neural network with both symbolic and connectionist components. The selection of rules is explicit, but the decision criteria about the salience of rules are learned via gradient descent.

---

[9] This section uses information and ideas from: (Ashby, Paul & Maddox, 2011), (He, 2020), and (Sun, 2023).

The set of all possible rules corresponding to the problem space is considered to be available. On each trial, the model selects one of these rules for application. The one-dimensional rules are defined by a specific threshold (the decision criterion).

If the response on a trial is correct using a rule, then that rule is deterministically selected again in the next trial. If the response is incorrect, then each rule in the rule set can be selected with a probability that depends on its weight, which results from its reward history, the tendency to perseverate, and the tendency to select unusual rules. These last criteria clearly address the modeling of the performance of human subjects on classification tasks.

In some advanced versions of COVIS, the working memory is implemented by separate networks. One is responsible for maintaining candidate rules, testing them, and switching between rules. Another network is responsible for generating or selecting new candidate rules.

COVIS predicts that the most time-consuming processing happens when the current rule is found to be incorrect. When a correct rule is finally selected, there is a sudden shift from suboptimal to optimal performance, and this mimics human results.

*16.6.2. The Procedural System*

Initially, this system was implemented as a typical perceptron; however, there is psychological evidence that people do not classify instances based on decision bounds (Ashby & Waldron, 1999), therefore an alternative architecture was proposed, called the "striatal pattern classifier".

This is a feedforward network with an input layer of 10000 units (corresponding to the sensory cortex), two units in the hidden layer (corresponding to the striatum), and two units in the output layer (corresponding to the premotor cortex). The units are modified Izhikevich neurons, presented in Section 3.2. Only the synapses between the input layer and hidden layer can be modified. A more biologically detailed variant of COVIS includes lateral inhibition between striatal units and between cortical units.

In the input layer, each unit is tuned to a different stimulus and its response decreases as a Gaussian function of the distance in stimulus space between the stimulus preferred by that unit and the presented stimulus.

The activation of the units in the hidden layer is determined by the weighted sum of activations of the input units. The synaptic strengths are adjusted via reinforcement learning (RL), however, we must note that RL is used here in a biological sense, not in a machine learning sense.

There are three factors involved in strengthening input-to-hidden synapses: strong presynaptic activation, strong postsynaptic activation, and high dopamine levels. A synapse is strengthened if the input cell responds strongly to the presented stimulus and there is a reward for responding correctly. The strength of the synapse decreases if the response is incorrect, or if the synapse has an input cell that does not fire strongly to the stimulus.

Learning is also determined by the dopamine-mediated reward signals. Positive feedback that follows successful responses increases dopamine levels, which strengthen recently active synapses. Negative feedback causes dopamine levels to fall, which weaken recently active synapses. Dopamine levels serve as a teaching signal that increase the probability of successful responses and decrease the probability of unsuccessful ones. COVIS includes a model of dopamine release based on the difference between the obtained reward and the predicted reward.

*16.6.3. Deciding the Final Response*

Out of the two competing systems, the one with better overall performance decides the final response. There are two factors involved in this decision: the confidence each system has in the accuracy of its response and how much each system can be trusted. In the case of the explicit system, confidence is the absolute value of the discriminant function, which is 0 when the input instance is on the decision bound (low confidence), and large when the instance is far from the bound (high confidence). In the procedural-learning system, confidence is defined as the absolute value of the difference between the activation values in the two hidden units. The amount of trust that is placed in each system is a function of an initial bias toward the explicit system and the previous success history of each system.

**15.7. SUSTAIN[10]**

The *supervised and unsupervised stratified adaptive incremental network* (SUSTAIN) model (Love, Medin & Gureckis, 2004) relies on prototypes for classification. Its internal representation consists of prototypes associated with classes, and new (training) instances are assigned to existing prototypes or form new prototypes via unsupervised learning. Classification decisions are based on the information contained in these prototypes. The model also uses attentional tunings that are learned for each input dimension (problem attribute). Its architecture is presented in Figure 15.4.

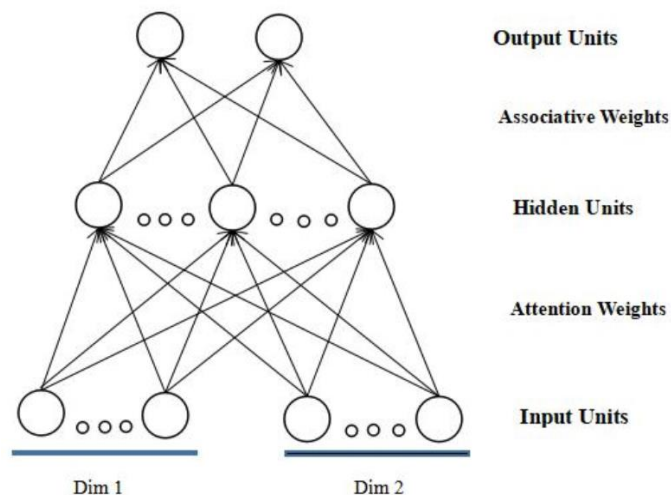

**Figure 15.4.** The architecture of SUSTAIN
(He, 2020)

SUSTAIN is built on five main principles:

1. *Initial preference for simple solutions*. It starts with single prototypes and progresses to more complex representations as needed, guided by selective attention directed toward the data dimensions that appear more promising for prediction on a prototype level;

---
[10] This section uses information and ideas from: (Love, Medin & Gureckis, 2004).

2. *Clustering similar inputs together*. It groups inputs through an unsupervised process that relies on similarity. As prototypes form, attention shifts to the dimensions that provide consistent matches;
3. *The combination of unsupervised and supervised learning*. It relies on both types of learning: it expands prototypes based on similarity when the classification is correct, but it creates new prototypes when the similarity-based assignment of an instance to an existing prototype fails to provide a correct classification;
4. *Feedback influences the learned class structure*. Different feedback patterns and the order of training instance presentation may result in different representations;
5. *Prototype competition*. Prototypes compete to explain the instances, and the response of the winning prototype is influenced by the presence of other similar prototypes.

The mathematical formulation of the SUSTAIN model encapsulates its mechanisms for clustering and learning by means of receptive fields and activation processes. Each prototype has a receptive field for every input dimension, centered at its position along that dimension. The position reflects the prototype expectations for its members. Receptive field tuning, distinct from position, determines the attention devoted to an input dimension. The dimensions that offer consistent information receive more attention.

The receptive fields are assumed to have an exponential shape (Figure 15.5), so that their response decreases exponentially with the distance from the center. This choice of negative exponential function, on the one hand, is motivated by the exponential negative expression of similarity as a function of distance (Shepard, 1964), as presented in Section 15.2, and on the other hand, ensures a constant area of 1 underneath a receptive field, which facilitates the mathematical formulation of the algorithm:

$$\alpha(\mu) = \lambda \cdot e^{-\lambda \cdot \mu} \tag{15.33}$$

$$\mu_{ij} = \frac{1}{2} \sum_{k=1}^{v_i} \left| I^{pos_{ik}} - H_j^{pos_{ik}} \right| \tag{15.34}$$

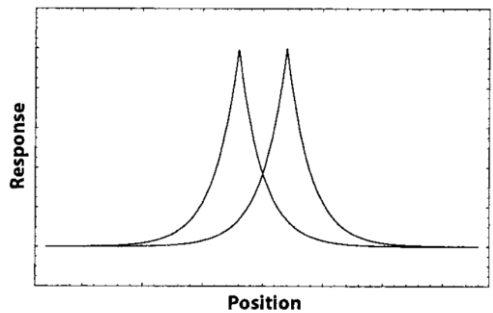

**Figure 15.5.** An example of receptive fields

Receptive fields can differ in their positions while having similar tunings, can exhibit varying tunings with the same positions, or any other combination. Highly tuned receptive fields, depicted by taller profiles, generate stronger but more localized responses, whereas shorter receptive fields produce lower responses that are more widely distributed.
(Love, Medin & Gureckis, 2004)

In these equations, $v_i$ is the number of nominal (symbolic) values on dimension $i$, $\lambda$ is the tuning of the receptive field, $\mu_{ij}$ is the distance between $I^{pos_{ik}}$ (the input on dimension $i$ for value $k$) and the field center $H_j^{pos_{ik}}$ of prototype $j$. $\alpha(\mu)$ is the response of the receptive field.

For a symbolic dimension (with discrete attribute values), the position of a prototype is equivalent to a probability distribution – it represents the probability of an instance that belongs to that prototype to have a specific value.

The activation of a prototype $j$ is:

$$H_j^{act} = \frac{\sum_{i=1}^{D}(\lambda_i)^r \cdot e^{-\lambda_i \cdot \mu_{ij}}}{\sum_{i=1}^{D}(\lambda_i)^r} \quad (15.35)$$

where $D$ is the number of dimensions, $\lambda_i$ is the tuning of the receptive field for dimension $i$, and $r$ is a user-defined "attentional focus" parameter.

Prototypes compete and inhibit one another. Following this lateral inhibition, the output of the winning prototype $j^*$ is influenced by the activations of the competing prototypes:

$$H_{j^*}^{out} = \frac{(H_{j^*}^{act})^\beta}{\sum_{i=1}^{n}(H_i^{act})^\beta} \cdot H_{j^*}^{act} \quad (15.36)$$

where $\beta$ is another user-defined "cluster (or prototype) competition" parameter.

The output of the other prototypes becomes 0: $H_{j \neq j^*}^{out} = 0$.

Then, prototype activation combines the responses from the $np$ prototypes for the class unit $k$:

$$C_{zk}^{out} = \sum_{j=1}^{np} w_{j,zk} \cdot H_j^{out} \quad (15.37)$$

In this equation, $w_{j,zk}$ is the weight from prototype $j$ to class unit $C_{zk}$. Given the previous competition, only the winning prototype $j^*$ actually contributes to the activation of the class units. These activations are further converted into probabilities following Luce's choice rule:

$$P(c) = \frac{e^{\delta \cdot C_{zc}^{out}}}{\sum_j e^{\delta \cdot C_{zj}^{out}}} \quad (15.38)$$

Here, $\delta$ is user-defined "decision consistency" parameter.

After responding, feedback guides learning. The target values for class units are adjusted based on the predicted and expected responses using the same "humble teacher" idea presented in equation (15.21). In case of SUSTAIN, only the limits are different (0 instead of –1):

$$t_{zk} = \begin{cases} \max\left(C_{zk}^{out}, 1\right) & \text{if } I^{pos_{zk}} = 1 \\ \min\left(C_{zk}^{out}, 0\right) & \text{if } I^{pos_{zk}} = 0 \end{cases} \quad (15.39)$$

For the dimension $z$ corresponding to the predicted class, if $t_{zk} \neq 1$ for the $C_{zk}$ with the largest output of all $k$'s, a new prototype is created, centered on the misclassified input instance. This also happens in an unsupervised setting, if the winning prototype activation is below a threshold $H_{j*}^{act} < \theta$.

Otherwise, the position of the winning existing prototype is adjusted towards the input instance, with an idea similar to ART:

$$\Delta H_{j*}^{pos_{ik}} = \eta \cdot \left(I^{pos_{ik}} - H_{j*}^{pos_{ik}}\right) \quad (15.40)$$

The positions and receptive field tunings of prototypes are adjusted as well, together with the weights from the winning prototype to output units are adjusted:

$$\Delta \lambda_i = \eta \cdot e^{-\lambda_i \cdot \mu_{ij*}} \cdot \left(1 - \lambda_i \cdot \mu_{ij*}\right) \quad (15.41)$$

$$\Delta w_{j*,zk} = \eta \cdot \left(t_{zk} - C_{zk}^{out}\right) \cdot H_{j*}^{out} \quad (15.42)$$

where $\eta$ is the learning rate. Equation (15.41) results from the attempt to increase the $\alpha$ function from (15.33) differentially, and (15.42) is actually the delta rule.

**15.8. DIVA**

The *divergent autoencoder* (DIVA) (Kurtz, 2007) is a fully connected, feedforward connectionist model employing backpropagation. Its structure involves an input layer for features, a shared hidden layer, and multiple output nodes representing different classes. It allows encoding and decoding inputs for learning and classification decisions based on the quality of reconstruction. DIVA dedicates a reconstructive learning channel to each class. It integrates these channels through a shared hidden layer, allowing recoding across all classes while reconstructing inputs based on specific class channels. Learning is error-driven, but focuses on reconstructive success rather than classification success. It builds statistical models for each class by recoding and decoding inputs. The representation of classes is influenced by alternative classes in the learning task. Thus, unlike simple autoencoders, DIVA includes the idea of contrast between classes. Its general architecture is presented in Figure 15.6.

Luce's choice rule is used also here to generate output probabilities, but it uses the inverse of the sum of squared errors (SSE) on each channel instead of the activation levels of the output nodes:

$$P(c) = \frac{1/SSE(c)}{\sum_k \left(1/SEE(k)\right)} \quad (15.43)$$

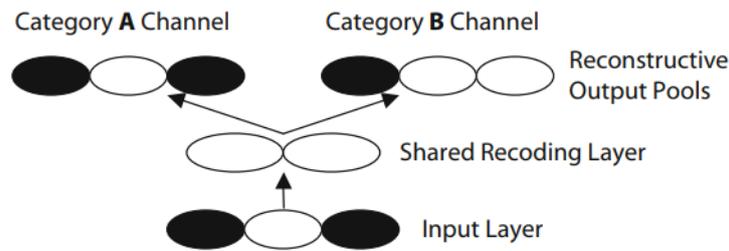

**Figure 15.6.** The architecture of DIVA
(Kurtz, 2007)

Although learning is similar to standard backpropagation, DIVA significantly differs in performance from traditional multilayer perceptrons. It does not convert inputs into a linearly separable space for output class nodes, but learns optimized recodings for accurate feature reconstruction per class. In standard autoencoders, this generates maximal input separation. However, in divergent autoencoding, multiple discrimination spaces overlay due to shared input-to-hidden weights for class-specific transformations. This ease of generating multiple models facilitates multiclass classification.

In a two-choice (*A/B*) classification task, such as the one presented in Figure 15.6, DIVA uses separate channels to reconstruct the inputs labeled *A* and *B*. During learning, the corrective class feedback determines which channel to train for a specific trial. Error-driven updates adjust shared input-to-hidden weights collectively, but update hidden-to-output weights independently for each class. DIVA continues error-driven updates even after achieving the correct class response (unless the reconstruction is perfect) and avoids changing weights along an incorrect class channel.

The learning approach, unlike MLPs, simulates real-world trials as single training trials, not multiple incremental ones, e.g., using a high learning rate of 1. Thus, DIVA performs well on classic categorization benchmark problems.

The conceptual framework of DIVA presents a novel perspective on category learning. Unlike models relying on specific exemplars or similarity matching, DIVA does not use nodes for exemplars or categorization/classification based on the similarity between inputs and reference points. The reference point framework in cognitive psychology posits that individuals classify stimuli based on their proximity to specific reference points which serve as anchors for classification and are often dictated by prototypical examples within a category.

Instead of dimension-based attention adjustments, DIVA transforms inputs based on tasks, creating a distributed representational space. It replicates characteristics linked to rules, prototypes, and exemplars without explicitly employing any of these methods.

# 16. Cognitive Architectures

*Cognitive architectures* serve as computational frameworks designed to replicate the functional organization and operations of the human mind. The fundamental premise underlying these architectures is the belief that human cognition can be divided into a collection of core cognitive processes and structures. The main goal is to present a computational account of these processes and structures. A common approach involves the integration of modules representing key psychological concepts, in order to approximate their functioning using computational methods.

Essentially, cognitive architectures aim to encapsulate cognitive mechanisms, which are information processing structures persisting across various temporal frames and tasks. These architectures are defined by a set of mechanisms that collectively outline the operational structure of the framework. They strive to capture the essence of human cognition by translating it into computational terms, thereby advancing understanding and the ability to simulate intelligent behavior.

In the following sections, we will present three such architectures, each with a different focus: psychological modeling, computational aspects of problem solving, and biological plausibility of the underlying processes.

## 16.1. ACT-R[11]

One of the best known classic cognitive architectures is the *adaptive control of thought – rational* (ACT-R). Developed by John Anderson (1983; 1996), it offers a framework for understanding human cognition. The architecture is rooted in the idea that human cognitive processes result from the interaction of multiple modules, including perception, memory, and reasoning. These modules operate concurrently and in a coordinated manner, with ACT-R delineating the interconnections, their interaction with the environment, and their adaptability for various tasks. The architecture distinguishes between procedural and declarative memory, and a modular structure in which components communicate through buffers.

ACT-R serves as both a cognitive architecture and a theory of cognition. It posits a fixed set of mechanisms that use knowledge to execute tasks, thus predicting and explaining the sequential steps of cognition that underline human behavior. The theory has undergone continuous evolution over more than four decades, leading to variations in the acronyms used to describe it, such as the alternate term "atomic components of thought" (Anderson & Lebiere, 1998).

The evolution of ACT-R traces back to its origins as a model of human memory, later maturing into a unified theory of cognition. One of its main strengths lie in modeling memory, a trait shared with the *human associative memory* (HAM) model (Anderson & Bower, 1973), which laid the groundwork for a theory of cognition. This model operates on the concept of information processing through buffers, as well as coding and recoding external stimuli into usable chunks connected to one another.

ACT-R is a hybrid architecture, incorporating both symbolic elements such as rules and declarative memory and subsymbolic components that modify their relationships and usage. An important moment in ACT-R development occurred with the introduction of rational analysis of cognition (Anderson, 1990). This methodology addresses the challenge of understanding the complexity of the human mind by assuming rationality (i.e., optimality) in the corresponding mechanisms. The underlying belief was that human cognitive processes represent an evolutionary local maximum of adaptability, leading to optimal memory retrieval and decision-making mechanisms. This assumption of optimality helps to narrow down the search space for the underlying mechanisms of human cognition. Thus, the ACT theory evolved into ACT-Rational, marked by revisions to memory and learning equations that better describe human processes.

---

[11] This section uses information and ideas from: (Anderson, 1989), (Anderson & Lebiere, 1998), (Whitehill, 2013), and (Ritter, Tehranchi & Oury, 2019).

*16.1.1. General Architecture*

The architecture comprises several interconnected *modules*, each responsible for specific aspects of cognitive processing (Figure 16.1). Each module handles particular types of information or cognitive functions. The main modules of ACT-R are presented as follows.

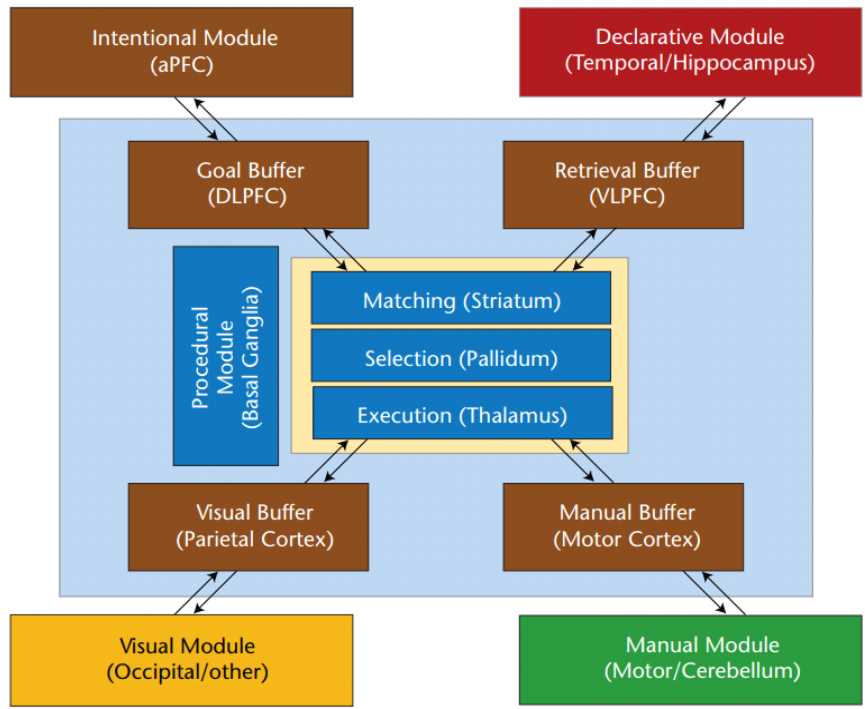

**Figure 16.1.** The architecture of ACT-R
(Laird, Lebiere & Rosenbloom, 2017)

The *declarative memory* mimics human memory, and contains knowledge *chunks*, which represent the currently known facts and active goals. The chunks can be considered to correspond to concepts defined by a list of symbolic properties. Memory retrieval may involve search terms, and strategies are needed if a memory cannot be retrieved. It is influenced by factors such as frequency of use, threshold, and noise. The architecture accounts for phenomena like the fan effect (when the time to access a memory is influenced by its connection to other frequently accessed memories), and some models use partial matching of rules, blending of memory retrievals, forgetting, and instance-based learning.

The *goal module* keeps track of the current goal of the system, and maintains the current task state. Its actions include creating and updating goal chunks, as well as saving or removing goal chunks into/from the declarative memory. ACT-R operates with a single goal stack, which means it does not explicitly address how to handle multiple conflicting or competing goals that often arise in real-life situations. In addition, the idea of goal preemption, where a more important goal takes precedence over a less important one, is not fully developed.

The *visual module* is responsible for identifying objects in the visual field. In general, it manages the perceptual input from the environment, and can simulate eye fixations and movements. It includes the "where" (location) and "what" (object identification) buffers, analogous to the components of the human visual system. The attention to a location enables the extraction of visual information, which becomes available to the central production system for decision making. The architecture can also include an *aural module* for auditory information.

The *motor module* controls the outputs of the system in the environment, e.g., movements and actions. It receives command requests through the motor buffer, as to what actions to perform in response to the goals of the system. The architecture can include several effector modules, for example, the "manual" and the "vocal" modules.

The *central production system* (CPS) coordinates the communication among modules using production rules – or, more simply, productions. It is the core component that manages the interaction between different modules. It applies production rules to recognize patterns in the buffers (described below), selects a matching rule, and updates the buffers accordingly. Production rules consist of conditions (the *if* part or the preconditions) and actions (the *then* part), and they drive the decision-making process in the system. When the preconditions of a rule are true, the rule may be activated or "fire". The CPS also includes a conflict-resolution phase to handle the situations when the preconditions of several rules match the current situation and to select only one rule that will actually fire. A rule can modify various buffers, including those of the visual and the motor modules.

We have mentioned the concept of buffers several times. *Buffers* in ACT-R are temporary storage locations that hold information for processing. They act as interfaces between the cognitive modules. They can receive information from the external environment, such as sensory input as well as other task-related data, and provide a storage area for this information before it is processed by the modules. Buffers also allow modules to expose only a subset of their data (what is relevant to the current task) to other modules.

The interaction among the modules occurs in *cycles*, with the CPS coordinating the flow of information and actions. A whole cognitive cycle, involving the recognition of patterns in buffers, rule selection, application, and subsequent buffer updates, takes about 50 ms, which is considered to be the minimum cycle time for cognition in several cognitive architectures. Time estimates for actions are based on human performance, but the system is capable of running in real-time or at a simulated faster pace. The architecture allows for modeling individual differences by adjusting parameters in equations governing action prediction and learning through rule learning and changes to chunk activation.

The following example – adapted from (ACT-R Research Group, 2014) – shows the basic structure of knowledge chunks and goals in the declarative memory and the production rules (with a Lisp syntax). The example is about classifying some animals based on their properties. Although the original example contains several animals represented within a semantic network, we will consider a simplified example with one animal (canary) and one category (bird). The chunks "property" and "is-member" are defined as:

```
(chunk-type property object attribute value)  ;; property has three slots: object, attribute, and value
(chunk-type is-member object category result)
```

The following chunks include information related to a canary:

```
(p1 ISA property object canary attribute color value yellow)
(p2 ISA property object canary attribute sings value true)
(p3 ISA property object canary attribute category value bird)
```

The goal is to check whether the canary is a bird.

```
(g ISA is-member object canary category bird)  ;; the goal has no "result" slot yet
```

The problem can be solved using two production rules. The first one requests the retrieval of categorical information and the second one uses that information to set the "result".

```
(p retrieve
  =goal>                      ;; IF part: tests whether the goal contains the ISA, object and category slots
    ISA  is-member            ;; ISA must match the "is-member" value (the category membership)
    object  =obj              ;; =obj is a variable, it will be assigned the matched value of the "object" slot
    category  =cat            ;; =cat is the variable for the "category" slot
    result  nil               ;; without a "result" slot
==>                           ;; THEN part
  =goal>                      ;; change the goal
    result  pending           ;; to add the "result" slot with the "pending" value
  +retrieval>                 ;; request a memory retrieval of a chunk
    ISA  property             ;; where "ISA" is "property"
    object  =obj              ;; =obj is a variable matched with the value of "object "
    attribute  category       ;; "attribute" must be "category"
)

(p verify
  =goal>                      ;; if a goal is matched
    ISA  is-member
    object  =obj
    category  =cat
    result  pending           ;; with "result" "pending"
  =retrieval>                 ;; and a chunk has been retrieved
    ISA  property
    object  =obj
    attribute  category
    value  =cat
==>
  =goal>                      ;; then modify the goal
    result  yes               ;; to set the "result" to "yes"
)
```

This example shows how the preconditions match the slot values of the chunks with required values or into variables. It also demonstrates some basic operations to retrieve chunks from the declarative memory.

### 16.1.2. Decision Making

As we have seen, declarative memory is comprised of knowledge chunks that can satisfy the preconditions of production rules. When all conditions of a production rule are met, it is considered a match. When multiple rules match, a decision is needed on which one to execute.

The decision-making process relies on the expected value ($v$) associated with each matching production rule. This value is learned by the agent through experience and is determined by factors such as the probability of goal achievement ($p$), the reward for achieving the goal ($g$), and the total expected cost ($c$) related to the execution of the rule:

$$v = p \cdot g - c \tag{16.1}$$

After matching a rule $s_i$, the decision involves comparing the expected gain of waiting for a potentially superior rule (with a higher value $c_j$) against the cost of waiting. The cost of waiting is represented as a constant that approximates the memory retrieval cost of matching another production rule in the future.

One decision-making mechanism involves a binary decision at each time step to either accept the current rule as the best choice with a certain probability or to wait for a potentially higher-valued rule to match.

If, at a given time, another rule with a higher value matches, the decision process resets, considering the new value as the baseline. This iterative process continues, with the agent dynamically assessing whether to accept or wait based on the evolving values of matching production rules. A somehow equivalent mechanism assumes that the agent uses a fixed waiting time during which it will wait for better production rules before firing the matched one.

### *16.1.3. Learning*

Learning in ACT-R involves the creation of new knowledge chunks and production rules, as well as the strengthening of memories through use.

The initialization of declarative memory in ACT-R involves either the creation of new chunks as encodings of external events or the writing of the chunks to memory as a result of an executed production rule. One of the most important learning mechanisms in ACT-R is the strengthening of declarative memories. When memories are retrieved and used, their activation is increased. The level of activation affects the retrieval time of a memory, with higher activation leading to quicker retrieval.

The strength of a chunk in the memory is denoted as its *activation*. Activation is increased when the chunk is used, i.e., matched in a production rule that fires. This aligns with the notion that practicing or using something repeatedly makes its memory stronger, and its recall faster.

The activation of a chunk $i$ is defined as (Anderson, 1993):

$$A_i(t) = B_i(t) + \sum_j w_j \cdot s_{ij} \tag{16.2}$$

where $B_i(t)$ signifies the base activation of the knowledge chunk at time $t$, and the summation accounts for the associative strength of the chunk $i$ with related chunks $j$ that are part of its context, e.g., simultaneously processed for the same goal. **w** represent the salience of chunks $j$ and **s** represent the strengths of association to $i$ from $j$ in the present context.

The *base activation* $B_i(t)$ increases when $i$ is used at time steps $t_k$:

$$B_i(t) = \ln \sum_k t_k^{-\delta} + \beta \tag{16.3}$$

In this equation, $\beta$ is a constant and $\delta$ is the decay rate.

The activation of a chunk is meant to express the log-odds that at time $t$, the chunk will match a rule that will fire. The relation between the log-odds $l$ and probability $p$ is: $l = \ln(p / (1 - p))$.

The *power law of learning* states that the time it takes to correctly recall a piece of information (latency) decreases over time as a power function which has the general form $L(t) = a \cdot t^{-k}$. In ACT-R, the *latency* of recall of a chunk is defined in a similar way (Pavlik, Presson & Koedinger, 2007):

$$L_i(t) = \phi \cdot e^{-A_i(t)} + \kappa \tag{16.4}$$

where $A_i$ is the chunk activation, and $\phi$ and $\kappa$ are constants.

In terms of creating new production rules, when two rules fire in close proximity, they can be merged into a single rule, and the timing of rule firings can in fact be influenced by the retrieval of declarative memory. The creation of new production rules involves several mechanisms:

- *Proceduralization.* Production rules frequently consist of variables that reference particular values governing their operation. With repetitive task execution, proceduralization facilitates the emergence of new productions within procedural memory, embedding hardcoded parameter values. For example, if the addition "1 + 2" is repeatedly performed, a new production rule may be created to simplify the process, such as: "If the goal is to add 1 and 2, then the result is 3";
- *Composition.* If two productions perform problem-solving steps sequentially, they can be combined to save computational resources. For example, let us consider the equation $(2/3) x = 1$. Assume $P_1$ multiplies both sides of the equation by 3, and $P_2$ divides both sides by 2. Then, a new production rule $P_3$ can combine these operations and multiply by $3/2$;
- *Generalization and discrimination.* Generalization takes place when multiple similar production rules have been stored, allowing the system to learn a more general rule inductively. For example, knowing the plurals "cats" and "dogs", the agent may generalize that adding an "s" forms plurals. However, there are exceptions to this, and additional rules can be learned to handle them.

Similarly to the activation of chunks, each production rule has a *production strength*, representing the log-odds that the rule will fire at time $t$ (Anderson, 1993):

$$S_r(t) = \ln \sum_k t_k^{-\delta} + \beta \tag{16.5}$$

The *latency* of a production rule is the period between the first moment when all of the knowledge chunks in its precondition were in memory, to the time when $r$ actually matches:

$$L_r(t) = \sum_i \beta \cdot e^{-\zeta \cdot (A_i(t) + S_r(t))} \tag{16.6}$$

In these equations, $\beta$ and $\zeta$ are constants.

## 16.2. SOAR[12]

The *states, operators, and reasoning* (SOAR) architecture (Laird, Newell & Rosenbloom, 1987), is a descendant of the Newell and Simon's *logic theorist* (1956), the first automated reasoning program, and the *general problem solver* (1963). Inspired by human cognition and aligned with Allen Newell's vision of a unified theory of mind, SOAR was designed with the goal of addressing problem solving across various scenarios, based on the premise that there are regular patterns in cognitive processes. It primarily employs symbol structures for knowledge representation, supplemented by numeric metadata to guide retrieval and learning. While ACT-R aims to model human performance in psychological experiments, SOAR focuses on computational perspectives.

### 16.2.1. The Problem-Space Computational Model

SOAR implements the *problem-space computational model* (PSCM), a framework for organizing knowledge and behavior in intelligent agents. Rooted in Newell and Simon's research on human problem solving, it views behavior as sequences of decisions towards achieving goals. Task environments, identified by the relevant aspects in an environment for a specific task (like playing chess), are central to their analysis. The authors conceptualize the space of potential actions within these environments, giving rise to the idea of *problem spaces*, essential for reasoning and decision making. The PSCM evolved to refine these problem spaces as the core of a cognitive architecture, which results from understanding how an agent interacts within task environments, and from identifying the relationships between the task, environment, and agent characteristics.

The problem space framework encapsulates an agent's decision-making process within a problem-solving context. The agent exists in a state, facing a choice of actions represented as operators. Once selected, an operator alters the situation, moving the agent to a new state, either through external action or internal changes. A problem space encompasses the states accessible to the agent via its operators, defining a problem as an initial state and a set of desired states. To solve a problem, the agent navigates from the initial state to desired states by selecting and applying operators, a process termed "problem space search". Solving a problem depends on factors such as the number and availability of operators and the agent's knowledge. Limited knowledge forces the agent to explore the problem space through trial and error, while greater knowledge enables direct operator selection. However, if the desired state does not exist within the problem space, the agent must either alter its approach or abandon the task.

In the problem space framework, the agent consistently resides within a state, distinguished between internal and external contexts. For internal problems, the agent entirely controls its state. In external scenarios, the agent's state is shaped by sensory input, internal interpretations, and structures derived from memory-based inferences, representing a partial snapshot of the environment. States are unconstrained in their representations, using both symbolic and numeric data.

Operators effect persistent state changes and hold preconditions, validating object properties or relations before application. Upon operator selection, actions are executed, possibly comprising multiple actions per operator. For internal problems, two approaches exist: one destructively modifies the current state, while the other creates a new state while preserving unchanged structures. External problem solving, however, follows to the former approach due to practicality and reactivity concerns, maintaining a single state altered by operator actions or environmental dynamics. The framework allows only one operator selection at a time, prohibiting simultaneous parallel execution. Conflict resolution between conflicting actions is not needed due to this singular operator selection premise.

---

[12] This section uses information and ideas from: (Hastings, 1997), (Laird, 2012), (Laird & Derbinsky, 2016), (Laird, 2022), and (Laird, et al., 2023).

Knowledge search involves retrieving relevant information stored in the long-term memory, essential for operator selection. The relationship between problem search and knowledge search constitutes the core of an agent's behavior. Reactivity in an agent depends on the computational cost of knowledge search; thus, cognitive architectures must efficiently organize and retrieve knowledge to preserve reactivity. Problem search involves navigating through a problem space, which consists of states and operators, to reach a desired goal state by selecting and applying operators. Knowledge search, on the other hand, involves retrieving relevant information stored in long-term memory to guide the selection and application of operators during problem search, i.e., using past experiences or acquired knowledge to choose the best path forward in solving a problem.

*16.2.2. General Architecture*

Similar to ACT-R, the architecture of SOAR (Figure 16.2) consists in a network of task-independent modules that work collaboratively to define the agent's behavior. These modules include a variety of functions, including short-term and long-term memory systems, processing components, learning mechanisms, and interfaces for facilitating communication between them.

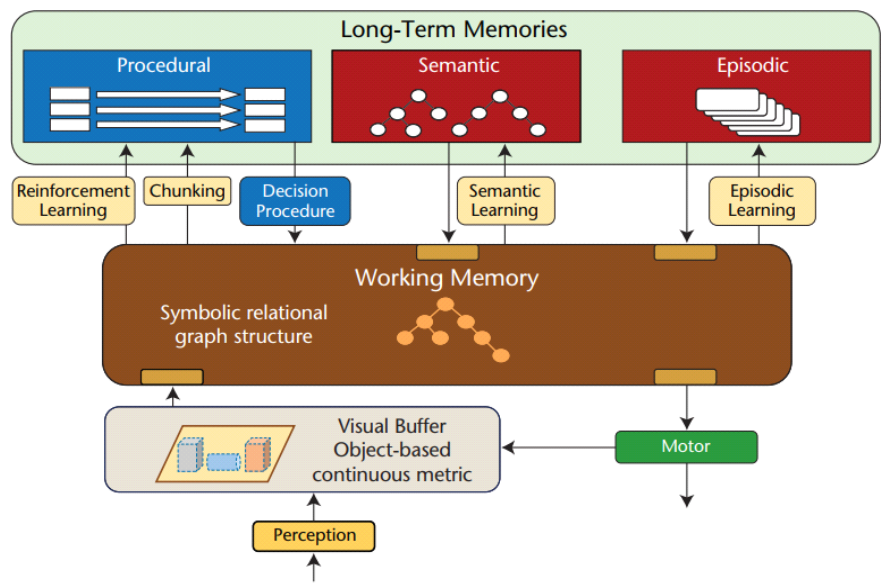

**Figure 16.2.** The architecture of SOAR
(Laird, Lebiere & Rosenbloom, 2017)

At the core of its structure is the *working memory*, which bears the responsibility of maintaining the agent's situational awareness. It serves as a repository for a wide array of information, including perceptual input, intermediate reasoning outcomes, active goals, hypothetical states, and specialized buffers for interacting with other memory systems.

Unlike ACT-R, SOAR features three long-term memory systems primarily responsible for storing symbolic knowledge: the *procedural memory*, which contains skills and "how-to" knowledge, the *semantic memory*, which contains facts about the world and the agent, and the *episodic memory*, which preserves memories of past experiences. Procedural memory dynamically responds to the contents of the working memory, and can actively modify it. It initiates retrievals from the semantic and episodic memory into the working memory and can initiate actions through interfaces such as the *spatial-visual system* and the *motor system*.

Learning mechanisms, closely associated with both procedural and episodic memories, allow the agent to acquire new knowledge and adapt its behavior over time. Agent behavior emerges from the interaction between working memory, representing the agent's current goals and situation, and procedural memory, which includes its repository of skills and processing knowledge.

### *16.2.3. The Spatial-Visual System*

The spatial-visual system (SVS) is a key component of the architecture designed to bridge the gap between symbolic and subsymbolic representations. SVS enables SOAR agents to efficiently integrate abstract knowledge with real-world perceptual data. SVS serves multiple roles, including grounding symbolic knowledge in perceptual data and supporting non-symbolic data processing. It acts as an intermediary between symbolic working memory, perception, motor control, and long-term, modality-specific memories.

This system accommodates both 2D and 3D representations, allowing agents to issue commands to SVS through operators to extract symbolic properties and relations from non-symbolic data. It also enables hypothetical reasoning over spatial-visual representations, facilitating the simulation of potential actions, outcome prediction, and decision making based on the current state of the environment. In practical scenarios, SVS is essential for tasks like robot arm movement planning, where it maintains parallel representations of real-world objects and symbolic knowledge.

### *16.2.4. Working Memory*

Working memory serves as the primary store for the current state of information. It holds facts, goals, and problem-solving steps for ongoing tasks. Working memory is dynamically updated when sensory data arrives, and productions (rules) fire and alter its contents. Representing the immediate context of the system, it stores both temporary and long-term information, facilitating decision making by providing access to relevant knowledge. The content of working memory influences the selection of actions, aiding in the continual processing and refinement of potential solutions within the cognitive system.

Its content is represented by triplets in the form of (state, attribute, value). For example, a graph (where S1, I1, etc. are nodes) can be represented as follows:

```
(S1 ^io I1 ^superstate nil ^type state)
(I1 ^input-link I3 ^output-link I2)
```

Likewise, some situation in the blocks world can be represented as:

```
(s1 ^block b1 ^block b2 ^table t1)
(b1 ^color blue ^name A ^ontop b2 ^type block)
(b2 ^color yellow ^name B ^ontop t1 ^type block)
(t1 ^color gray ^name Table ^type table)
```

*16.2.5. Semantic Memory*

Semantic memory focuses on encoding factual knowledge about the agent and the environment, including both general, context-independent world information and specific details about the agent's environment, capabilities, and long-term goals. In contrast to procedural memory, semantic memory encodes knowledge using symbolic graph structures rather than rules. A retrieval from semantic memory is initiated by creating a cue in the semantic memory buffer, which partially specifies the concept to be retrieved.

SOAR employs activation mechanisms such as base-level activation and spreading activation during retrieval, which simulate human access to long-term semantic memory by considering factors such as recency, frequency of access, and contextual relevance.

Semantic memory is stored separately from working memory to manage the computational cost of matching procedural knowledge against a growing working memory. It can be initialized with knowledge from external sources or incrementally built up during agent operations, including information about the environment, language processing, interactions with other agents, and hierarchical task structures learned from instructions. However, SOAR lacks an automatic learning mechanism for semantic memory.

*16.2.6. Episodic Memory*

Episodic memory is a distinct memory type designed to capture an agent's past experiences over time. Each episode represents a snapshot of the structures in the agent state at a specific moment, enabling the agent to recall the temporal context of past experiences.

Similar to semantic memory, episodic memory retrieval is initiated through a cue created in the episodic memory buffer by procedural knowledge. However, in contrast to semantic memory, episodic retrieval cues represent partial specifications of complete states, not single concepts. The retrieval process includes recency bias, and the retrieved episode is recreated in the buffer, allowing the agent to review past experiences.

SOAR stores only the differences between episodes and employs indexing to minimize retrieval costs. While the memory size increases over time due to the automatic storage of new episodes, retrieval costs for older episodes rise gradually with the growing number of stored episodes. Agents have control over memory costs by selecting which aspects of the state to store, often excluding frequently changing low-level sensory data. Episodic memory supports various capabilities, including virtual sensing of remembered locations and objects, learning action models, operator evaluation knowledge, prospective memory, and the ability to reconstruct and learn from past courses of action.

*16.2.7. Producing Deliberate Behavior*

Compared to ACT-R, which relies on rules with preconditions and actions, SOAR has a more complex mechanism involving rules (with a different meaning than in ACT-R) and operators (which are actually responsible for performing actions – both internally, e.g., memory retrieval, performing mathematical operations, changing the memory content, and externally, i.e., interacting with the environment).

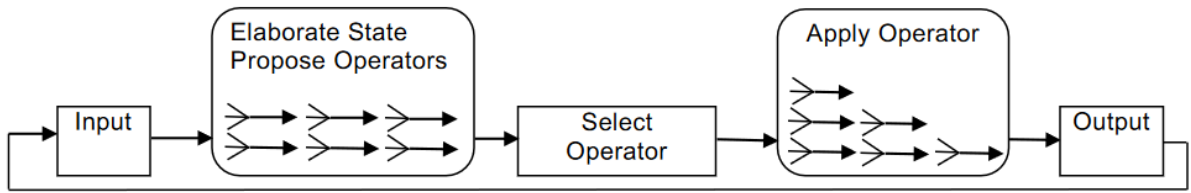

**Figure 16.3.** The decision cycle
(Laird, 2022)

The execution cycle of SOAR involves five phases (Figure 16.3):

- *Input*, where new sensory data enters working memory;
- *Proposal*, where rules interpret the data, propose operators, and compare them;
- *Decision*, where a new operator is chosen or an impasse leads to a new state;
- *Application*, where the chosen operators are applied, leading to changes in working memory;
- *Output*, where commands are sent to the external environment.

It is important to note that all rules fire and retract when conditions change, continuing in the proposal and application phases until a state of *quiescence* is reached, i.e., no further changes occur in the set of matching rules.

During these phases, if the preferences for the current operator change, the current state is immediately altered, but a new operator is not selected until the next decision phase. Additionally, if the rules proposing the selected operator retract during the apply phase, the phase immediately ends. Complex behavior arises from the execution of multiple decision cycles.

The process of working with operators in SOAR involves three rule functions: proposing potential operators, evaluating proposed operators, and applying the selected operator. In addition, there are rules for state elaboration.

In the following, we will present an example involving the blocks world (Soar Cognitive Architecture, 2020) that shows how operators are proposed and selected. Let us consider a situation with three blocks (*A*, *B*, *C*), where initially *B* is on top of *A*, and *C* is on the table: (on A Table), (on C Table), (on B A).

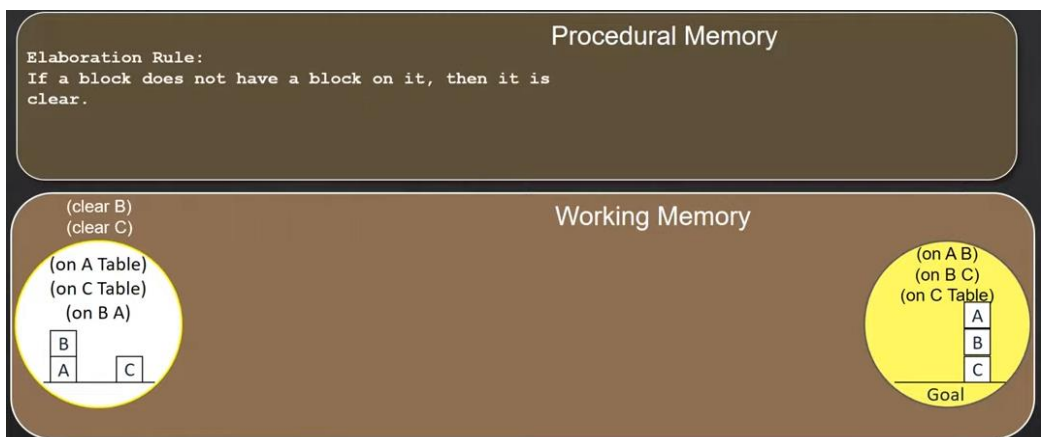

**Figure 16.4.** Example: state elaboration
(Soar Cognitive Architecture, 2020)

The first step is *elaboration*, where rules check whether certain objects or structures exist in the current state and propose corresponding operators. In Figure 16.4, we can see that two new facts (clear B) and (clear C) are introduced in the working memory, because the definition of the initial state did not include them explicitly.

Then, other rules examine the current situation and *propose* relevant operators based on specific conditions. This step often integrates task-specific knowledge to avoid overwhelming the agent with unnecessary operator suggestions (Figure 16.5).

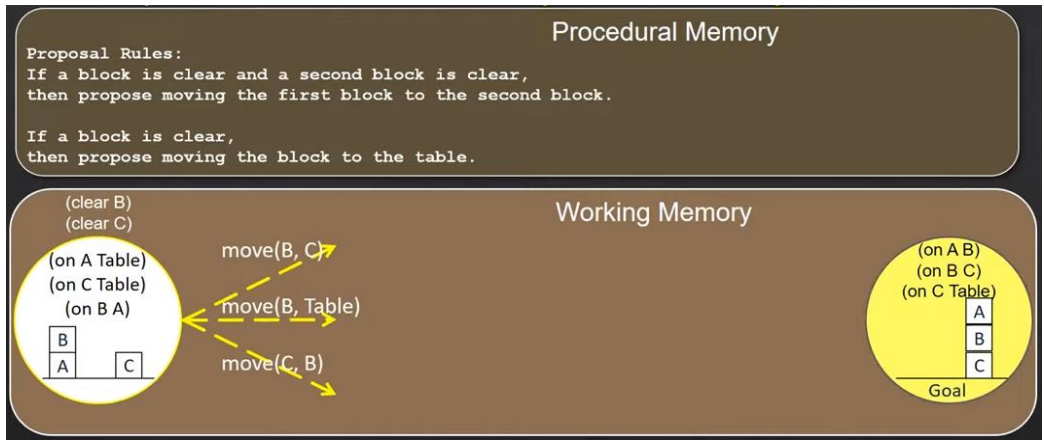

**Figure 16.5.** Example: operator proposal
(Soar Cognitive Architecture, 2020)

If another block *D* were clear on the table, operators for all the possible moves would be proposed, e.g., move(B, D), move(C, D), move(D, C), etc.

The third step is the *evaluation*, where operator evaluation rules assess the proposed operators and other contents of working memory, assigning preferences that guide operator selection. Preferences can indicate feasibility (+ accept / – reject) or desirability (> best / >> better / = indifferent / << worse / < worst). An acceptable preference states that the operator is a candidate for selection. In order to be further considered, candidates must be acceptable, and hence this is the default. For an operator to be selected, there must be at least one preference for it, specifically, a preference to say that the value is a candidate for the operator attribute of a state (usually done with the "acceptable" preference). There may also be others, e.g., to say that the value is the "best". Numeric preferences can encode expected future rewards of operators, often in conjunction with reinforcement learning mechanisms, presented in Section 16.2.10.

For our example, the evaluations are presented in Figure 16.6.

Since the operator ordering is clear, the next steps are to *select* the move(B,C) operator and to *apply* it in order to actually change the current state (Figure 16.7).

As mentioned above, SOAR allows multiple rule instantiations and unlimited access to declarative memory. However, it enforces a bottleneck by permitting only one operator to be selected at a time, using its preference mechanism. To maintain logical consistency, operators can automatically unselect if their preconditions are no longer met. Consequently, SOAR operators typically cannot execute long action sequences, which are instead executed by a series of operators. This impacts decision timing and granularity, requiring multiple operators for complex actions.

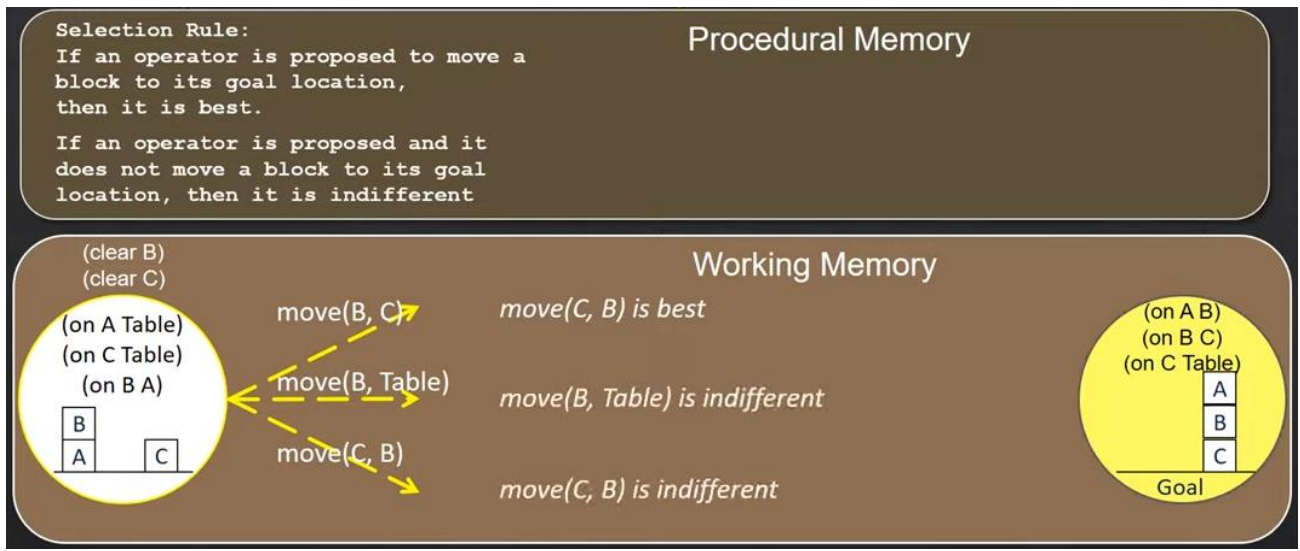

**Figure 16.6.** Example: operator evaluation
(Soar Cognitive Architecture, 2020)

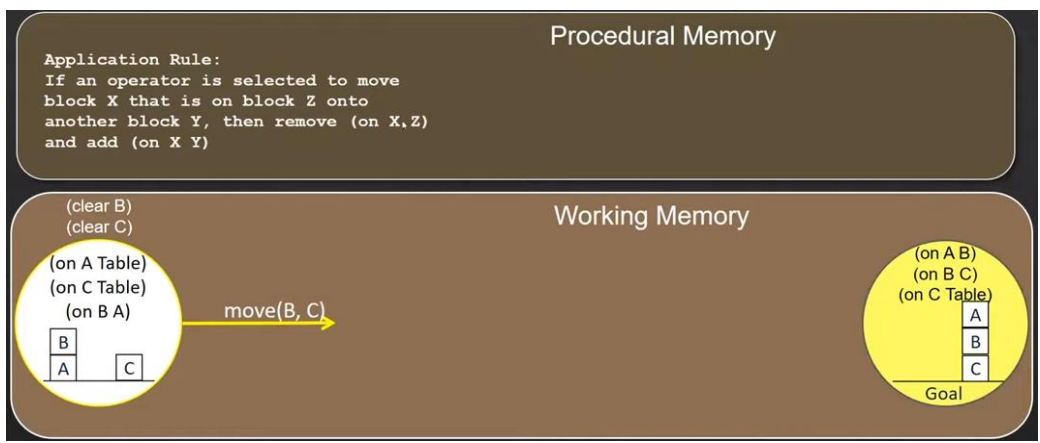

**Figure 16.7.** Example: operator application
(Soar Cognitive Architecture, 2020)

The operations described above are implemented in a specific programming language, which is quite complex. We include below an example (Laird & Congdon, 1996) related to moving a block including only "propose" and "apply" rules:

```
sp {blocks-world*propose*move-block  # sp = soar production
  (state <s> ^problem-space blocks  # the state has a problem-space named "blocks"
    ^thing <thing1> {<> <thing1> <thing2>}  # the block moved is different from the block moved to
  ^ontop <ontop>)
  (<thing1> ^type block ^clear yes)  # the block moved must be type block
  (<thing2> ^clear yes)  # the block moved and the block moved to must be both be clear
  (<ontop> ^top-block <thing1>
    ^bottom-block <> <thing2>)  # the block moved must not already be on top the block being moved to
-->
  (<s> ^operator <o> +)  # propose an operator with an "accept" preference
  (<o> ^name move-block
    ^moving-block <thing1>  # the substructure of the operator (its name, its "moving-block" and the "destination")
    ^destination <thing2>)}
```

```
sp {blocks-world*apply*move-block*remove-old-ontop
   (state <s> ^operator <o>   # an operator has been selected for the current state
      ^ontop <ontop>)
   (<o> ^name move-block  # the operator is named move-block
      ^moving-block <block1>  # the operator has a "moving-block" and a "destination"
      ^destination <block2>)
   # the state has an ontop relation, the ontop relation has a "top-block" that is the same as the "moving-block"
   # of the operator. the ontop relation has a "bottom-block" that is different from the "destination" of the operator
   (<ontop> ^top-block <block1>
      ^bottom-block { <> <block2> <block3> })
-->
   (<s> ^ontop <ontop> -)}   # create a "reject" preference for the ontop relation

sp {blocks-world*apply*move-block*add-new-ontop
   (state <s> ^operator <o>)
   (<o> ^name move-block
      ^moving-block <block1>
      ^destination <block2>)
-->
   (<s> ^ontop <ontop> +)  # create an acceptable preference for a new ontop relation
   # create default acceptable preferences for the substructure of the ontop relation:
   # the top block and the bottom block
   (<ontop> ^top-block <block1>
      ^bottom-block <block2>)}
```

## 16.2.8. Impasses

An impasse signifies a point where the agent cannot proceed due to insufficient or conflicting knowledge. There are three types of impasses, each linked to a specific kind of knowledge failure. A "state no-change" occurs when no operators are proposed, often indicating a need to create new operators relevant to the current situation. An "operator tie/conflict" arises when multiple operators are proposed, but evaluation preferences fail to determine which to select, and this needs preferences adjustment. An "operator no-change" happens when the same operator persists across multiple cycles, indicating insufficient knowledge to apply it or a situation where an action requires multiple cycles to execute externally.

  SOAR handles impasses by creating *substates*, allowing localized reasoning. These substates have their own preference memory, enabling the selection and application of operators, aiming to resolve impasses without disrupting the processing in superstates. Substates play a dual role as a state and a *subgoal*. They use a similar processing cycle as the *top state*, using procedural memory to match structures and propose, evaluate, and apply operators. If a substate lacks sufficient knowledge, it results in a new impasse, leading to a stack of substates. In this way, a hierarchy of operators and substates can be created. Each impasse type demands specific knowledge for resolution, e.g., creating new operators, altering preferences or creating or removing working memory elements to facilitate the selection of a new operator.

  In a blocks world scenario that also involves the robot arm (gripper) that moves the blocks, this hierarchical approach can deal with complex actions by breaking them into simpler operations. Consider the "move block" operator. Initially, only lower-level, primitive actions are available ("pick up", "move", and "put down"), corresponding to the physical capabilities of the gripper.

"move block" is in this case an abstract action formulated at a higher level. As the "move block" action is not directly executable, an impasse occurs due to the absence of rules for its execution. To resolve this, SOAR creates a substate, a smaller problem-solving context focusing on executing the "move block" action, operating as a subgoal for achieving the action. Within this substate, the agent searches for actions that match the abstract "move block" operation. This involves creating sub-substates for actions such "pick up", "move", and "put down". Finally, the agent executes these primitive actions in a step-by-step manner within these sub-substates, and is able to move the block (Figure 16.8).

This hierarchical approach allows the agents to handle complex actions by breaking them down into smaller, executable steps. By resolving impasses through nested problem-solving contexts, an agent progressively achieves higher-level tasks by performing simpler, more concrete actions.

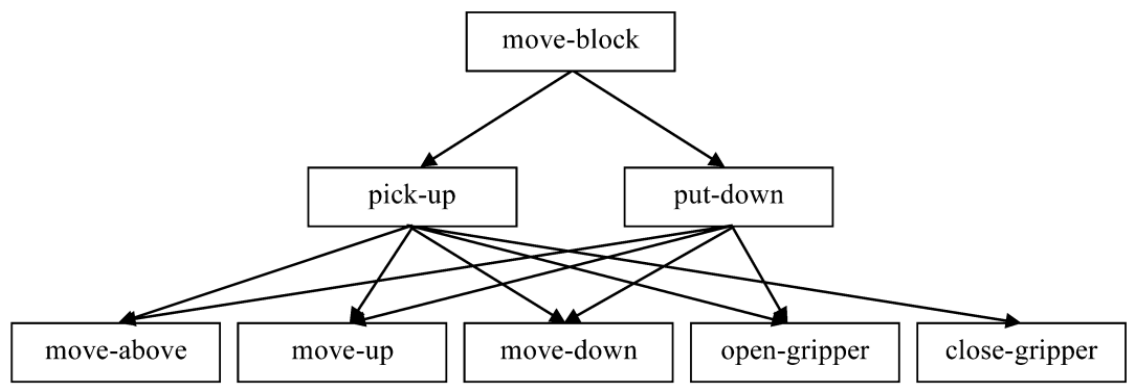

**Figure 16.8.** The operator structure of the hierarchical blocks world example
(Laird, 2012)

*16.2.9. Chunking*

Both ACT-R and Soar use the concept of "chunking", yet their implementations differ. In ACT-R, chunking involves grouping information into chunks, which aides memory and reduces cognitive load. These chunks enhance retrieval but do not impact the underlying rules. On the other hand, chunking in SOAR involves the creation of new rules based on past problem-solving experiences, allowing the system to generalize knowledge. The chunks in SOAR establish new operators or productions, altering the knowledge base of the system and affecting future decision making, whereas in ACT-R chunks primarily focus on enhancing memory retrieval efficiency without altering the rules themselves.

In SOAR, chunks are formed when a result is returned from an impasse, linking the pre-impasse situation to the outcome. They correspond to different impasse resolutions, generating specific types of chunks based on the resolution type.

When a substate processes information to overcome an impasse, chunking translates this processing into rules that replicate the successful results of the substate. Thus, chunking eliminates future impasses and the need for repeated substate processing. When an impasse occurs and a substate is generated to resolve it, chunking analyzes this sequential problem-solving process and translates the sequence of steps into a set of rules. These rules replicate the successful resolution achieved within the substate. Once these rules are established, the agent does not need to engage in the sequential, step-by-step reasoning process again when faced with a similar situation in the future; instead, it can immediately apply the rules generated through chunking in parallel.

Activated automatically when a substate yields a result, chunking reviews the historical trace of substate processing, and identifies critical structures in the superstate necessary to achieve the substate results. These structures form the conditions of the rule, while the outcomes become the actions of the rule.

### *16.2.10. Reinforcement Learning*

Reinforcement learning (RL) is integrated into the SOAR architecture as a means of refining operator selection based on feedback, including factors such as goal accomplishment and failure, which actually express rewards. RL is implemented through the creation of operator evaluation rules known as *RL rules*, which produce numeric preferences.

RL rules are designed to encode the states and proposed operators to which they apply, with their numeric preferences denoting the expected rewards (the *Q* values) for those specific states and operators. Following the application of an operator, all associated RL rules are updated in response to the acquired reward and the anticipation of future rewards. This update mechanism ensures that even states lacking direct rewards have associated operators conveying the expected reward back to the relevant RL rules. RL influences operator selection primarily in scenarios of uncertainty, supplementing non-RL preferences when they prove insufficient.

For example, RL rules can model how a robot moves and orients itself concerning an object it intends to pick up. Each rule tests various distances and orientations relative to the object, associating *Q* values with specific operators. With experience, RL rules adapt to favor operators that accomplish the task more quickly. Additional evaluation rules can be included to ensure collision avoidance.

Rewards within SOAR are generated by rules examining state features and defining a reward structure for the state. Rules can calculate rewards by evaluating intermediate states or converting sensory data into a reward representation.

RL rules can be acquired through chunking, with initial values set through substate processing and subsequent fine-tuning based on the agent's experience. Furthermore, SOAR naturally supports hierarchical RL across a range of problem solving scenarios, including model-based or model-free RL.

### **16.3. The Standard Model of the Mind**

The standard model of the mind (Laird, Lebiere & Rosenbloom, 2017) is envisioned as a way to achieve consensus within the cognitive science community about the fundamental components of a cognitive architecture. Its philosophy assumes that human-like minds can be described as computational entities whose structures and processes closely resemble those found in human cognition. They are computational models of human cognitive functioning. This proposal suggests that cognitive architectures provide the appropriate abstraction for defining a standard model. However, it is important to note that the standard model itself is not a cognitive architecture; instead, it is a conceptual framework.

The development of the standard model of the mind began with initial consensus discussions (Burns et al., 2014) and was later extended and refined through a synthesis of ideas from three existing cognitive architectures: ACT-R, SOAR and Sigma (Rosenbloom, Demski & Ustun, 2016).

The proposed model covers various aspects of cognitive architectures, including their structural organization, information processing, memory systems, content representation, learning mechanisms, and perceptual and motor functions. It aims to identify areas of agreement among cognitive architectures and areas where there may be differences or gaps in understanding.

The standard model assumes that computational capabilities similar to a physical symbol system are available. The *physical symbol systems hypothesis* posits that such systems have the necessary and sufficient means for general intelligent action (Newell & Simon, 1972). However, the model departs from the traditional view by not assuming that computation at the deliberate act level is purely symbolic. The model is agnostic about whether symbols are arbitrary labels or patterns over vectors of distributed elements (such as those presented in Section 12).

Non-symbolic (numeric) information in the model serves two roles. First, it represents explicitly quantitative task information, such as distances or times. Secondly, it annotates task information representations (both symbolic and non-symbolic) to modulate how they are processed. This numeric information acts as metadata about the data.

The standard model also acknowledges the need for statistical processing and incorporates forms of statistical learning, such as Bayesian and reinforcement learning.

The structure of the standard model defines how information and processing are organized into distinct components, since the computational mind is considered to be composed of independent modules, each with specific functionalities. Its fundamental components include perception and motor, working memory, declarative long-term memory, and procedural long-term memory (Figure 16.9). Working memory serves as the communication buffer between the other components. Each module can be further decomposed into submodules or multiple instances for different modalities (e.g., various perceptual and motor modalities). Long-term memories, both declarative and procedural, have associated learning mechanisms for automatic storage and modification of information.

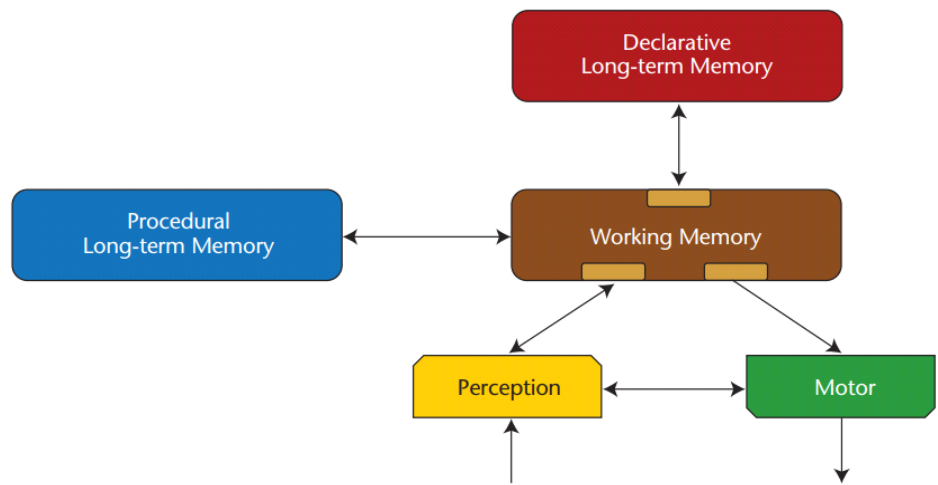

**Figure 16.9.** The proposed architecture of the standard model of the mind
(Laird, Lebiere & Rosenbloom, 2017)

The model works based on cognitive cycles, driven by procedural memory that initiates the processing required to select a single deliberate act in each cycle. Each action can modify the working memory, perform abstract reasoning steps, simulate external actions, retrieve knowledge from the long-term declarative memory, or influence perception. Complex behaviors, both internal and external, are considered to result from sequences of these cognitive cycles.

While there is significant parallelism within the internal processing of the procedural memory, the selection of a single deliberate act per cycle introduces a serial bottleneck, like the operator selection in SOAR. However, parallelism can occur across components, each with its own independent time course.

The standard model suggests that no additional specialized architectural modules are required for performing complex cognitive activities such as planning or language processing. These activities are achieved through the composition of primitive acts, involving sequences of cognitive cycles. Architectural primitives specific to these activities (e.g., visuospatial imagery for planning, phonological loop for language processing) can be included as needed.

The standard model assumes that all forms of long-term knowledge are learnable, including symbol structures and associated metadata. Learning is incremental and occurs online during system behavior, with experiences serving as the basis for learning. It typically involves a backward flow of information through internal representations of experiences. Longer-term learning can result from the accumulation of knowledge acquired over shorter-term experiences, possibly involving explicit deliberation over past experiences. Specific mechanisms are assumed to exist for all types of long-term memory, including procedural and declarative memories. Procedural memory involves at least two independent learning mechanisms: one for creating new rules based on rule firings and another for tuning the selection between competing deliberative acts through reinforcement learning. Declarative memory also includes at least two mechanisms: one for creating new relations and another for tuning associated metadata.

Multiple perception modules can exist, each focused on a specific sensory modality, e.g., vision, hearing, etc., and each may have its own perceptual buffer. An attentional bottleneck is assumed to constrain the amount of information that enters working memory, but the internal representation and processing within perception modules are not strictly defined. Information flow from the working memory to perception is possible, allowing expectations or hypotheses to influence perceptual classification and learning.

The motor modules are responsible for converting symbol structures and their associated metadata, stored in working memory buffers, into external actions. Multiple motor modules can exist, each corresponding to different effectors, e.g., arms, legs, etc.

## 16.4. Semantic Pointer Architecture[13]

The *semantic pointer architecture* (SPA) (Eliasmith, 2015) differs from traditional cognitive architectures by prioritizing biological principles. Unlike theories rooted in symbolic representations and processing, SPA integrates biologically plausible representations, computations, and dynamics within large-scale neural networks. This framework aims to mirror the cognitive processes observed in biological systems and tries to unify the modeling of diverse empirical data, both physiological and psychological.

### 16.4.1. Using a VSA for Representation

All computations in SPA are performed on abstract conceptual representations known as *semantic pointers*, which are high dimensional vector representations of the kind we have described in

---

[13] This section uses information and ideas from: (Eliasmith et al., 2012), (Bekolay et al., 2014), (Eliasmith, 2015), (Eliasmith, Gosmann & Choo, 2016), (Choo, 2018), and (Dumont et al., 2023).

Section 12. In fact, it relies on the holographic reduced representation (HRR) that allows the encoding and combination of various concepts, as well as the extraction of parts of the encoded combinations. However, the SPA operations do not depend on the specific VSA chosen, and the structure of the SPA itself remains independent of the VSA choice.

Semantic pointers are semantic because they have higher similarity rankings when conceptually similar and lower rankings when dissimilar. Similarity between semantic pointers with similar meanings is determined by their dot product. Semantic pointers that are similar in meaning produce higher dot product values when compared. They are referred to as "pointers" because they can be de-referenced to extract the encapsulated information.

In SPA, vectors are typically 512-dimensional. They are generated from a normal distribution with specific properties, such as an expected magnitude of 1 and uniform distribution of Fourier coefficients. The discrete Fourier transform is used to compute the Fourier transform of semantic pointer vectors. Semantic pointers maintain a fixed dimensionality regardless of the number of operations, ensuring the consistency of the neural implementation.

The binding operation involves circular convolution, which is computationally expensive but can be optimized in the Fourier domain, where it becomes element-wise multiplication of complex-valued vector coefficients. Circular convolution has properties that make it suitable for the manipulations of structured representation. It maps input vectors to approximately orthogonal results, meaning the dot product between the result and the original vectors is close to zero. This allows for the addition of new, unfamiliar items without making the result unrecognizable. When using VSA representations, information is gradually lost during binding operations, resulting in information reduction or compression. To handle these approximate results, a "clean-up memory" is required, which maps noisy representations back to allowable representations in the codebook, ensuring that unbinding results are recognizable.

### 16.4.2. The Structure of SPA

SPA addresses four cognitive aspects: semantics, syntax, control, and memory/learning. Traditionally, cognitive systems were characterized as symbol processing systems relying on syntax to convey semantics. Connectionist approaches focused on semantics, representing meaning in vector spaces. SPA tries to explain how semantic information is captured by semantic pointers and how representational structures are constructed, accounting for syntax. Thus, SPA combines syntax inspired by symbolic approaches and semantics inspired by connectionist approaches, using a biologically plausible substrate.

Semantic pointers can also be viewed from three perspectives: mathematically (vectors in high-dimensional space), physically (occurring activity in a neural network), and functionally (compressed representations pointing to semantic content).

Beside the representation of objects or concepts, semantic pointers can also be generated from perceptual input and used for motor actions.

A SPA subsystem working as a schema adaptable to specific brain areas is shown in Figure 16.10. Details such as hierarchy levels, transformations, control mechanisms, and error flow vary based on the specific brain region (e.g., vision, hearing, etc.). Figure 16.11 illustrates a higher-level schema for organizing SPA models, capturing their organizational structure. While not all models contain identical components, they ideally avoid conflicting elements.

The author suggests that a "whole brain" view can be achieved by linking these standard subsystems into a larger system. Their interconnectedness allows for the construction of unified models of biological cognition by proposing and refining such higher-level schemas, with the ultimate goal of "building a brain".

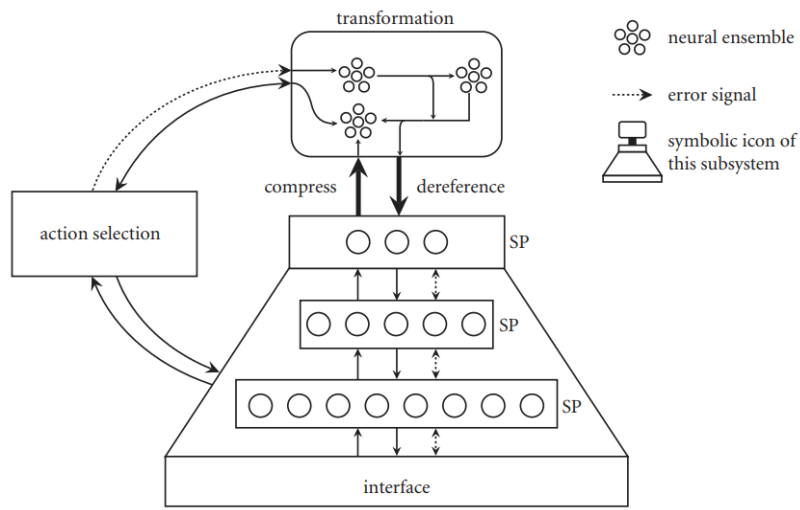

**Figure 16.10.** A schematic for a subsystem of the SPA
(Eliasmith, 2015)

In Figure 16.10, the information processing begins with a high-dimensional representation entering the subsystem, typically through a hierarchical structure, where it is compressed into semantic pointers. Moving within this hierarchy compresses or expands these representations. These semantic pointers can be altered and manipulated by other elements within the system. Transformations can be updated based on error signals, originating from both action selection and internal sources. The action selection component plays a role in directing information flow within the subsystem.

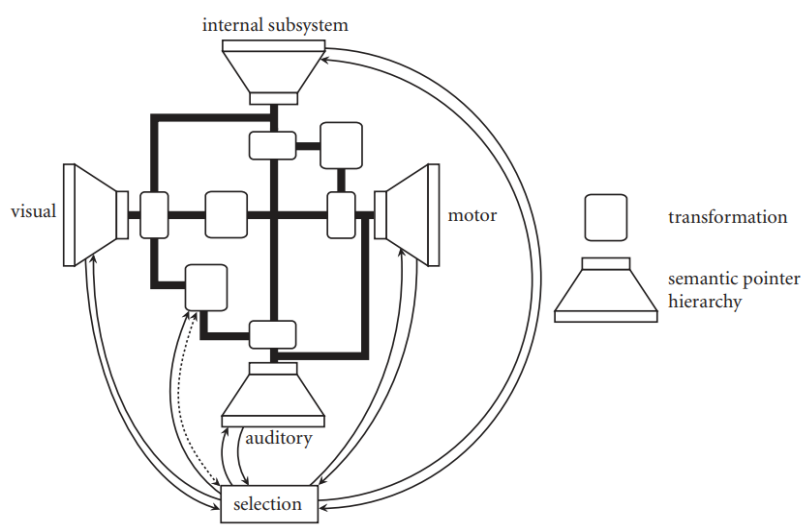

**Figure 16.11.** A schema of SPA models
The thick black lines show representation transfers and the thinner ones represent control and error signals.
(Eliasmith, 2015)

Figure 16.11 depicts multiple interacting subsystems similar to those in Figure 16.10. An "internal subsystem" is highlighted, emphasizing its importance in functions such as working memory and conceptual hierarchy encoding. Both action and perception involve hierarchies that operate in both forward and backward directions. These hierarchies help classify stimuli, generate semantics, and facilitate learning. The semantic pointers at the top of the hierarchy can be seen as pointing to memories and can be used to elicit details of motor or perceptual memories. These pointers facilitate constructive processes in the brain, allowing for movements and perceptions that have not been directly experienced. While top-level semantic pointers are the most compressed and typical, other levels in the hierarchy can also handle pointers to more detailed semantic information.

Biological systems do not process perception and action sequentially; they are processed concurrently, influencing each other throughout their respective hierarchies. Perception and action are deeply interconnected, and motor control relies on perceptual input. The interaction between perception and action occurs at multiple levels, making the system more dynamic and complex. Therefore, the entire perception-action system can be better understood as a series of nested controllers rather than separate hierarchies.

### 16.4.3. SPAUN

The principles of the SPA were embodied into the *SPA unified network* (SPAUN) (Eliasmith et al., 2012), a computational model designed to perform various cognitive tasks based on visual and symbolic input, and to manifest its responses through a two-joint arm. One of its design objectives is to be entirely self-contained, just like a biological brain, where all inputs are sensory information, and all outputs are motor actions. It unifies several previously built cognitive functions models, e.g., visual processing, working memory, and action planning using the SPA architecture.

Thus, SPAUN is constructed as the combination of six distinct models of various brain functions: vision, motor control, memory, inductive reasoning, action planning and control, and learning. Each individual model has its own way of representing information, but they use a common "language" within SPA.

SPAUN interacts with its environment through a single eye that perceives handwritten or typed digits and letters and a physically modeled arm with mass and length, which enables it to manipulate its surroundings. Its natural interfaces and internal cognitive processes allow it to perceive visual input, remember information, reason, and produce motor output, such as writing numbers or letters.

The functional architecture of SPAUN is shown in Figure 16.12. It consists of three hierarchies (visual, motor, working memory), an action selection mechanism (the basal ganglia model), and five subsystems for various information processing functions. This actually mimics the operations of various brain regions, and is based on the schema depicted in Figure 16.11.

The model can handle eight different cognitive tasks, each chosen to represent various challenges faced by biological cognitive systems. These tasks range from simple perceptual and motor tasks to more complex ones, such as question answering and fluid reasoning.

To initiate a task, SPAUN is presented with a letter-number pair (e.g., "A 4") that specifies the task to be performed. Subsequent input symbols guide the processing, leading to the generation of motor commands that produce a response. SPAUN waits for further input after responding to a task.

Semantic pointers are used with different sizes at different levels in the model. For example, 50D and 54D semantic pointers are used in the visual and motor hierarchies, respectively, and 512D semantic pointers elsewhere.

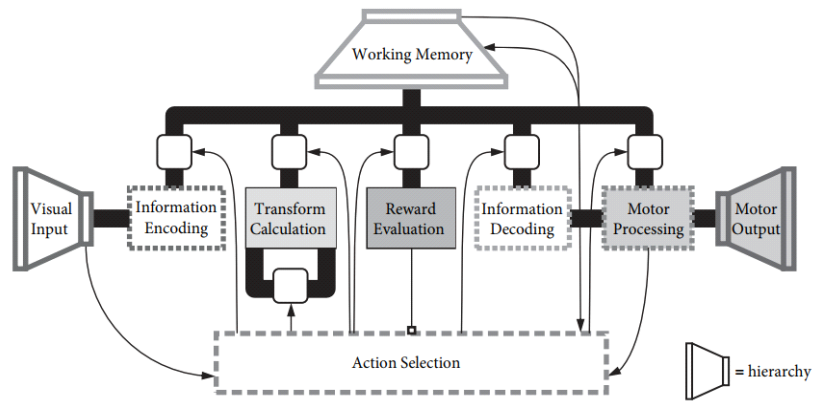

**Figure 16.12.** The functional architecture of SPAUN

The thick lines denote information exchange within the cortex, while thinner lines represent communication between the action selection mechanism and the cortex. Rounded boxes signify gating elements capable of regulating information flow within and between subsystems. The circular end of the line linking reward evaluation and action selection indicates that this connection influences connection weights.

(Eliasmith, 2015)

SPAUN is computationally intensive, requiring significant computational resources, such as 24 GB of RAM in year 2012. The model involves around 2.5 million spiking LIF neurons and approximately 8 billion synaptic connections, which makes it the world's largest functional brain model. It also incorporates four neurotransmitters (GABA, AMPA, NMDA, and dopamine) and simulates their known characteristics. Running one second of simulation time took approximately 2.5 hours (in the 2010s).

The model architecture is not specific to the tasks it performs. The eight tasks were chosen to demonstrate that despite being vastly different in nature, all of the models can be integrated into a functional end-to-end cognitive system. They are:

- *Copy drawing:* to show that the compressed semantic pointers generated by the visual system, and those used by the motor system retain enough feature information such that for each digit, a generalized relationship between the two sets of semantic pointers can be found;
- *Digit recognition:* to show that the vision system can recognize and reproduce the digits presented as inputs;
- *N-arm bandit task:* to show that the basal ganglia network model has the ability to adapt itself when provided with the appropriate error feedback;
- *List memory:* to show that the working memory model is able to remember and recall lists of digits after being presented;
- *Counting:* to show that the basal ganglia network model is able to perform an internally guided task (e.g., silent counting) in the context of a larger integrated system. A secondary goal is to demonstrate that information stored within the memory system can be modified in order to accomplish this task;
- *Question answering:* to show that the internal SPA representation is flexible enough to be probed for information using different types of queries;
- *Rapid variable creation:* to show the ability to perform an induction task which involves finding the variable inputs among a set of static digits;

- *Fluid induction:* to show the ability of performing a pattern induction task similar to the sequential variants of the Raven's progressive matrices. Raven's progressive matrices are non-verbal tests used to measure the abstract reasoning ability. They consist of visual patterns arranged in matrices, where test-takers must identify the missing piece that completes the pattern based on logic, reasoning, and understanding of spatial relationships.

### 16.4.4. NEF

In contrast with SPA, which focuses on high-level concepts ("what"), the *neural engineering framework* (NEF) (Eliasmith & Anderson, 2004) addresses the low-level implementation of these concepts ("how"), inspired by how groups of neurons perform various functions in the brain. NEF can be viewed as a "neural compiler" that models groups of neurons to realize high-level brain functions based on knowledge of individual neuron responses. NEF was used to implement the SPAUN model, and can be used in general to build computational networks.

NEF is designed based on three core principles:

- Neural *representations* involve nonlinear encoding and weighted linear decoding over neural populations and time;
- *Transformations* of neural representations depend on variables represented by neural populations and are determined using weighted linear decoding;
- Neural *dynamics* consider neural representations as state variables in dynamic systems, allowing analysis using control or dynamic systems theory.

In addition, it acknowledges the fact that neural systems are inherently noisy, and thus their analysis must account for noise effects.

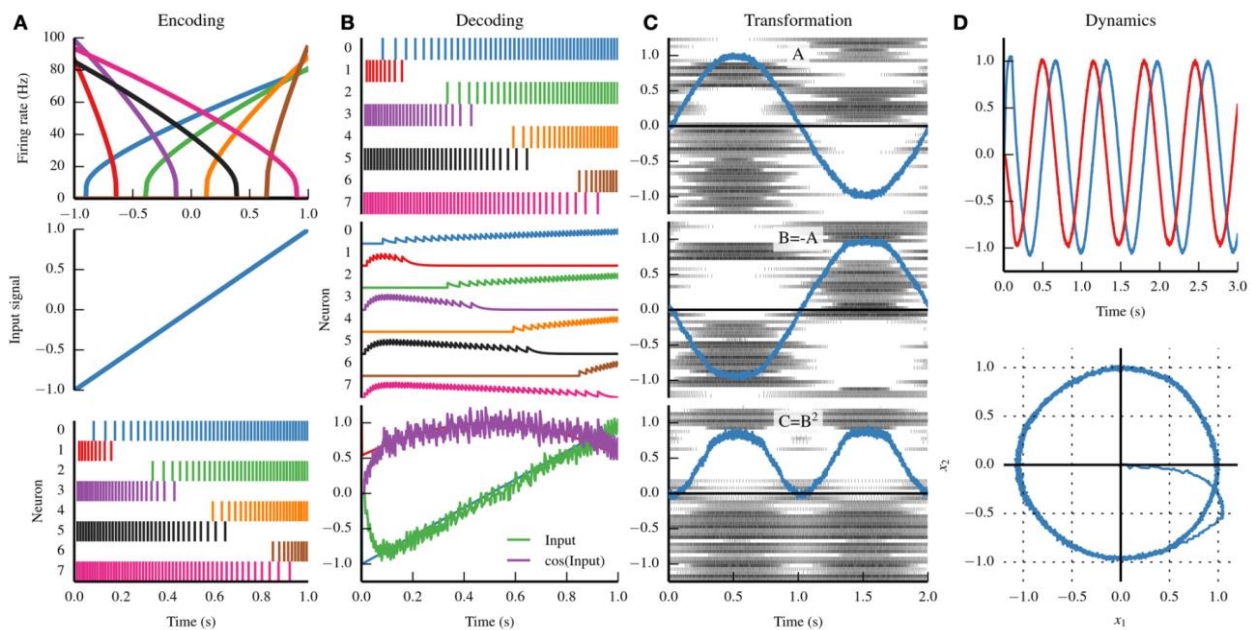

**Figure 16.13.** Examples illustrating the core principles of NEF
(Bekolay et al., 2014)

*16.4.4.1. Representation*

NEF operates by encoding information into time-varying vectors of real numbers, allowing mathematical manipulation of neural computations. Information encoding involves injecting specific currents into single neuron models based on the vector being encoded. The encoding process relies on tuning curves, which describe how likely a neuron is to respond to a given input signal. These curves depend on factors such as the neuron's gain (how quickly activity rises), bias (activity given no signal), and encoding weight (the direction in input vector space that causes maximum activity). Although NEF typically uses LIF neurons, the encoding process is not tied to any specific neuron model. When more neurons are involved, the original encoded vector can be estimated through a decoding process. This method extends the concept of population coding to handle vectors of any dimensionality.

In Figure 16.13.A1, eight neurons display distinct tuning curves as part of their encoding process. When presented with a time-varying signal (Figure 16.13.A2), they translate it into train spikes (Figure 16.13.A3). Their frequency is determined by how intensely the input signal activates the tuning curves at any given moment.

Decoding in NEF involves filtering the spike trains to account for postsynaptic currents, followed by combining them using weight values computed by solving a least-squares minimization problem. Thus, NEF can be used to compute arbitrary nonlinear functions in neural networks. By calculating decoders for neurons in a way that allows the neural population to represent the input value **x**, the same method can be used to solve for decoders enabling the computation of arbitrary functions. The decoders for a population of neurons computing a function $f(\mathbf{x})$ can be determined by using $f(\mathbf{x})$ in place of **x** in the decoder calculation. The approximation accuracy improves with the number of neurons in the network.

Figure 16.13.B shows the decoding process. Figure 16.13.B2 shows the integrated postsynaptic currents created by the spikes processed with a decaying exponential filter. This is in fact a very rough estimation of the input signal by each individual neuron. Then, in the simplest case, these currents are summed to create a more accurate representation of the input signal. However, the process uses a set of weights in the summation, which can be computed to account for transformations of the input signal, e.g., a cosine function (Figure 16.13.B3). The same encodings of the input signal are used to decode such functions of the input.

*16.4.4.2. Transformations*

We have already seen an example of transformation in the previous example with the cosine function with a single (time-varying) input dimension. However, this can be generalized to compute functions with input vectors of different dimensions encoded by different neuron populations. The connection weights linking neuron populations are determined as a product of decoding weights from the sending population, encoding weights of the receiving population, and any linear transformation needed to compute a specific function.

Figure 16.13.C1 we can see a sine input signal encoded by a population *A*. Populations of neurons have the ability to transmit signals to other groups by decoding the intended function from the former and subsequently encoding the decoded estimation into the latter. The combination of these two processes into a single step involves the computation of a set of weights describing the connection strength between the neurons in the two populations. In Figure 16.13.C2, we can see the negative of the input signal projected to population *B*, and in Figure 16.13.C3, the square of that signal projected to population *C*.

*16.4.4.3. Dynamics*

Some neural systems require persistent activity via recurrent connections. When these connections are present, the vectors represented by groups of neurons are similar to state variables within a dynamic system. The dynamics governing these systems can be studied using control theory methods. An example is presented in Figure 16.13.D, which shows the neural implementation of a harmonic oscillator with negative feedback across its two dimensions.

*16.4.5. NENGO*

The *neural engineering object* (NENGO) (Bekolay et al., 2014) is a graphical neural simulation environment that offers a practical way to apply the principles of NEF to model and simulate neural representations. The NENGO framework served as the backbone for SPAUN, and represents a platform for building large-scale neural models that mirror brain dynamics.

# 17. Discussion: Challenges for AGI Systems

## 17.1. Dual-Process Theory

In our discussion concerning the necessary capabilities of AGI, we will briefly delve into a concise presentation of the *dual-process theory* (Wason & Evans, 1975; Kahneman, 2011; Evans & Stanovich, 2013) in cognitive psychology. This theory distinguishes between type 1 and type 2 systems, each characterized by distinct modes of processing information and making decisions.

    *Type 1 systems* are often referred to as "intuitive" or "automatic". They are characterized by fast, effortless processing of information and are typically employed in situations where a quick decision is needed. These systems rely heavily on past experience and do not require much conscious effort or attention. An example of a type 1 system is driving a car by an experienced driver on a familiar route. Once the driver has become familiar with the route, he/she is able to drive without much conscious effort or attention. However, if the driver encounters a new and unfamiliar situation, such as driving in a foreign country, type 1 decisions are no longer possible.

    *Type 2 systems*, on the other hand, are often referred to as "rational" or "deliberative". They are characterized by slower, more effortful processing of information and are typically employed in situations that require careful consideration and decision making. These systems rely on conscious thought and attention and are more accurate and flexible than type 1 systems. An example of a type 2 system is solving a complex math problem. This task requires careful consideration, and relies heavily on conscious effort and attention. The problem solver must carefully analyze the problem, consider different strategies, and select the best solution. This type of task cannot be performed using a type 1 system.

    Type 1 processes are typically believed to generate default responses, they are fast and operate in parallel, functioning without depending on working memory. Moreover, they work unconsciously and are characterized by their associative nature, resembling the cognitive processes

observed in animals. Type 2 systems, in contrast, rely on the use of working memory and potentially involve mental simulations to estimate the consequences of actions. These processes operate slowly and in a serial manner, and possess a limited processing capacity. They make use of explicit knowledge, e.g., rule-based mechanisms. Notably, the type 2 processes are characteristic of human cognition.

While the dual-process theory provides insights into how individuals process information and make decisions, some criticisms were put forward, and several such cases are discussed in (Evans & Stanovich, 2013). For example, one common critique revolves around the multiple and sometimes vague definitions assigned to the two systems. In addition, some scholars argue that the binary classification oversimplifies the complexity of human cognition, suggesting instead that there exists a continuum of processing styles rather than two distinct types.

Current deep learning systems belong to type 1. Although they can solve complex problems, as mentioned earlier, the errors they occasionally make underscore their deficiency in type 2 capabilities. An illustrative example of this is the generation of adversarial samples for image recognition models. These samples involve making subtle, imperceptible changes in the input data, resulting in substantial alterations in the output of the model. In an extreme scenario, even a single pixel modification, imperceptible to the human eye, can cause the neural network to classify an object entirely differently. For example, a deer can be misclassified as an airplane and a baby bassinet as a paper towel (Su et al., 2019). Such adversarial attacks raise significant concerns, particularly in critical applications like autonomous vehicles, where slight changes in lighting conditions at some point may lead to an accident.

Still, the undeniable success of deep learning can also be explained by the fact that people themselves often carry out many of their daily actions instinctively, without extensive deliberation. They frequently rely on simple rules to navigate unfamiliar situations. The use of classical logic is not necessarily natural; it must be learned and applied through conscious effort.

A simple example is the difficulty of drawing sound conclusions in the presence of complex premises, or even when dealing with simple logical rules, but including unusual negations (Evans & Handley, 1999). Thus it is more difficult to use *modus tollens* with negative antecedents, such as: "If the letter is not B, then the number is 7. The number is not 7. Therefore, the letter is B" than with positive antecedents, such as: "If the letter is B, then the number is 7. The number is not 7. Therefore, the letter is not B."

The same limitations appear in tests such as the Wason selection test (Wason, 1968), which investigates deductive reasoning and conditional reasoning abilities. Participants are presented with a set of four cards, each containing information on one side. These cards typically feature a letter on one side and a number on the other. The participants' task is to determine which card(s) they need to flip over in order to determine whether a specific conditional rule, usually in the format "If *P*, then *Q*", is true or false. The test is designed to reveal the often counterintuitive nature of conditional reasoning, as participants tend to make systematic errors when attempting to identify the correct cards to flip. For example, given the cards: A D 3 7, and the rule "If there is an A on one side of the card, then there is a 3 on the other side of the card", only about 10% of the of participants select A and 7 (*P* and not *Q*), while many select only A (*P*), or A and 3 (*P* and *Q*) (Evans, Newstead & Byrne, 1993).

Neither identifying complex causal relationships through counterfactual reasoning is straightforward. For instance, during World War II, the British Royal Air Force wanted to protect their fighter planes from German anti-aircraft fire. The challenge revolved around determining the optimal placement of armor on the planes. An initial assessment based on the count of bullet holes in various parts of returning aircraft suggested that the armor should be added to the areas with the most damage. However, an insightful mathematician offered a different perspective: he pointed out that the areas without bullet holes were actually the vulnerable spots because the planes hit in those locations never returned to base. Consequently, the armor should have been placed in the areas with no bullet holes, i.e., on the engines (Ellenberg, 2015).

There are currently several attempts to combine the mostly neural type 1 systems with reasoning type 2 systems, resulting in hybrid neuro-symbolic systems. While there are a number of interesting papers addressing it, this direction is still in its early stages. As mentioned earlier, the reasoning component may not necessarily follow the principles of mathematical logic, but may instead employ specific human cognitive mechanisms.

The integration of these capabilities holds the potential for AI/AGI to leverage the strengths of both approaches, offering the possibility of meaningful analysis in addition to learning from available data. This combination has the potential to drive innovative advancements across the field and may bring us closer to singularity, i.e., the point where machines surpass human intelligence.

**17.2. Jackendoff's Challenges**

Ray Jackendoff (2002) identified four challenges for cognitive neuroscience. They were posed in the context of language processing, but they need to be addressed in the general sense if the goal is to design systems that exhibit AGI.

The first one is the *binding problem*, discussed in Section 13. Some proposals suggest that neural firing synchrony, where neurons encoding related features fire together, could solve this issue. However, when applied to linguistic structures such as sentences, the challenge deepens due to the complex interconnections among multiple elements.

The second challenge is the *problem of 2*, or the representation of multiple objects. It arises when identical words occur within a sentence or the same type of entities need to be processed. For example, in the sentence "The little star is beside a big star", it is not clear how the neural activations for the word "star" can distinguish its two occurrences. A possible solution is duplicating units in memory for each potential entity, but it encounters scalability issues when applied to larger contexts or longer sentences with a big vocabulary. An alternative solution suggests that working memory contains "dummy" nodes that act as pointers to long-term memory and encode the relationships among the items being pointed to, bound by temporal synchrony. However, there are still concerns regarding the neural plausibility of these proposals and the adequacy of temporal synchrony in this case.

The third one, the *problem of variables* addresses the limitation of encoding typed variables with the existing models of neural activations in order to handle relationships and linguistic rules effectively. Even simple relationships such as two words rhyming lead to difficulties in the encoding of their relationship. The brain cannot list all possible rhymes; also, people have the ability to recognize rhymes without explicit learning, even in foreign languages. Analogy and reasoning based on existing rhymes do not extend effectively to new words. Instead, rhyming must be encoded as a pattern with variables: "any phonological string rhymes with any other if everything from the stressed vowel to the end is identical in the two strings, and the onset preceding the stressed vowel is different".

The forth challenge is related to *binding in working memory vs. long-term memory*. A discrepancy is identified between transient (STM) and lasting (LTM) connections in memory for linguistic structure processing. Short-term connections are though to be linked to spreading activation or firing synchrony, while long-term connections are often attributed to the strength of synaptic connections. However, the combinatorial nature of language presents a difficulty because both transient and lasting connections may encode the same type of relations. For example, idioms such as "kick the bucket" need to be stored in LTM due to their non-literal meanings, yet they possess syntactic structures similar to combinatorically built phrases such as "lift the shovel". When retrieved during language processing, both types of structures are expected to have a similar instantiation in the brain's areas responsible for syntax, challenging the assumption that one is encoded through synaptic weights and the other through firing synchrony. The transfer from STM to structured LTM, especially in episodic memories brings an additional difficulty to the problem because in this case important elements seem to be stored immediately after only one occurrence, and not by some gradual strengthening of synaptic efficacies. Also, neurally-encoded information cannot be directly transferred to another memory area like in a computer system.

Although this is one of the Jackendoff's challenges, we should mention here that the generally accepted theory for the so called "memory consolidation" is that STM events are primarily stored in the hippocampus, which then replays them during non-REM sleep so as to send activation waves to the neocortex, leading to the gradual strengthening of cortical associations ensuring long-term storage. However, it is not obvious whether the cases of one-time events that are subsequently remembered for the rest of one's life are easily explained by this theory. There are studies that propose alternative explanations at least on some details of the theory, so we can say that the exact mechanisms behind the transfer from the STM to LTM have not been completely elucidated yet.

In one form or another, some of the methods presented in the previous sections proved to be able to address one or several of these challenges. Still, one would need to find a unified cognitive model that could consistently address all of them.

**17.3. Stability-Plasticity Dilemma**

The *stability-plasticity dilemma* refers to the balance between two essential processes in the brain. Stability refers to the preservation of established neural connections and the resistance to change in response to new information or experiences. Stable neural circuits are essential for maintaining long-term memories, well-learned skills, and consistent cognitive functions. On the other hand, plasticity is the brain's ability to change and adapt. It involves the formation of new neural connections, the strengthening or weakening of existing ones, and the ability to acquire new information and skills. Plasticity is essential for learning, memory formation, and recovery from brain injuries. An excessive focus on stability can hinder learning and adaptation, making it challenging to acquire new knowledge or adjust to changing environments. Conversely, too much plasticity can lead to instability, causing the loss of important information and disrupting established cognitive functions. Finding the right balance between stability and plasticity is crucial for the brain's optimal functioning, involving mechanisms that allow the brain to consolidate essential information while remaining flexible enough to incorporate new learning experiences. Cell assembly methods (Sections 6 and 7) and the ART model (Section 15.1) are, e.g., techniques that attempt to deal with this challenge.

**17.4. Symbol Grounding Problem**

The *symbol grounding problem* (Harnad, 1990) is another fundamental challenge in artificial intelligence and cognitive science. It deals with the issue of how symbols, which are abstract representations used in computation and communication, can be connected or "grounded" in the real world, enabling them to carry meaning that is understandable to humans. In AI, symbols are typically used to represent concepts, objects, actions, or ideas. For example, words in natural language are symbols that represent real-world entities. However, symbols by themselves lack inherent meaning. They are arbitrary and derive their meaning from the way they are used in a particular system or context. To be truly meaningful, symbols need to be connected to the real world, where they refer to specific objects, qualities, or actions. This connection allows individuals to understand and manipulate symbols based on their experiences and interactions with the physical environment. On the other hand, connectionist methods like NNs do not rely on manually devised symbols and rules. Instead, they can learn their representations from primary sources, such as sensory data. In this way, NNs can overcome issues specific to symbolic AI, e.g., brittleness in handling inconsistencies or noise, as well as the substantial human engineering effort. The representations learned by NNs are grounded in their input data, unlike symbols that rely entirely on human interpretation for their connection to real-world concepts (Greff, van Steenkiste & Schmidhuber, 2020).

# 18. Conclusions

So far, we have covered many concepts. Unlike typical "horizontal" reviews that focus on specific subfields, this review adopts a "vertical" approach. It aims to offer a panoramic view of selected concepts, spanning from low to high levels. It is also important to acknowledge that it is not exhaustive, and other valuable contributions about each topic are available. The selected works were those considered to be interesting for an AI audience.

     Rather than being initially designed in a top-down manner, the content gradually accumulated in a bottom-up manner and was structured afterwards, as it summarizes the author's exploration into neuro-symbolic methods. One surprising revelation from this exploration is that, despite active ongoing research, the fundamental cognitive mechanisms governing brain processes remain somewhat elusive. Various models attempt to clarify specific mechanisms and can be seen as pieces of a larger puzzle that still needs to be solved.

     These concepts can serve as building blocks for creating intelligent computational systems that can process information, learn from experience, and perform tasks with human-like adaptability and versatility, bridging the gap between human cognition and artificial (general) intelligence.

# Abbreviations

AGI: artificial general intelligence
AI: artificial intelligence
ALCOVE: attention learning covering map
AP: action potential
ART: adaptive resonance theory
ATRIUM: attention to rules and items in a unified model
BG: basal ganglia
BSC: binary spatter code
CA: cell assembly
CLS: complementary learning system
CNN: convolutional neural network
COVIS: competition between verbal and implicit systems
DBN: deep belief network
DFT: discrete Fourier transform
DIVA: divergent autoencoder
DPAAN: dynamically partitionable auto-associative network
EpCon: episodes to concepts
ESN: echo state network
ESP: echo state property
FHRR: Fourier holographic reduced representation
fMRI: functional magnetic resonance imaging
GABA: gamma-aminobutyric acid
GCM: generalized context model
HAM: human associative memory
HDC: hyperdimensional computing
HRR: holographic reduced representation
HTM: hierarchical temporal memory
HV: hypervector
IDFT: inverse discrete Fourier transform
IF: integrate and fire
kWTA: k winners take all
LCA: leaky, competing accumulator
LEABRA: local error-driven and associative, biologically realistic algorithm
LIF: leaky integrate and fire
LSM: liquid state machine
LTD: long-term depression
LTM: long-term memory
LTP: long-term potentiation
MAP: multiply, add, permute
MLP: multilayer perceptron
MTL: medial temporal lobe
NBP: neural binding problem
NEF: neural engineering framework
NMDA: n-methyl-d-aspartate
NN: neural network
NVAR: nonlinear vector autoregression
PCA: principal component analysis
PFC: prefrontal cortex
PSCM: problem-space computational model
RBF: radial basis functions
RBM: restricted Boltzmann machine
RC: reservoir computing
ReLU: rectified linear unit
RL: reinforcement learning
RP: random projection
RULEX: rule plus exception model
SDR: sparse distributed representation
SFA: spike frequency adaptation
SNN: spiking neural network
SOAR: states, operators, and reasoning
SPA: semantic pointer architecture
SR: successor representation
SSE: sum of squared errors
STDP: spike timing dependent plasticity
STM: short-term memory
SUSTAIN: supervised and unsupervised stratified adaptive incremental network
SVM: support vector machine
SVS: spatial-visual system
T2FSNN: time to first spike coding for deep spiking neural networks
TEM: Tolman-Eichenbaum machine
TPR: tensor product representation
VSA: vector symbolic architecture
WTA: winner takes all